\pdfoutput=1
\documentclass[11pt]{article}

\usepackage[final]{acl}

\usepackage{times}
\usepackage{latexsym}
\usepackage{subcaption}
\usepackage{threeparttable}
\usepackage{multirow}
\usepackage[T1]{fontenc}
\usepackage[utf8]{inputenc}

\usepackage{microtype}

\usepackage{inconsolata}

\usepackage{graphicx}

\usepackage{subcaption}
\usepackage{booktabs}
\usepackage{amsmath}
\usepackage{amsfonts} 
\usepackage{enumitem}
\usepackage{float}
\usepackage{soul}

\usepackage[capitalize,noabbrev]{cleveref}
\usepackage[font=small,labelfont=bf]{caption}
\usepackage[subtle]{savetrees}

\newcommand{\tworow}[1]{
  \begin{tabular}{@{}c@{}}#1\end{tabular}
}

\title{Semantic Agreement Enables
Efficient Open-Ended LLM Cascades}
\author{
 \textbf{Duncan Soiffer},
 \textbf{ Steven Kolawole},
 \textbf{ Virginia Smith}
\\
 Carnegie Mellon University
\\
 \small{
   \textbf{Correspondence:} \href{mailto:dsoiffer@cs.cmu.edu}{dsoiffer@cs.cmu.edu}, \href{skolawol@cs.cmu.edu}{skolawol@cs.cmu.edu}
 }
}

\begin{document}
\maketitle

\begin{abstract}
% \vspace{-0.25em}
% Cascade systems for open-ended text generation face a fundamental challenge: determining output reliability when generation quality exists on a continuous spectrum with multiple valid outputs. We introduce semantic agreement between multiple model outputs as a training-free deferral signal that captures meaning-level consensus rather than token-level confidence.
% % Unlike confidence-based methods optimized for next-token prediction, s
% Semantic cascades maintain black-box APIs compatibility, operate without training, and preserve performance across model updates. When ensemble models converge semantically despite surface variations, this consensus indicates output reliability more accurately than next-token confidence metrics.
% Across tasks with models from 1B to 70B parameters, semantic methods achieve superior deferral decisons and competitive quality at 40\% computational cost with up to 60\% latency reductions.
% % Adding weaker models often improves performance.
% Semantic agreement provides a practical framework for efficient LLM deployment that delivers substantial cost and latency improvements.
% % while aligning with production constraints

% \gs{Can you add a short motivating sentence first to explain what cascade systems are/why they are useful?}
Cascade systems route computational requests to smaller models when possible and defer to larger models only when necessary, offering a promising approach to balance cost and quality in LLM deployment.
However, they face a fundamental challenge in open-ended text generation: determining output reliability when generation quality lies on a continuous spectrum, often with multiple valid responses. 
To address this, we propose \textit{semantic agreement}---meaning-level consensus between ensemble outputs---as a training-free signal for reliable deferral.
We show that when diverse model outputs agree semantically, their consensus is a stronger reliability signal than token-level confidence.
Evaluated from 500M to 70B-parameter models, we find that semantic cascades match or surpass target-model quality at 40\% of the cost and reduce latency by up to 60\%.
Our method requires no model internals, works across black-box APIs, and remains robust to model updates, making it a practical baseline for real-world LLM deployment.
\end{abstract}
% \vspace{-1em}
\section{Introduction}
%\vspace{-0.5em}
Large language models (LLMs) have enabled impressive progress across a range of language tasks; however, this progress comes at a steep computational cost. Larger models typically produce higher-quality outputs but are slower, more expensive, and less scalable for real-time or large-scale use. To mitigate this cost-quality tradeoff, cascade systems have emerged as a practical deployment strategy: route inputs to smaller models whenever possible, and defer to larger models only when necessary ~\citep{chenfrugalgpt, yue2024large, kolawole2024abc}.
\begin{figure}
    \centering
    \includegraphics[width=1\linewidth]{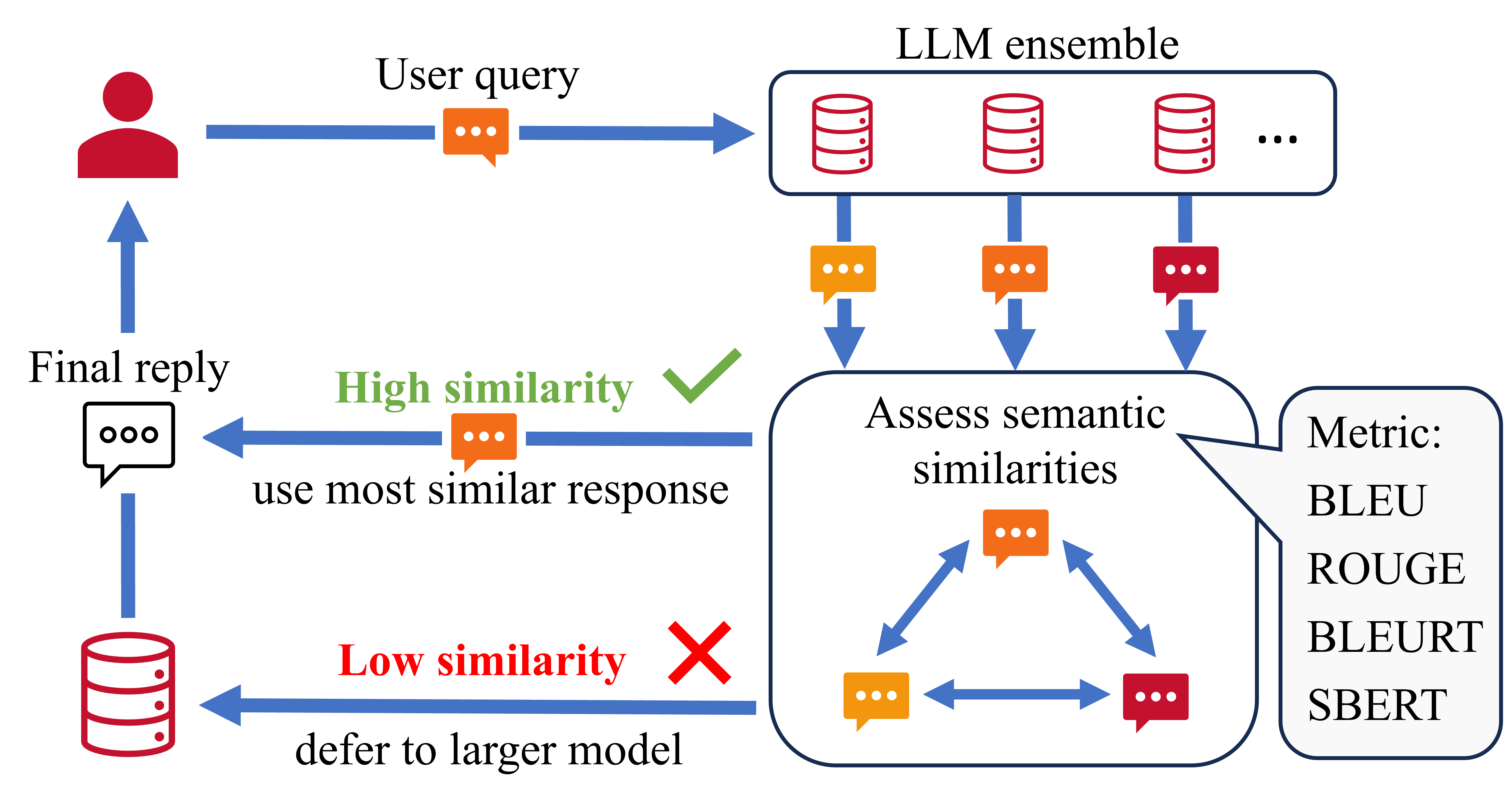}
    \caption{\textbf{Overview of the semantic cascade framework.} An ensemble of small models generates responses to a user query. Semantic similarity metrics assess agreement among outputs. High semantic agreement indicates reliability: the system returns the highest-scoring ensemble response. Low agreement signals uncertainty: the query defers to a larger target model, ensuring that expensive model is called only when necessary.}
    \label{fig:figure_1}
\end{figure}
While cascades are well-established in classification and other tasks with objectively assessable outputs, they remain underexplored---and fundamentally more challenging---in open-ended settings. In open-ended generation, output quality lies on a continuous spectrum and multiple valid outputs may exist, complicating deferral decisions. Determining when to defer requires estimating quality without the ground truth references available to traditional cascade systems: the conventional ``correct" vs. ``incorrect" paradigm no longer applies. Moreover, existing LLM cascading approaches rely on learned routing mechanisms requiring substantial training investment through domain-specific fine-tuning or model-specific engineering~\citep{chenfrugalgpt, ong2024routellm, aggarwal2024automix}. These systems must be retrained whenever models or distributions shift---a scenario that is common in today's fast-evolving LLM landscape---leading to recurring costs and reduced production agility.

Modern production deployments compound these challenges. Dominant commercial models such as GPT-4, Claude, and Gemini are accessible only through black-box APIs, exposing no internal representations required by existing cascade methods. Frequent model updates render trained routers obsolete, while heterogeneity across model families limits the portability of learned strategies. Enterprise deployments demand adaptive methods that generalize across architectures and deliver operational efficiency gains that translate directly to cost savings.

% \textbf{Research Gap \& Our Solution}. We bridge the gap between theoretical cascade advances and deployment realities by proposing semantic agreement between multiple model outputs as a training-free deferral signal. When ensemble models produce semantically consistent responses, their collective output likely represents reliable generation. Conversely, semantic disagreement suggests uncertainty, indicating deferral would be beneficial. This approach captures a dimension of output reliability that token-level confidence methods completely miss, while eliminating training requirements and maintaining compatibility across black-box APIs. 

To address these challenges, we introduce a simple, training-free alternative: using \textit{semantic agreement} among multiple model outputs as a deferral signal (overview in Figure~\ref{fig:figure_1}). When independently generated outputs are semantically consistent---even if lexically different---their agreement suggests the underlying meaning is reliable. 
% models produce semantically consistent responses, their collective output represents reliable generation. Conversely, semantic disagreement suggests uncertainty, indicating deferral would be beneficial. 
In contrast, semantic divergence signals uncertainty, suggesting that deferring to a larger model may be warranted.
Our method requires no model-specific tuning, no access to internals, and generalizes across model families and API versions---while capturing aspects of output reliability that token-level confidence methods miss entirely.

We evaluate semantic agreement across translation, summarisation, question answering, and reading comprehension using models from 500M to 70B parameters. Our method achieves competitive or superior quality to target large models at 40\% of large-model computational budgets and reduces latency by up to 60\%, while often delivering superior deferral decisions than token-level confidence methods. These results position semantic agreement as a strong, practical baseline framework for building efficient, adaptive LLM  systems in production.
\section{Related Work}
%\vspace{-0.5em}

\paragraph{Cascading in Language Models} Recent cascade systems for LLMs demonstrate effectiveness across tasks, but face fundamental limitations for open-ended generation~\citep{chenfrugalgpt, ong2024routellm, aggarwal2024automix}. These approaches require extensive training that can become obsolete with model/test data distribution updates, and prior works focus primarily on classification or objectively assessable tasks. When underlying models are updated---common in today's rapidly evolving LLM landscape---routing systems incur the recurring costs of complete retraining, though recent work~\citep{kolawole2024abc, feng2024graphrouter, jitkrittum2025universal} explores circumventing this bottleneck.

%\vspace{-0.4em}
\paragraph{Confidence-Based Deferral for Generation} Extending cascades to open-ended generation exposes fundamental limitations of confidence-based approaches. \citet{tokenleveluncertainty} explores token-level uncertainty for selective generation while \citet{narasimhan2024faster} extends this with speculative decoding. However, token-level confidence signals are designed to optimize for next-token prediction rather than semantic coherence or factual accuracy---qualities that ultimately determine generation utility. More critically, confidence-based approaches require access to model internals, precluding deployment with proprietary systems that dominate enterprise usage. This architectural constraint represents a fundamental barrier when  production deployments rely on black-box APIs rather than locally hosted models with accessible internal states.

%\vspace{-0.4em}
\paragraph{Ensemble Methods and Agreement-based Signals} Traditional ensembles focus on improving quality through combination rather than addressing deferral decisions \citep{llm_ensemble_survey, lakshminarayanan2017simple, wangwisdom,wangfusing, jiang2023llm}. ABC \citep{kolawole2024abc} uses ensemble agreement for cascade deferral while \citet{yue2024large} explores consistency-based quality estimation for numerical reasoning tasks, but both are limited to fixed output spaces.
Other works, like ModelSwitch~\citep{chen2025we} leverage agreement across multiple models to switch dynamically during repeated sampling, demonstrating that consistency correlates with accuracy.
Our work, instead, addresses the distinct challenge of cascade deferral decisions in open-ended generation where valid outputs exist without objective correctness measures, and ``correctness" must be defined through semantic meaning rather than exact matching.

%\vspace{-0.4em}
\paragraph{Semantic Similarity in NLP}
Finally, we note that existing semantic similarity metrics primarily serve as evaluation tools rather than active system components. Classic approaches like BLEU~\citep{papineni2002bleu} and ROUGE~\citep{lin2004rouge} measure surface-level lexical overlap but miss deeper semantic relationships. Recent methods using contextual embeddings—BLEURT~\citep{sellam2020bleurt}, BERTScore \citep{zhangbertscore}—show stronger correlation with human judgements, but their application remains confined to post-hoc assessment. In this work, we consider a paradigm shift from evaluation to active decision-making, treating semantic similarity as an architectural component for resource allocation in generative cascade systems.

\section{Methods}
%\vspace{-0.6em}

% \gs{Should add a brief intro here to explain what you will discuss in this section. It would also be good to remind the reader what a ``cascade'' is and why this can be useful. Then, what are the components of a cascade? (with the main one being the deferral protocol)}
We present a semantic cascade framework comprising three key components. First, we define our deferral protocol (\S\ref{sec:method-defer-protocol}), which determines when to route queries from an ensemble of small models to a larger target model. Second, we introduce semantic agreement signals (\S\ref{sec:semantic-signals}) that assess meaning-level consensus between ensemble outputs without requiring ground-truth references. Finally, we establish token-level confidence baselines for comparison (\S\ref{sec:token-level}).

\subsection{Deferral Protocol} \label{sec:method-defer-protocol}
A semantic cascade comprises $n$ lightweight ensemble models, $M_1, \dots, M_n$, and a larger, high-capacity target model $M_T$. Given an input $x$, each ensemble model produces a full response $y^{(i)} = M_i(x)$. A deferral score function $s\left(y^{(i)},~ \dots,~ y^{(n)}\right)$ then determines whether to defer to $M_T$, based on semantic agreement across outputs.  We adopt the convention that higher scores reflect greater certainty; the system retains ensemble predictions for high scores and defers for low scores. When not deferring,
 the system selects a final output $y = y^{(i)}$ corresponding to the response with highest score according to an output scoring function $o\left(y^{(i)},~ y^{(1)}, ~\dots, ~y^{(n)}\right)$.
% We formalize the deferral process as follows. A cascade consists of $n$ (small) ensemble models $M_1, \dots, M_n$, and a (large) target model $M_T$. Given input $x$, each ensemble model generates output $y_i=M_i(x)$. Then, based on deferral score $s(y_1,\dots, y_n)$, the system chooses whether to defer to $M_T$. When not deferring, it selects $y=y_i$ for the $y_i$ with the highest output score $o_i(y_i, y_1,\dots,y_n)$.

% We adopt the convention that higher scores reflect greater certainty. Accordingly, the system defers to the target model for inputs with lower deferral scores $s$, while retaining predictions with higher $s$.
%Concretely, w
% We adopt the convention that larger scores indicate higher certainty, so outputs with low deferral score $s$ are deferred before outputs with high $s$.
% , and we set $y=\argmax_{y_i} o(y_i, y_1,\dots,y_n)$ when not deferring.
% This training-free framework allows cascades to remain adaptable as models evolve without requiring expensive retraining of routing mechanisms, i.e., multiple signals that are used as-is.
% Note that this system can be extended to cascades with arbitrarily many layers of models, but for simplicity we focus on the two-layer case in this paper.

% We evaluate multiple signals for $s(y_1,\dots,y_n)$ and $o_i(y_i, y_1,\dots,y_n)$, all of which require no additional training or access to ground-truth labels.

\subsection{Semantic Agreement Signals} \label{sec:semantic-signals}
%\vspace{-0.5em}

Semantic similarity metrics offer a natural approach to assess agreement between multiple model outputs without %requiring 
ground-truth references. We explore increasingly sophisticated metrics capturing different aspects of similarity, from surface-level lexical overlap to deeper semantic representation. This progression allows systematic examination of how different dimensions of semantic agreement affect cascade performance.

%\vspace{-0.4em}
\paragraph{Classic Overlap Metrics}
We begin with reference-based metrics: BLEU measures n-gram precision, ROUGE-N measures n-gram overlap, and ROUGE-L measures longest common subsequence overlap---all producing scores between 0 and 1, where higher scores indicate better matches. While limited to surface-level similarity, these metrics provide computational efficiency and interpretability. 
% \gs{can you provide more detailed information here about these metrics? eg what do they range between? maybe add a citation?} %In the ensemble setting, we compute pairwise overlap across generations $y_1^{(i)}, y_1^{(j)}$ to assess consistency.

%\vspace{-0.4em}
\paragraph{Pretrained Metrics}
BLEURT is a regression model based off BERT, fine-tuned to reflect human judgments of generation quality. It requires text pairs as input and outputs a similarity score. SBERT \citep{reimers2019sentence} provides a reference-free approach by embedding each output $y^{(i)}$ into a vector representation $z^{(i)}$ and computing pairwise cosine similarities in the embedding space.

%\vspace{-0.4em}
\paragraph{Implementation}
In an ensemble setting, we use these similarity metrics as the output scoring function $o$ by computing the mean pairwise similarity between $y^{(i)}$ and all other $y^{(j)}$. We then use the output scores to determine the deferral score by $s=\max_i o\left(y^{(i)},~ y^{(1)},~ \dots,~ y^{(n)}\right)$. This approach identifies the output with highest agreement across the ensemble, while providing a confidence signal for deferral decisions.

\subsection{Token-Level Confidence Baselines}\label{sec:token-level}

For comparison, we evaluate token-level confidence metrics that extend classification confidence to generation \citep{tokenleveluncertainty}:
\textbf{Chow-Sum} (sum of token log probabilities across output sequence),
\textbf{Chow-Avg} (sum of token log probabilities normalized by sequence length), and
\textbf{Chow-Quantile} ($q$-th quantile of token log probabilities). We analyze Chow-Quantile for quantiles $q\in\{0.0, 0.1, \dots, 1.0\}$, which provides a balance between expressivity and avoiding spuriously high ``Best Quantile'' results due to noise.
% \vspace{-0.6em}
% \begin{itemize}[leftmargin=*]
% \itemsep0em
% \item \textbf{Chow-Sum}: Sum of token log probabilities across output sequence.
% \item \textbf{Chow-Avg}: Sum of token log probabilities normalized by sequence length. %Per-token average confidence normalized by sequence length.
% \item \textbf{Chow-Quantile($q$)}: $q$-th quantile of token log probabilities.% identifying uncertain portions.
% \end{itemize}
% Chow-Sum (sum of token log probabilitites), Chow-Avg (length-normalized sum, and Chow-Quantile($q$) (q-th quantile of token log probs).

%\vspace{-0.3em}
\paragraph{Handling Model Heterogeneity}
% \vspace{-0.6em}
presents significant challenges for token-level approaches. Ensembling heterogeneous models based on confidence scores fails due to differences in training dynamics, architectures, and vocabularies. Baseline token probabilities vary significantly across models and reflect different calibration levels, causing systematic dominance by particular models when using raw confidence scores (see \cref{apdx-token-level-ensemble} for empirical evidence). 

 We address this calibration challenge without an expensive post-training fix through z-score normalization: we run each model on a subset of training data, compute confidence metric statistics, then normalize during inference. We use mean confidence as the deferral score, selecting the output from the model with the highest normalized confidence when not deferring.

Note that unlike semantic agreement, which inherently requires multiple model outputs for comparison, token-level confidence operates on individual model responses and does not require ensembling. However, to provide a comprehensive comparison, we evaluate both individual token-level cascades (using single models) and token-level ensemble cascades that aggregate confidence scores across multiple models.

\subsection{Practical Deployment Advantages}
Apart from out-of-the-box applicability to open-generation cascading, our semantic cascade framework addresses critical production constraints that existing methods overlook. 

\textbf{Training-free operation} eliminates the substantial overhead of learned routing mechanisms. Unlike approaches requiring domain-specific fine-tuning or model-specific engineering, semantic agreement operates directly on model outputs without additional training data or the need for labels, reducing deployment time and removing distribution shift risks between training and production environments. As a consequence, when underlying models are updated---a common occurrence in production API environments---semantic cascades adapt automatically to new model capabilities.

\textbf{Black-box compatibility} enables deployment across diverse model providers. Semantic methods function with any text generation API by operating on outputs rather than internal model states, extending to proprietary systems where confidence scores are unavailable.

% \textbf{Update resilience} maintains performance across model versions without retraining. When underlying models are updated (common in production API environments), semantic cascades adapt automatically to new model capabilities, while token-level approaches may require signal calibration for each model version.

\textbf{Robustness to heterogeneity} emerges as an important property. Semantic cascades demonstrate a unique resilience to varying model quality within ensembles,
%semantic ensembles extract meaningful signals from (dis)agreement between models of different quality, [commented out because I put elsewhere]
%Adding lower-performing models often preserves or improves cascade performance,
%unlike token-level ensembles where performance degrades toward the weakest component. This %counterintuitive
%property enables flexible ensemble composition and graceful degradation under model availability constraints.
enabling flexible ensemble composition and graceful degradation under model availability constraints.

%\vspace{-0.6em}

\section{Experimental Setup}
% \vspace{-0.6em}

%\vspace{-0.6em}
\paragraph{Models and Scaling Design}
We evaluate semantic cascades across model families representing diverse architectural approaches and parameter scales. Our ensemble models span the FLAN-T5~\citep{flan-t5}, mT0 \citep{mt0}, Gemma \citep{gemma3}, Llama \cite{llama3}, and Qwen \citep{qwen2.5} families, selected for their multilingual capabilities and availability at parameter tiers of roughly 1B, 3B, and 8B, while we use Llama3.1-70B as our primary target model. %, enabling analysis of cascade behavior across realistic production model gaps.
For consistency, all outputs are produced via greedy decoding.

To systematically evaluate strategies for combining models across tiers, we test ensemble configurations ranging from homogeneous 1B model groups to heterogeneous combinations mixing 8B, 3B, and 1B models. 
%To systematically evaluate scaling effects, we test ensemble configurations ranging from homogenous 1B model groups to heterogeneous combinations mixing 8B, 3B, and 1B models. This design allows investigation of our core hypotheses: that larger model gaps result in better cost savings, that stronger base models enable more intelligent deferral decisions, and that off-the-shelf models can be randomly plugged into our cascade system. 

%\vspace{-0.6em}
\paragraph{Task Selection}
We evaluate across translation (WMT19 DE$\to$FR~\citep{barrault2019findings}, WMT14 FR$\to$EN/EN$\to$FR~\citep{bojar2014findings}), sumarization (CNN/DailyMail~\citep{nallapati2016abstractive}, XLSum~\citep{hasan2021xl}), open-book question answering/reading comprehension (SQuAD1.1~\citep{rajpurkar2016squad}), and closed-book question answering (TriviaQA~\citep{joshi2017triviaqa}). This selection spans extractive tasks with single correct answers to abstractive generation with many valid outputs, enabling identification of task characteristics that predict semantic cascade effectiveness.
% \gs{add links to these datasets} %Translation and summarization represent high-volume production workloads where efficiency gains translate directly to operational cost savings.

%\vspace{-0.6em}
\paragraph{Evaluation Protocol}
For cost-efficiency analysis, we examine performance at fixed computational budgets (40\% of target model FLOPs) and at the latency required to achieve target quality thresholds (98\% of target model performance). 
We model parallel execution---ensemble models on separate GPUs or API endpoints---while tracking total computational cost through the sum of FLOPs. This provides realistic efficiency assessments for production environments where ensemble models execute concurrently.

We additionally assess cascade performance using \emph{deferral curves}: plots of deferral rate against output quality. As a scalar summary statistic, we report the area under the deferral curve (AUC-DF), with higher values indicating better overall performance---though the range of AUC-DF values varies across datasets. This evaluation approach, established in prior cascade literature \cite{tokenleveluncertainty}, is useful for assessing the strength of the deferral signal conferred by different cascades, as it isolates deferral decisions and their impact on performance from other factors.

\begin{table*}[htbp]
\centering
\caption{\textbf{Semantic cascades achieve superior quality at constrained computational budgets.} At 40\% of target model (Llama3.1-70B) budget (FLOPs), semantic methods match or exceed 70B performance on most tasks. Latency measurements show substantial reductions (60\% on SQuAD,  39\% on CNN/DM) at 98\% of the target model's quality, demonstrating that better deferral decisions and output selection translate to both computational and time efficiency gains over significantly larger individual models and the best token-level cascades. To capture performance across the whole curve and avoid noise induced by selecting based off only a single point, `best' cascades are determined by AUC-DF. %, which avoids potential bias induced by measuring at only one point.
Full results are presented in \cref{apdx:experimental_results}.}
\label{tab:efficiency_results}
\small
% \resizebox{\linewidth}{!}{
\begin{tabular}{@{}lcccc@{}}
\toprule
\textbf{Task} & \textbf{Metric} & \textbf{Best Semantic Ensemble} & \textbf{Best Token-Level} & \textbf{Large Model (70B)} \\
\midrule
\multirow{2}{*}{\textbf{SQuAD}} 
& Performance at 40\% Budget & \textbf{0.843} & 0.823 & 0.827 \\
& Latency at 98\% Quality (ms) & \textbf{163} & 221 & 410 \\
\midrule
\multirow{2}{*}{\textbf{CNN/DM}} 
& Performance at 40\% Budget & \textbf{0.261} & 0.259 & 0.267 \\
& Latency at 98\% Quality (ms) & \textbf{4,041} & 5,123 & 6,632 \\
\midrule
\multirow{2}{*}{\textbf{WMT FR$\to$EN}} 
& Performance at 40\% Budget & \textbf{0.747} & 0.744 & 0.747 \\
& Latency at 99.5\%* Quality (ms) & \textbf{516} & 602 & 1,319 \\
\midrule
\multirow{2}{*}{\textbf{TriviaQA}} 
& Performance at 40\% Budget & 0.750 & \textbf{0.776} & 0.807 \\
& Latency at 98\% Quality (ms) & 315 & \textbf{311} & 368 \\ 
\bottomrule
\end{tabular}
% }
\begin{tablenotes}
\small
\item Best token-level base model: SQuAD/CNNDM: Qwen2.5-7B; WMT/Trivia-QA: Llama3.1-8B.
\item Ensemble models: SQuAD: [Qwen2.5-7B, mT0-Large, FLAN-T5-Large]; CNNDM: [Qwen2.5-7B, mT0-Large, Llama3.2-3B]; WMT: [Qwen2.5-7B, Llama3.1-8B, FLAN-T5-Large]; TriviaQA: [Llama3.1-8B, Qwen2.5-7B, Llama3.2-3B].
\item *Several models' baseline performance start above 98\% of the target model's, so we adjust for this tight clustering
\end{tablenotes}
\end{table*}

\begin{figure*}
\centering
    \begin{subfigure}{0.2425\linewidth}
    \includegraphics[width=\linewidth]{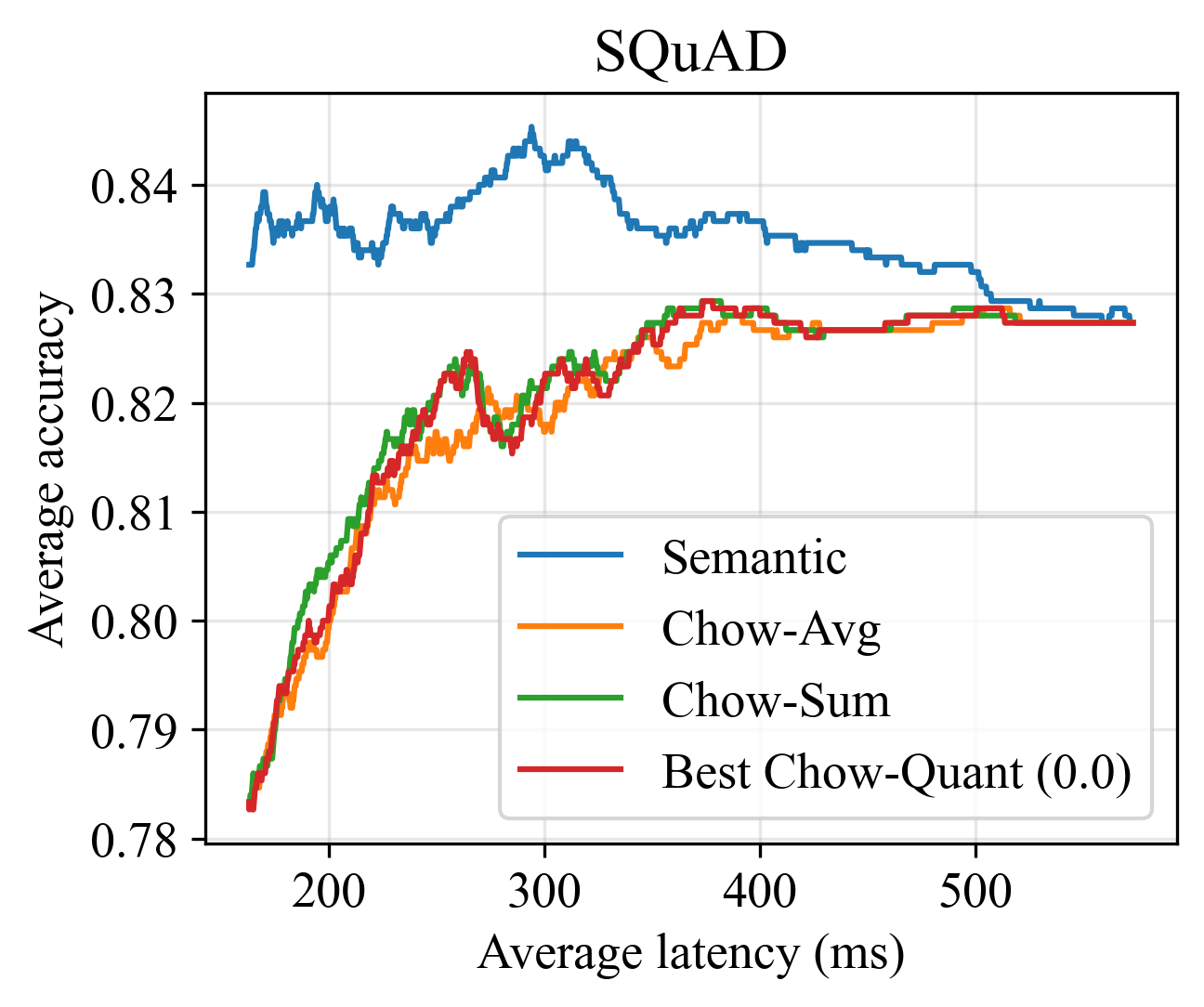}
    \end{subfigure}
    \hfill
    \begin{subfigure}{0.2475\linewidth}
    \includegraphics[width=\linewidth]{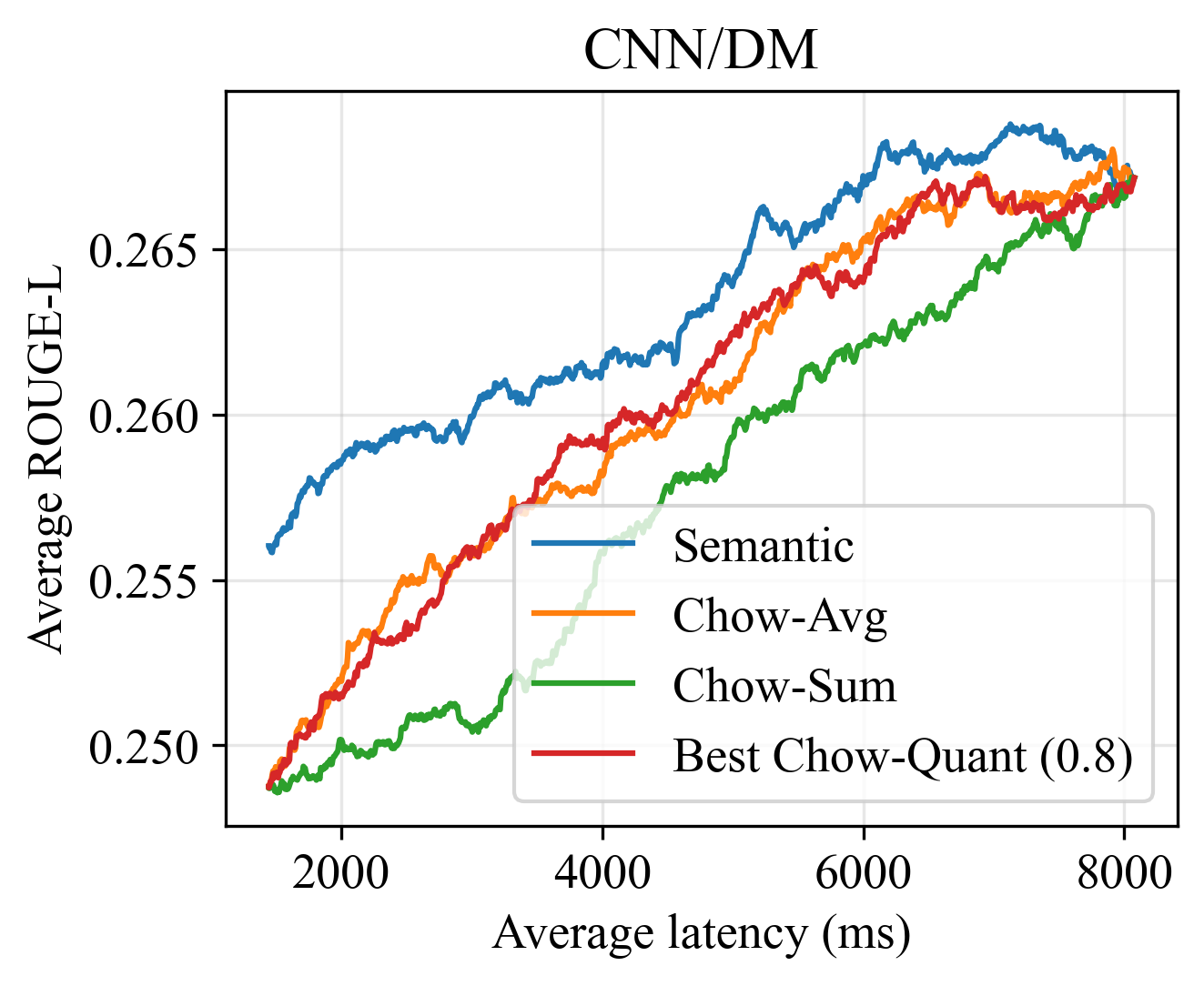}
    \end{subfigure}
    \hfill
    \begin{subfigure}{0.248\linewidth}
    \includegraphics[width=\linewidth]{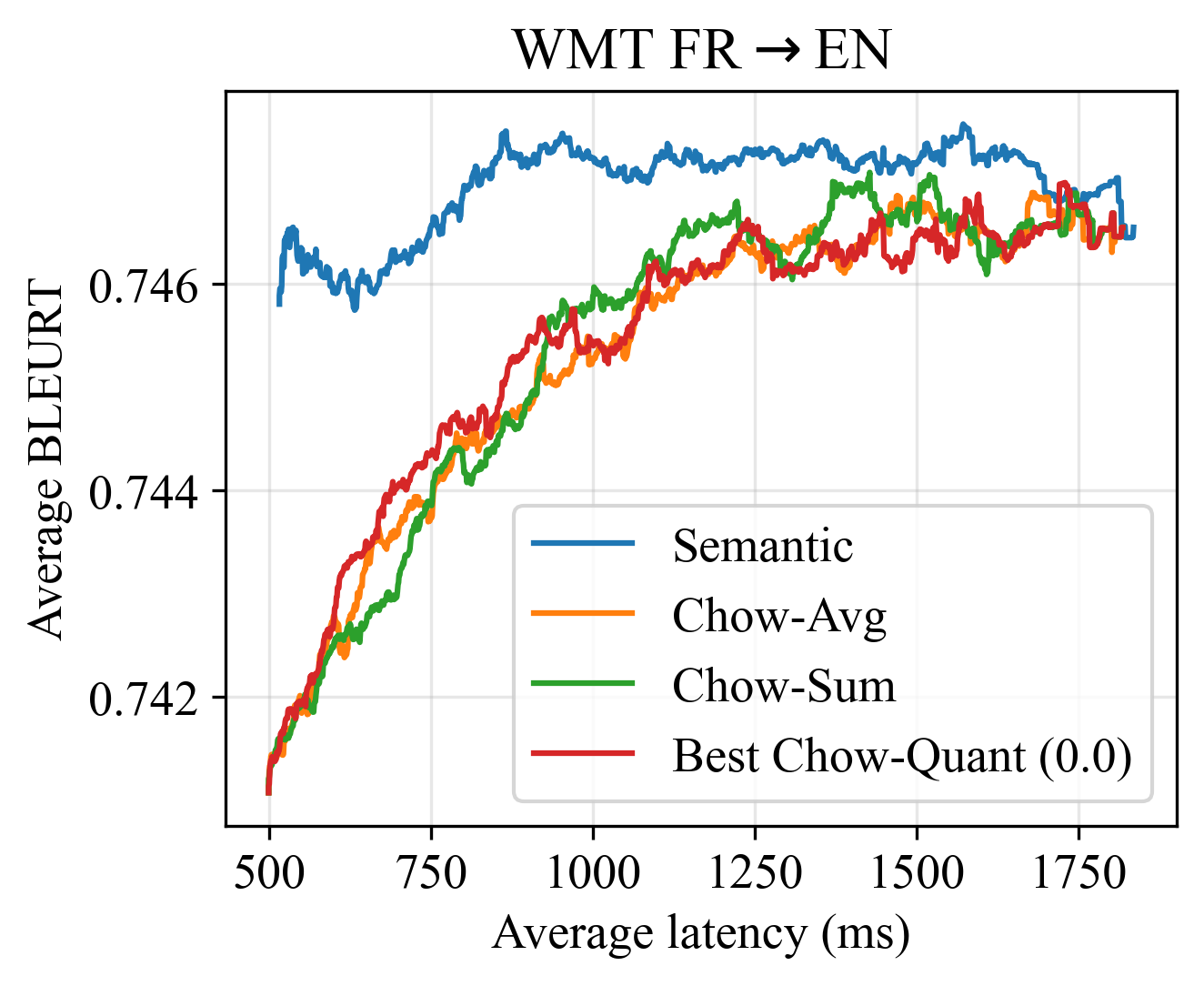}
    \end{subfigure}
    \hfill
    \begin{subfigure}{0.249\linewidth}
    \includegraphics[width=\linewidth]{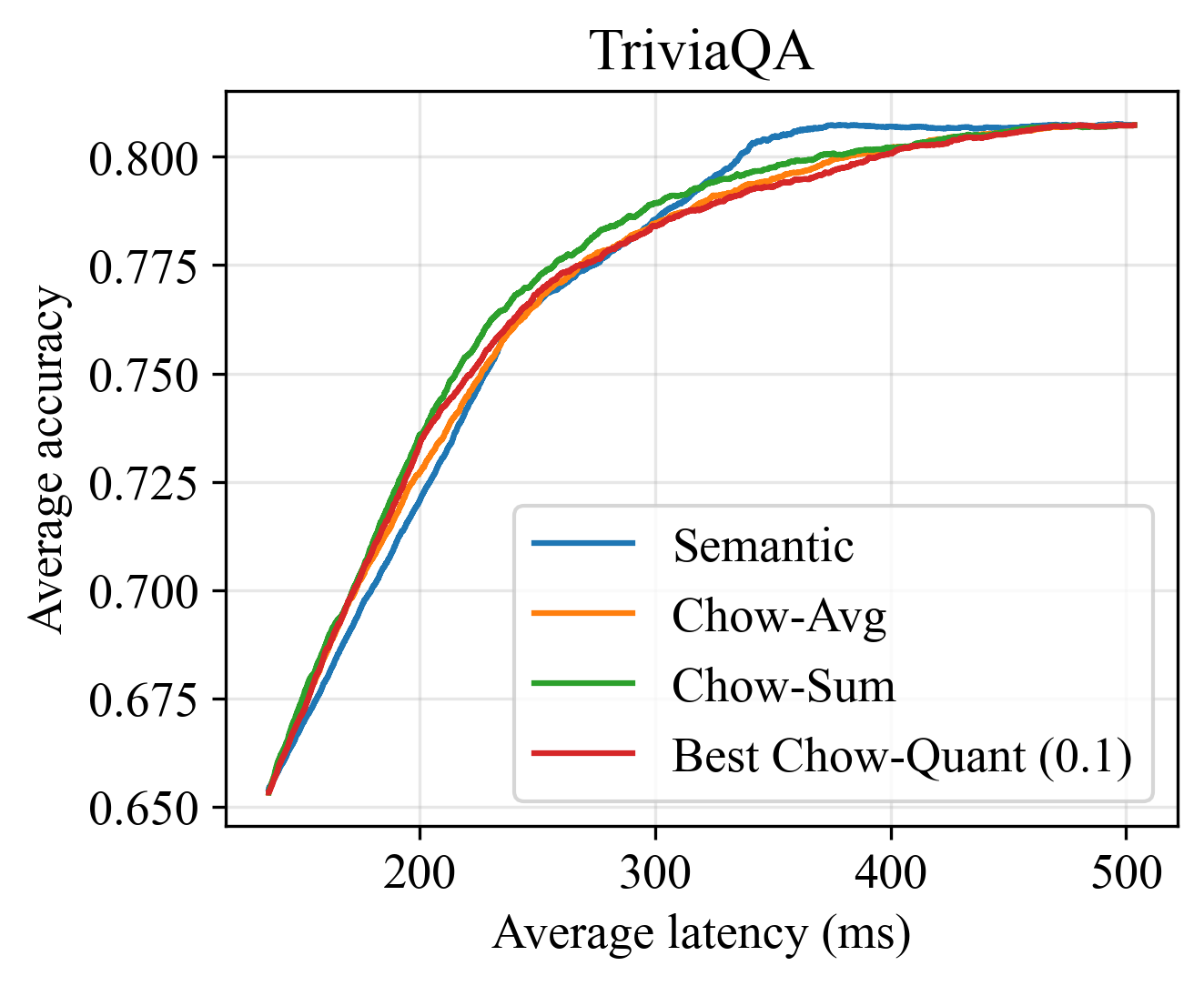}
    \end{subfigure}

    \caption{\textbf{Semantic deferrals achieve superior efficiency-quality tradeoffs across diverse generation tasks.} On SQuAD, CNN/DM, and WMT FR$\to$EN, semantic methods consistently outperform token-level confidence across all quality and latency levels and even surpass the target model in some cases. These results suggest semantic agreement captures output reliability dimensions that token-level confidence misses. The curves shown correspond to the cascades from \cref{tab:efficiency_results}.}    
    \label{fig:latency_curve}
\end{figure*}

\section{Results \& Analysis}

\subsection{Cost-Efficiency Analysis}
Superior deferral decisions and response selection translates directly to efficiency gains across production-relevant scenarios. \cref{tab:efficiency_results} demonstrates that semantic cascades achieve competitive quality at 40\% of target model computational budget: SQuAD (84.3\% vs 82.7\% target accuracy), CNN/DailyMail (.261 vs .267 target ROUGE-L), and WMT FR$\to$EN (.747 vs .747 target BLEURT). 
% \gs{this table is too far away from the text. also, I would present this table differently: I think it would make more sense as two separate figures, side-by-side: one that shows that quality is preserved (eg a bar chart where the numbers are all the same, roughly), and another that shows computational budget is reduced (so a bar chart where the budget is small for semantic cascades}

Latency advantages prove substantial when targeting 98\% of large model quality. Compared to the target model, semantic cascades achieve 60\% latency reduction on SQuAD, 39\% reduction on CNN/DailyMail, and 61\% reduction on WMT FR$\to$EN, resulting in cascade systems which are 14-26\% faster than their token-level counterparts at the same (or higher) target quality.

The efficiency gains translate directly to operational cost savings, including in API-based deployments. Achieving near-equivalent quality at 40\% computational cost enables significant cost reduction for token-based pricing models, while latency improvements enhance user experience through faster response times. 

To address concerns about reference-based evaluation in open-ended generation, we conducted additional experiments using reference-free metrics: COMETKiwi-XL \citep{comet-kiwi} a reference-free regression model for assessing translation quality, and G-Eval \citep{liu2023gevalnlgevaluationusing}, a framework for using an LLM (here, GPT-4-mini) to assess summarization with a consistent set of criteria. Results confirm our findings: semantic cascades maintain their advantage over token-level methods at 40\% computational budget while preserving latency benefits (CNN/DM: 3,937ms vs token-level 5,014ms; WMT FR$\to$EN: 522ms vs token-level 621ms at near-equivalent quality thresholds). The full table is presented in \cref{apdx:ref-free}.

\begin{figure*}
    \centering
    \begin{subfigure}{0.39\linewidth}
    \includegraphics[width=\linewidth]{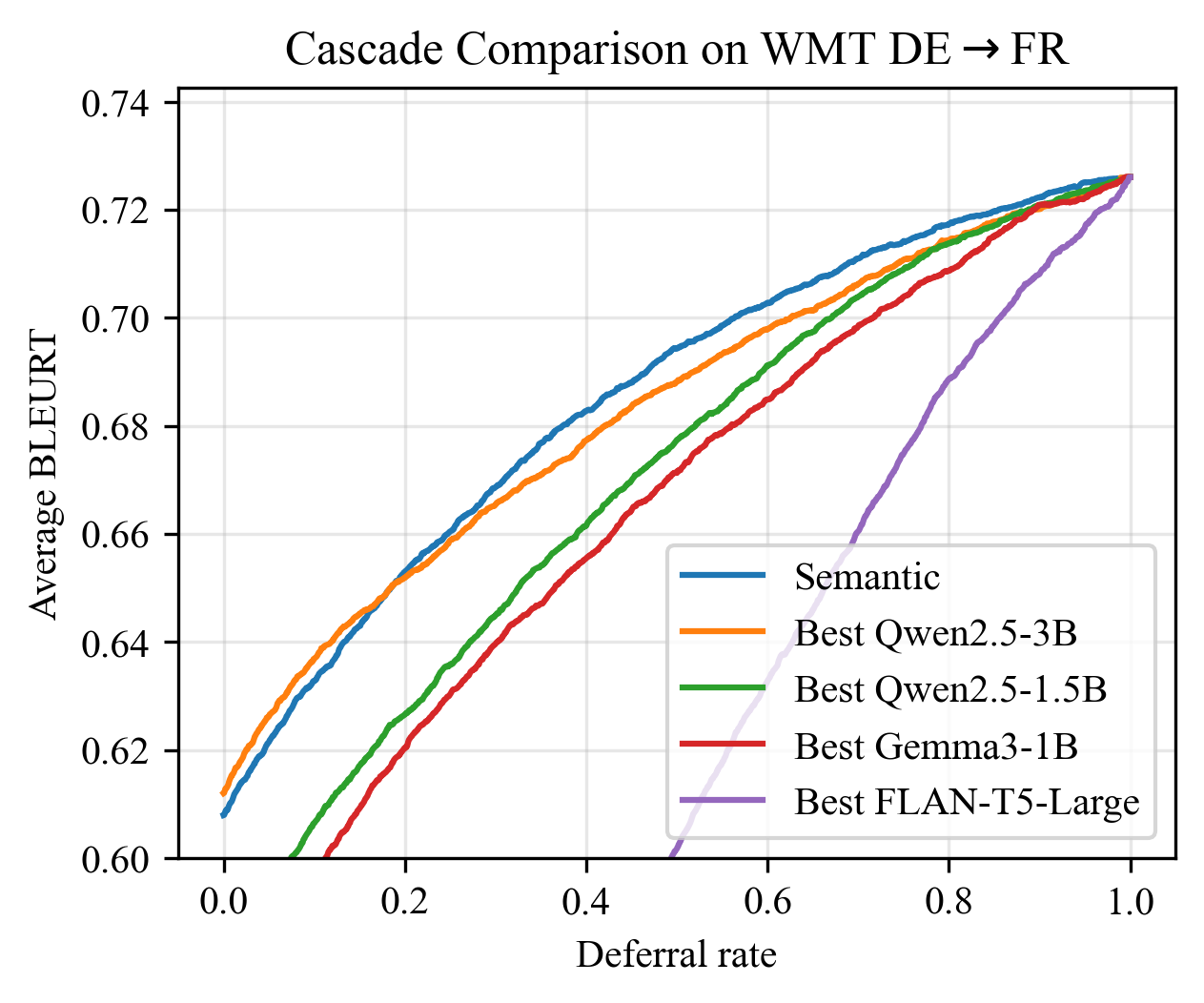}
    \end{subfigure}
    % \hfill
    \hspace{0.04\linewidth}
    \begin{subfigure}{0.39\linewidth}
    \includegraphics[width=\linewidth]{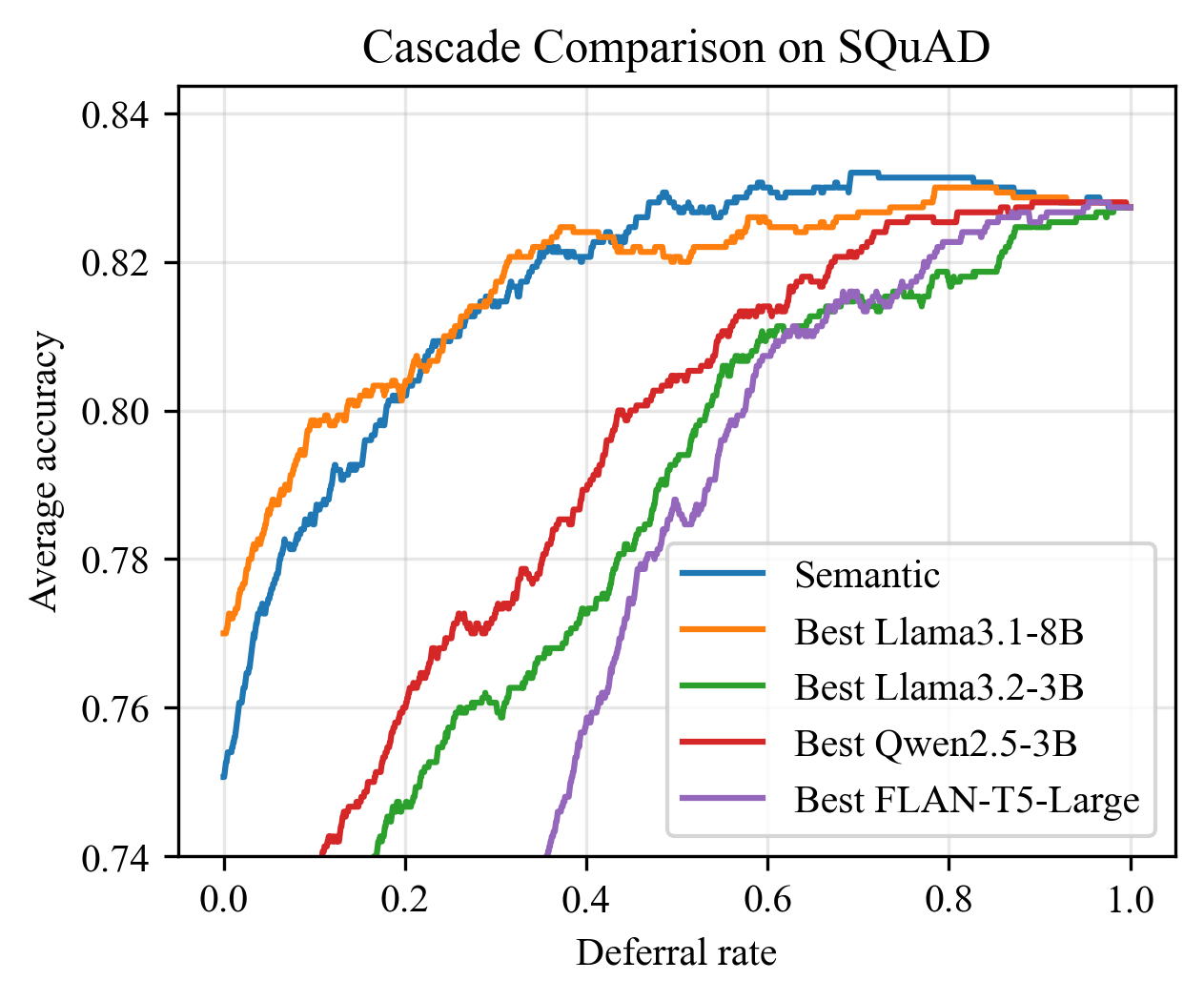}
    \end{subfigure}

    \caption{\textbf{Semantic cascades select stronger outputs than their constituent models and perform more effective deferral decisions than larger token-level cascades.} \textbf{(a)} Deferral curves for a semantic cascade of [Qwen2.5-1.5B, Gemma3-1B, FLAN-T5-Large], token-level cascades of its individual ensemble models, and a token-level cascade of Qwen2.5-3B, evaluated on WMT DE$\to$FR. \textbf{(b)} The same framework for a semantic cascade of [Llama3.2-3B, Qwen2.5-3B, FLAN-T5-Large] and larger token-level model Llama3.1-8B, evaluated on SQuAD.
    All cascades defer to Llama3.1-70B; for each cascade only the curve with highest AUC is shown. In both cases, the semantic ensemble has a lower baseline than the larger token-level model, but overtakes it due to superior deferral decisions. Additionally, the semantic ensemble considerably outperforms its constituent models across all deferral rates. 
    This demonstrates that semantic similarity's advantage comes not just from reliably selecting strong responses, but also from genuinely superior deferral decisions.
   }
    \label{fig:better-defer}
\end{figure*}

%highlight: trivia QA token-level performance drops precipitously due to heterogenous nature
% THIS WILL BECOME THE AUC TABLE
\begin{table*}[htbp]
\centering
\caption{\textbf{Semantic agreement is a stronger deferral signal.} AUC-DF values demonstrate that semantic deferral achieves near-dominance over single-model token-level cascades as measured by the quality of the deferral signal. Meanwhile, token-level ensembles require equally or more expensive models, yet yield smaller gains and can fail to outperform the best individual token-level cascade. Random represents the expected AUC-DF when deferring at random from the best base model.}
\label{tab:auc_main}
\small
% \resizebox{\linewidth}{!}{
\begin{tabular}{@{}lcccc@{}}
\toprule
\textbf{Task} & \textbf{Best Semantic Ensemble} & \textbf{Best Token-Level} & \textbf{Best Token-Level Ensemble} & \textbf{Random} \\
\midrule
{\textbf{SQuAD}} 
& \textbf{.8353} & .8217 & .8255 & .8053 \\
\midrule
{\textbf{CNN/DM}} 
& \textbf{.2635} & .2607 & .2611 & .2580 \\
\midrule
{\textbf{WMT FR$\to$EN}} 
& \textbf{.7470} & .7454 & .7453 & .7438 \\
\midrule
{\textbf{TriviaQA}} 
& .7720 & \textbf{.7741} & .7630 & .7303 \\
\bottomrule
\end{tabular}
% }
\begin{tablenotes}
\small
\item Semantic ensemble and individual token-level models are the same as in \cref{tab:efficiency_results}
\item Token-level ensemble: SQuAD: [Llama3.1-8B, Qwen2.5-7B, mT0-Large]; Others: [Llama3.1-8B, Qwen2.5-7B]
% SQuAD: [Qwen2.5-7B, mT0-Large, FLAN-T5-Large]; CNNDM: [Qwen2.5-7B, mT0-Large, Llama3.2-3B]; \\WMT: [Qwen2.5-7B, Llama3.1-8B, Gemma-1B]; TriviaQA: [Llama3.1-8B, Llama3.2-3B, Qwen2.5-3B].
% \item Token-level ensemble models SQUAD Qwen2.5-7B, Llama3.1-8B, mT0-Large 

\end{tablenotes}
\end{table*}

\subsection{Deferral Signals and Hetereogeneity}
\label{sec:heterogeneity}
% Semantic agreement captures output reliability through meaning-level consensus rather than next-token prediction confidence. When multiple models converge on semantically similar responses despite surface-level variations, this convergence indicates reliable generation quality.

\cref{tab:auc_main} demonstrates that across most tasks, semantic methods achieve the highest AUC-DF values, indicating their deferral choices and output selection are stronger than other methods. Additionally, semantic cascades often leverage cheaper models than token-level ensembling approaches while achieving superior AUC-DF values, and provide more flexible configuration options.

The best-performing token-level ensembles rely exclusively on the top 2–3 base models---which are typically the most expensive---and degrade significantly in the presence of a much weaker base model. Further, token-level ensembles do not reliably improve upon their single-best ensemble model despite using additional resources. 

Accordingly, \textbf{robustness to heterogeneity} and \textbf{ensemble flexibility} emerge as critical production advantages for semantic ensembles. Semantic ensembles extract useful signal from disagreements induced by weaker models, even in extremely simple setups (\cref{apdx:simple_ensemble}).
The strong performance of heterogeneous configurations mixing 8B, 3B, and 1B models demonstrates that semantic methods effectively leverage diverse model capabilities while remaining robust to performance disparities. %\gs{the improvements in Table 2 seem quite small relative to the token-level approach? am I reading the table correctly? I wonder if there is a better way to present these results}

% \paragraph{Ensemble composition flexibility} emerges as a critical production advantage. 

\subsection{Mechanistic Advantages of Semantics}
Semantic agreement's strength stems from two areas.
First, semantic similarity's ability to reliably select a strong response from the ensemble's several outputs leads to high baseline performances, an effect which is especially apparent in top performing ensembles (\cref{fig:latency_curve}). Second, semantic similarity provides a stronger indication of when to defer. Even when baseline ensemble performance is not higher than a token-level cascade's baseline---which occurs when the semantic ensemble is comprised of unilaterally weaker (and often cheaper) models---semantic similarity can still outperform token-level methods through more efficient deferral choices arising from accurate identification of (un)reliable model outputs (\cref{fig:better-defer}). %\gs{can you make Figure 2 larger (one column), given that we have space in an arxiv submission?}\sk{Both FIGS 1\&2 are unfixed yet}

Combined, these two facets explain how semantic cascades can match the performance and cost-quality tradeoffs of token-level deferral from a stronger model, and consistently match or surpass them when the model is included in the ensemble. %---provided the model is not too much stronger than the rest of the ensemble.
In addition, these aspects explain how semantic ensembles can outperform even the 70B target model on certain tasks (SQuAD, WMT FR$\to$EN) without making use of any model- or task-specific training.

%\vspace{-0.6em}
\paragraph{Difficult short-form question answering} exposes semantic methods' core limitation, however. TriviaQA reveals that when answers are typically short (1-3 tokens) \emph{and} baseline ensemble performance is low (only Llama3.1-8B and Llama3.1-70B from our models pool are reasonably effective for this task), frequent and uninformative disagreements provide insufficient signal for deferral decisions. Token-level confidence signals achieve higher performance than semantic methods in this regime---though semantic cascades usually retain near-competitive performance in terms of AUC-DF, accuracy, and latency relative to a token-level cascade of their single-best constituent model.

\subsection{Scaling and Model Gap Effects}
Larger model gaps between ensemble and target models amplify semantic advantages. The cost differential between correct and incorrect deferral decisions grows with target model size, making accurate deferral identification increasingly valuable. Additionally, as performance gaps narrow, semantic ensembles see improved gain from highest-similarity response selection. This scaling pattern suggests semantic cascades will provide greater benefits as model sizes continue to increase---especially as the computational cost of achieving marginal performance gains increases with larger model sizes \cite{kaplan2020scaling, henighan2020scaling}.

%\gs{can you add a section here where you explore the different metrics that could be used for semantic agreement? it seems like those are in the appendix but would be of interest to the reader} \sk{@duncan, maybe you can extend the section below, with more details, esp the additional experiments ran?}
\subsection{Semantic Metric Selection}
While no single similarity metric dominates across all tasks and ensemble compositions, we do observe patterns in metric effectiveness. BLEURT and SBERT, which capture richer semantic information than n-gram overlap metrics, generalize better across diverse generation tasks. Translation especially favors these embedding-based approaches, achieving AUC-DF improvements of 0.02-0.04 over ROUGE/BLEU variants. Conversely, short-form question-answering tasks (SQuAD, TriviaQA) marginally prefer n-gram methods due to their focus on exact matching. For practical deployment, BLEURT or SBERT serve as robust default choices, with n-gram metrics reserved for very short responses. Full metric comparisons are provided in \cref{appdx:auc_values}.

\section{Discussion \& Conclusion}

% Open-ended generation fundamentally breaks the assumptions underlying cascade systems. When multiple valid outputs exist without ground truth references, traditional confidence measures---optimized for next-token prediction---can fail to capture what actually matters: whether the generated content is reliable and meaningful.

% \textbf{Semantic agreement represents a fundamentally better-suited approach for open-ended cascades.} Our findings establish that semantic similarity measures meaning-level consistency, which better aligns with generation quality assessment, compared to token-level confidence---optimized for language modeling objectives. This explains why semantic signals provide superior deferral decisions even with smaller model ensembles, as demonstrated consistently across translation and reading comprehension, while maintaining competitive performance on summarization tasks.

We demonstrate that semantic agreement provides an alternative
%(and more principled foundation)
for generative cascade decisions. %The key insight is that meaning-level consensus reveals output reliability in ways that token-level confidence cannot access.
When models converge semantically despite surface variations, this convergence signals quality more accurately than individual model confidence scores.
The implications extend beyond efficiency gains to challenge fundamental assumptions about ensemble behavior. Semantic methods exhibit counterintuitive robustness: weaker models often improve ensemble performance by providing valuable (dis)agreement signals that reliably indicate when a query should be deferred to a more capable model.
% addresses the core challenge of determining when smaller models produce reliable outputs. The findings challenge conventional assumptions about cascade design: semantic ensembles with weaker models often outperform stronger individual models, and adding lower-performing components frequently improves rather than degrades performance. These counterintuitive results stem from semantic agreement's ability to detect consensus on meaning despite surface-level variations---a capability that proves especially valuable as generation tasks become more open-ended and subjective.

Perhaps most significantly, semantic cascades are a step toward addressing the deployment barriers that have limited cascade adoption in production environments. The training-free approach eliminates the recurring costs of model-specific engineering and avoids the restrictions imposed by task-specific adaptions, while black-box compatibility enables deployment across proprietary APIs that dominate enterprise usage. When models are updated---a frequent reality in commercial deployments---semantic cascades maintain performance without recalibration, providing the operational stability that production systems require.

The efficiency gains validate this approach beyond theoretical interest. Achieving near-equivalent or superior quality at 40\% computational cost while delivering substantial latency improvements demonstrates that better uncertainty estimation and output selection directly translates to operational value. As model sizes continue to grow and deployment costs rise, the ability to make accurate deferral decisions becomes increasingly critical for sustainable LLM deployment.
By capturing output reliability through meaning rather than mechanics, semantic cascades provide a more robust foundation for deploying LLM systems that can adapt to evolving model landscapes while maintaining the flexibility that modern applications demand.

Semantic agreement establishes meaning-level consensus as a viable alternative to confidence-based uncertainty estimation, opening new research in model combination and adaptive deployment strategies. These findings suggest that the added cost of more reliable uncertainty quantification---traditionally viewed as prohibitive for cascading---may be justified by better deferral decisions, especially when the method naturally guides output selection. This raises the possibility that other uncertainty quantification methods in the literature might offer surprising benefits when applied to cascading, and methods of combining semantic agreement with token-level confidence is an appealing line of future study.
% This demonstrates that the higher cost associated with more reliable uncertainty quantification---typically considered prohibitive for cascading---can be justified by superior deferral choices, especially when the approach inherently guides better output selection. 

%%

\section{Limitations}

Semantic agreement relies on the assumption that consensus indicates output quality, which may not hold universally across generation tasks. \textbf{Code generation} likely represents a fundamental limitation where semantic similarity measures would fail to capture syntactic correctness requirements. Semantically similar code snippets may be functionally incorrect due to syntax differences, API variations, or language-specific requirements that semantic metrics cannot distinguish. 

\textbf{Creative writing tasks} present a more nuanced challenge. While diversity and originality are often desirable in creative generation, semantic agreement may favor generic or conventional outputs where models converge on similar themes or expressions. However, semantic consistency could still capture  whether models remain on-task versus producing incoherent or off-topic content, suggesting this limitation may be context-dependent.

\textbf{Sequential execution environments} eliminate several of the efficiency advantages our approach provides. In a worst-case scenario where ensemble models must run sequentially rather than in parallel, the latency associated with semantic agreement becomes pure computational overhead that may not offset deferral benefits. This constraint particularly affects resource-limited deployments where parallel model execution is infeasible for the base models.

Finally, \textbf{evaluation challenges} persist in assessing generation quality on tasks with free-form output. Our quality measurements rely on reference-based metrics that may not fully capture semantic quality, particularly for abstractive tasks where multiple valid outputs exist. This deficiency is especially apparent for XLSum, where no deferral method achieves significantly better performance than random deferral partly as a result of the fact that evaluation is based on adherence to one highly abstractive ground-truth summary (\cref{sec:apdx-summ-problems}). While results under reference-free evaluation metrics also support the effectiveness of semantic cascades (\cref{apdx:ref-free}), no metric is a perfect substitute for manual expert review---reflecting the fact that automatic evaluation of open-ended generation remains a challenge for the whole field.

\section*{Acknowledgements}
    We thank Ameet Talwalkar and Don Dennis for their helpful comments at the initiation of this work.

% Bibliography entries for the entire Anthology, followed by custom entries
%\bibliography{anthology,custom}
% Custom bibliography entries only
\bibliography{main}

\appendix
\onecolumn

% maybe I can do a quick test with partial oracles to test relative strength of deferral signals of ensemble token-level vs ensemble semantic

% add section with observation that best similarity metric correlates with downstream task? have a commented out paragraph about this in discussion

% This is a macro that that will automatically resize a table if it is too wide, it gets applied to appendix data
\newsavebox{\CBox}
\newcommand{\autosizeTable}[1]{%
  \begin{lrbox}{\CBox}
    #1
  \end{lrbox}
  \ifdim\wd\CBox > \textwidth
    \resizebox{\textwidth}{!}{\usebox{\CBox}} % Resize if too wide
  \else
    \usebox{\CBox} % Otherwise, display normally
  \fi
}

\section{Experimental Details}
\paragraph{Models and Datasets}
For all Gemma, Llama, and Qwen models, we use their official instruction-tuned versions; FLAN-T5 and mT0 are already instruction-tuned. For consistency, we use the same SBERT model across datasets. For the SBERT model, we use the cased multilingual BERT Base model \citep{bert-multilingual}, as provided through HuggingFace, due to its multilingual capabilities, small size, and speed. 

On all datasets, we evaluate on the validation split as provided on HuggingFace, or the test split if no validation split is available. Within a dataset, all models are given the same prompt (up to chat templating) in order to better mimic the real world, where live user prompts cannot easily be tuned for best performance on a particular model. On summarization tasks we provide five examples of reference summaries in the prompt; in all other tasks prompts are zero-shot. Whether different prompting strategies for each model lead to different behaviors and relative performance gains for token-level and semantic methods is an interesting topic for future analysis.

For each model, we measure average end-to-end latency on a consistent subsample of each dataset. End-to-end latency is measured for each model individually and with a batch size of 1 to prevent batching-related undercounting. 
These trials are each preceded by a warm-start, where inference is performed on an irrelevant set of tokens and the results discarded. All experiments are performed on an L40S GPU. As Llama3.1-70B does not fit on a single L40S GPU, we parallelize it across four L40S GPUs instead. We note that this means GPU usage times are roughly four times larger than Llama3.1-70B's latency would suggest---meaning deferral strategies yield greater efficiency savings with respect to this metric should the costs of multi-GPU usage be taken into account.

\begin{table}[H]
    \centering
    \begin{tabular}{cccccccccc}
    \toprule
         & Qwen2.5-7B & Qwen2.5-3B & Qwen2.5-1.5B & Qwen2.5-0.5B \\
    \midrule
        Parameters & 7.62B & 3.09B & 1.54B & 494M\\
    \bottomrule
    \end{tabular}
    \caption{Number of Qwen2.5 model parameters, as reported by HuggingFace.}
\end{table}

\begin{table}[H]
    \centering
    \begin{tabular}{cccccccccc}
    \toprule
         & Llama3.1-70B & Llama3.1-8B & Llama3.2-3B & Llama3.2-1B \\
    \midrule
        Parameters & 70.6B & 8.03B & 3.21B & 1.24B  \\
    \bottomrule
    \end{tabular}
    \caption{Number of Llama model parameters, as reported by HuggingFace.}
\end{table}

\begin{table}[H]
    \centering
    \begin{tabular}{cccccccccc}
    \toprule
         & Gemma3-1B & FLAN-T5-Large & mT0-Large & BERT Base (multilingual, cased) \\
    \midrule
        Parameters & 1.00B & 783M & 1.23B & 179M \\
    \bottomrule
    \end{tabular}
    \caption{Number of model parameters, as reported by HuggingFace.}
\end{table}

\paragraph{Ensemble Deferral}
In \cref{sec:method-defer-protocol} it is described how, in the context of ensemble deferral, we set $s=\max_i o_i$. We also tested setting $s=\text{mean}_i o_i$, and observe nearly identical results. This is also true for token-level ensemble cascades.

%For token-level cascades, we analyze Chow-Quantile for quantiles $q\in\{0.0, 0.1, \dots, 1.0\}$. This provides a balance between expressivity and avoiding spuriously high ``Best Quantile" results due to noise.

We use the BLEU implementation as provided by SacreBLEU \citep{sacrebleu}.

%dont forget other things mentioned in comments in main report

%should describe what data was tested on (when random subsamples were taken, the splits),

% wait is what I did technically BERTScore and not SBERT?

% we should maybe really be using things like BARTScore or COMETKiwi maybe
\section{BLEURT Model Sizes}
\begin{figure}[H] 
    \centering
    \begin{subfigure}{0.32\linewidth}
    \includegraphics[width=\linewidth]{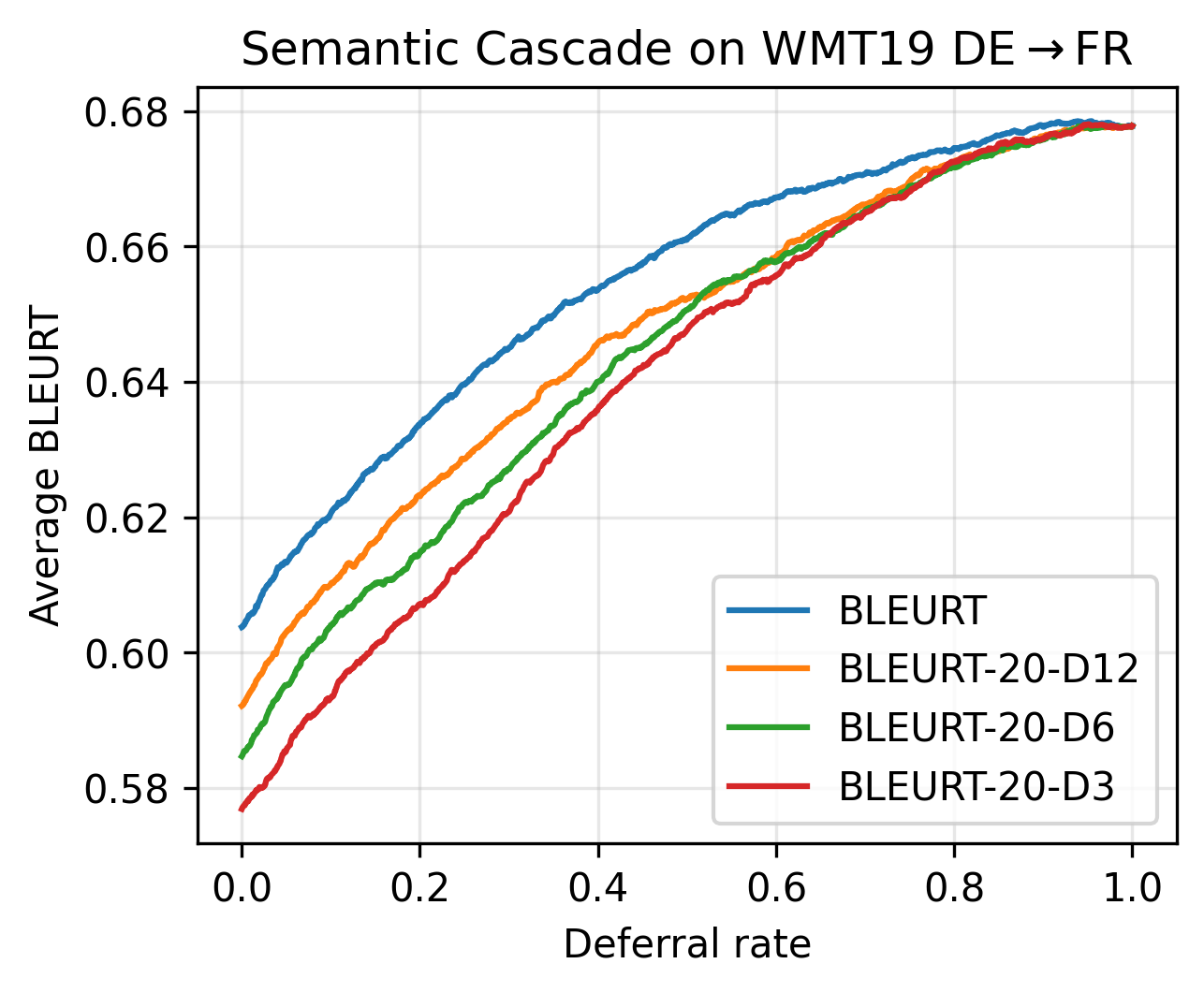}
    \caption{}
    \end{subfigure}
    \begin{subfigure}{0.32\linewidth}
    \includegraphics[width=\linewidth]{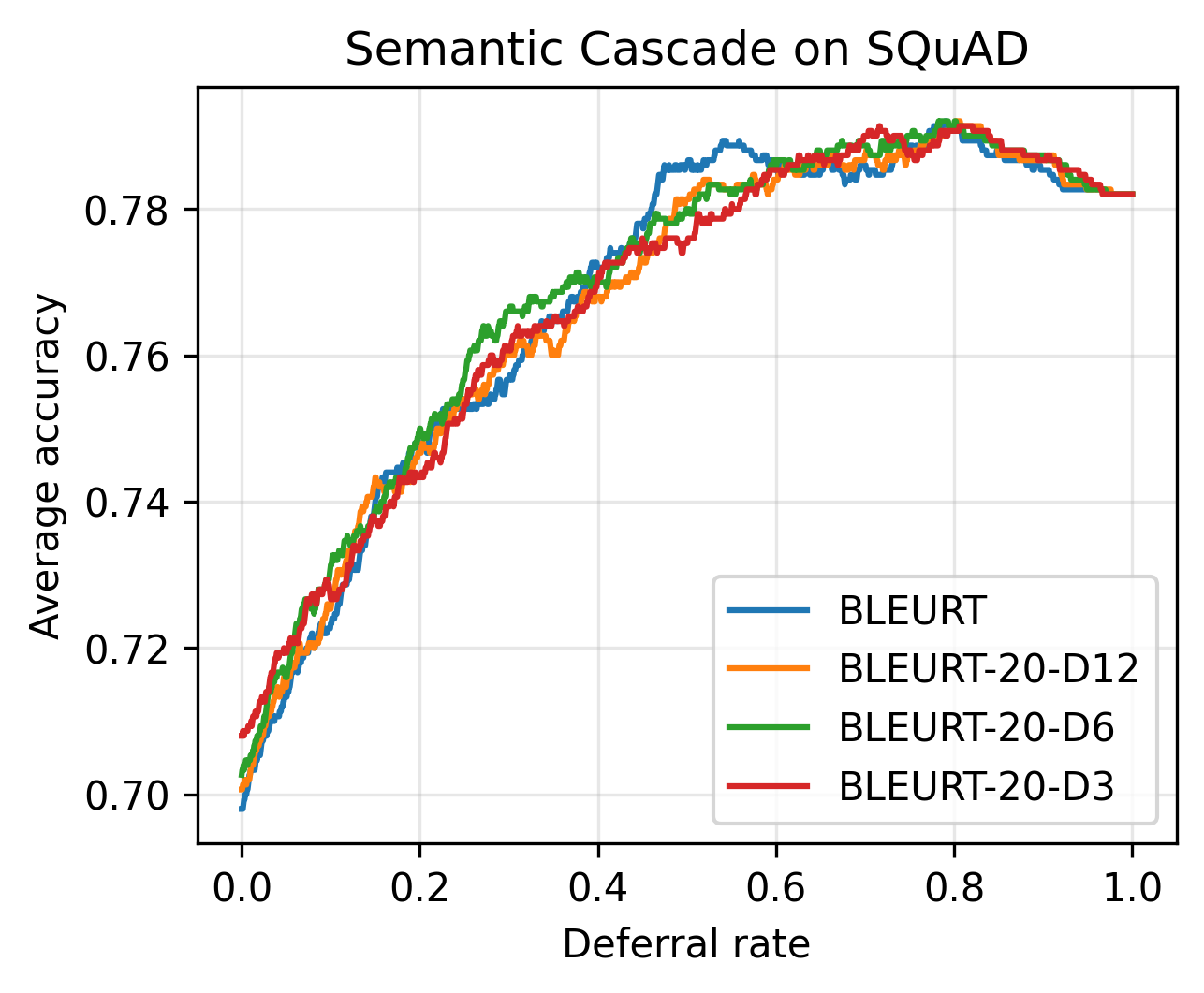}
    \caption{}
    \end{subfigure}
    \caption{\textbf{(a)} Comparison of deferral curves for different BLEURT sizes for a semantic cascade of Qwen2.5-1.5B, Gemma3-1B, mT0-Large, deferring to large model Llama3.1-8B. Smaller sizes of BLEURT lead to worse performance. \textbf{(b)} A similar comparison on SQuAD1.1 of the same semantic ensemble, deferring to Qwen2.5-7B. Smaller sizes of BLEURT do not impact cascade performance due to the short nature of responses and the binary evaluation scheme.}
    \label{fig:apdx-bleurt-comparison}
\end{figure}

There are several different official BLEURT model sizes available. We use BLEURT-20 in our experiments, which has 30 layers and 579M parameters, but there are also three lossily compressed versions available: BLEURT-20-D12 (12 layers, 167M parameters), BLEURT-20-D6 (6 layers, 45M parameters), and BLEURT-20-D3 (3 layers, 30M parameters). We find that on domains with longer output lengths, the size (quality) of the BLEURT model becomes important, but is largely irrelevant at very small scales (\cref{fig:apdx-bleurt-comparison}).

% On QA datasets with short outputs, the cost of using a more expensive (larger) BLEURT model are more relevant relative to the cost of running the ensemble models, but the accuracy impact of using a larger BLEURT model is minimal. Meanwhile, for longer-form tasks like translation, the cost of using a more expensive BLEURT model is negligible compared to the cost of running the autoregressive models to generate an output, and the accuracy impact of using a larger model is more significant in these cases.

% This suggests that, for short outputs, one can use a small semantic similarity model, mitigating the cost without impacting performance, and for long outputs one can use a large semantic similarity model, which is still cheap in relative terms and leads to higher quality outputs.

\section{Reference-Free Evaluations} \label{apdx:ref-free}
Due to concerns over the reference-based nature of our primary evaluation metrics for open-ended generation (translation and summarization), we present additional results under alternate evaluation schemes for these scenarios. Semantic cascades retain their advantage over token-level approaches under these conditions.

\begin{table*}[h]
\label{tab:apdx-ref-free}
\centering
\caption{Results for alternate evaluation metrics using COMETKiwi-XL (translation) and G-Eval (summarization)}
\small
\begin{tabular}{@{}lcccc@{}}
\toprule
\textbf{Task} & \textbf{Metric} & \textbf{Best Semantic Ensemble} & \textbf{Best Token-Level} & \textbf{Large Model (70B)} \\
\midrule
\multirow{2}{*}{\textbf{CNN/DM}} 
& Performance at 40\% Budget & \textbf{3.99} & \textbf{3.99} & 4.10 \\
& Latency at 98\% Quality (ms) & \textbf{3,937} & 5,014 & 6,632 \\
\midrule
\multirow{2}{*}{\textbf{WMT FR$\to$EN}} 
& Performance at 40\% Budget & \textbf{0.748} & 0.747 & 0.747 \\
& Latency at 99.75\%* Quality (ms) & \textbf{522} & 621 & 1,319 \\
\bottomrule
\end{tabular}
\begin{tablenotes}
\small
\item Best token-level base model: CNN/DM: Llama3.1-8B, WMT: Qwen2.5-7B.
\item Ensemble models: CNN/DM and WMT: [Llama3.1-8B, Qwen2.5-7B, Llama3.2-3B].
\item *Due to a tighter range of baseline scores using COMET compared to BLEURT, we adjust latency reporting to be at 99.75\% for this task.
\end{tablenotes}
\end{table*}

For G-Eval, we evaluate on a subset of the data, and average GPT evaluations over 5 trials for each response. We note that, unfortunately, G-Eval results are not fully reproducible due to the closed nature of ChatGPT and its stochastic responses.

\section{Worse-on-Average Models Provide Useful Signal for Deferral}\label{apdx:simple_ensemble}
\begin{figure}[H]
    \centering
    \begin{subfigure}{0.32\linewidth}
    \includegraphics[width=\linewidth]{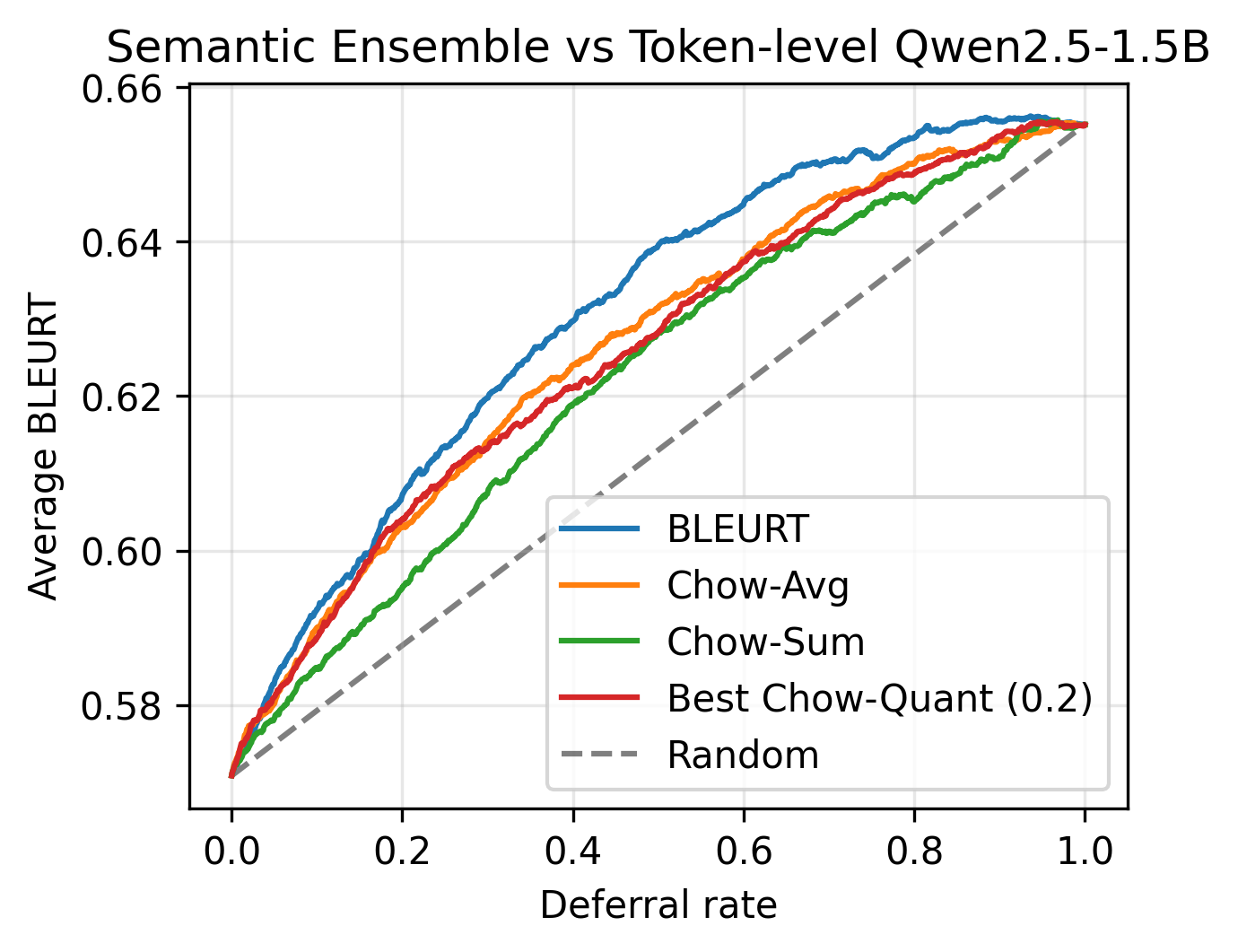}
    \caption{WMT19 DE$\to$FR}
    \end{subfigure}
    \begin{subfigure}{0.33\linewidth}
    \includegraphics[width=\linewidth]{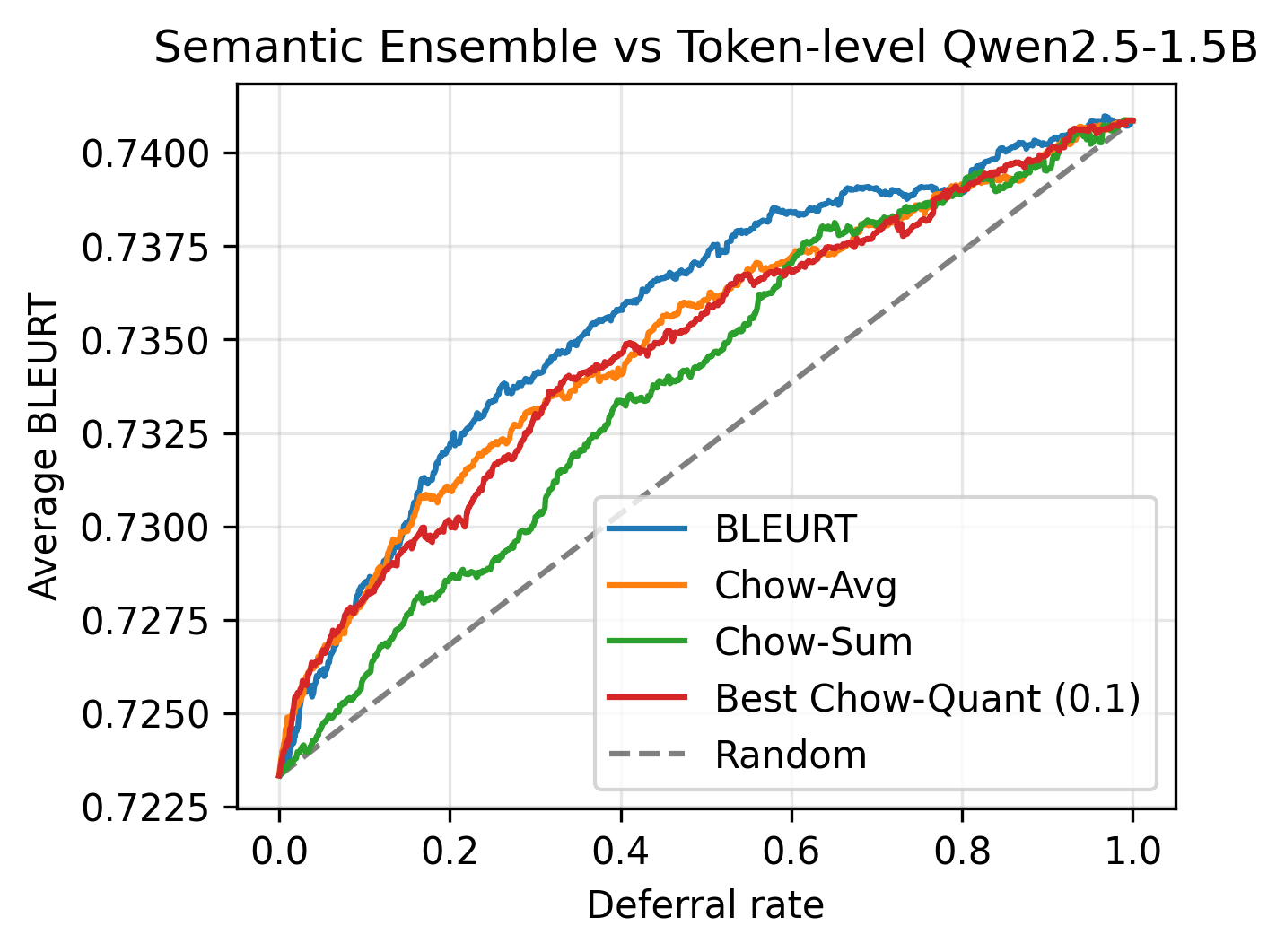}
    \caption{WMT14 FR$\to$EN}
    \end{subfigure}
    \begin{subfigure}{0.32\linewidth}
    \includegraphics[width=\linewidth]{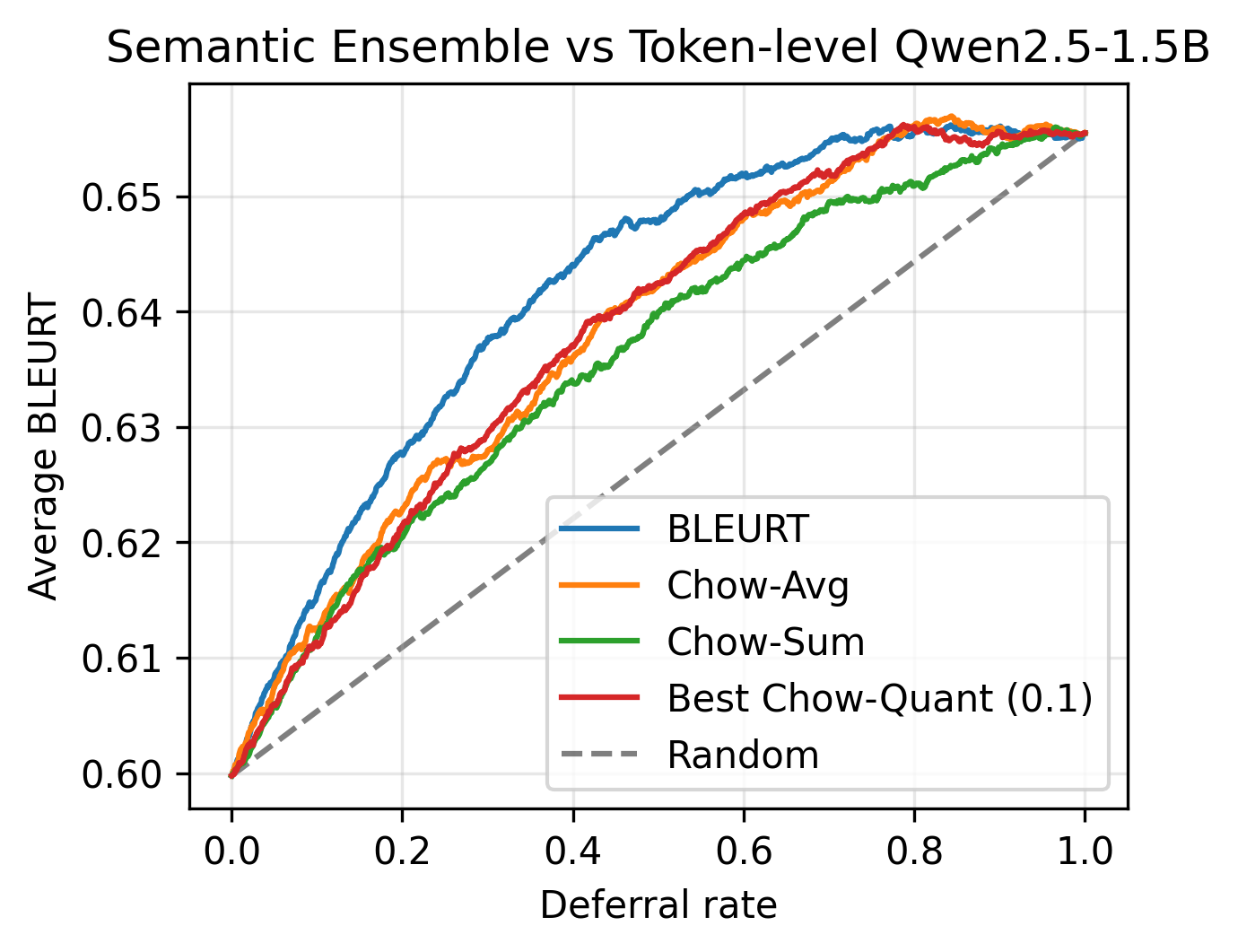}
    \caption{WMT14 EN$\to$FR}
    \end{subfigure}

    \caption{Deferral curve on WMT19 DE$\to$FR, WMT14 FR$\to$EN, and WMT14 EN$\to$FR for a simplest semantic ensemble of Qwen2.5-1.5B and Qwen2.5-0.5B, always using the outputs of Qwen2.5-1.5B when not deferring, plotted with deferral curves for token-level Qwen2.5-1.5B. Both cascades defer to Qwen2.5-7B. In all cases, the deferral curve from this semantic cascade improves over single-model token-level deferral signals, improving for instance the AUC to .6307 on DE$\to$FR over single-model token-level deferral signals (Best (Chow-Avg) AUC: .6263), and significantly over random deferral (AUC: .6130). This demonstrates how, even in a very simple ensemble, semantic similarity with a substantially \emph{worse} model can still provide a strong indication for when deferral is appropriate.}
    \label{fig:apdx-simplechow}
\end{figure}

\section{Difficulties with Assessing Summarization} \label{sec:apdx-summ-problems}
Summarization is difficult to fairly assess as a result of the fact that FLAN-T5 outperforms the ``target" large models by a large degree on CNN/DailyMail and XLSum (\cref{tab:baselines}). For this reason, we exclude FLAN-T5 from our main analysis. 
FLAN-T5 typically performs less well on other datasets, and as semantic cascading is usually able to account for its performance on other datasets and come close to matching its performance on summarization when it is included in the ensemble, semantic ensembling arguably remains a solid choice.

However, it should be noted that ROUGE scores are highly noisy, imperfect measures of summarization quality. Further, there are often many equally satisfactory ways of summarizing a given article, but there exists only one ground-truth summary per example in these datasets. Additionally, ROUGE does not account for factual inaccuracies. Together, these factors mean that ROUGE scores alone cannot reliably determine summarization quality.

Based on a qualitative analysis we performed on a small sample of model responses, the differences in ROUGE performance appear mostly to be a result of models giving more detail than FLAN-T5-Large (and using complete sentences in the case of CNN/DailyMail, where target summaries are often comprised of multiple sentence fragments), rather than these models outputting strictly ``worse" summaries. In other words, there is a stylistic mismatch between their outputs and the targets, but not necessarily a semantic one. Furthermore, we observed more factual inaccuracies in the summaries generated by FLAN-T5-Large and mT0-Large than in those generated by the large models (\cref{tab:apdx-sum_ex}).
Different prompt styles tailored to each model might better force the models to adhere to the desired style and be able to close this performance gap. However, for accurate assessments of summarization quality, human annotation is likely necessary. Other approaches like G-Eval \cite{liu2023gevalnlgevaluationusing}, which use LLMs to assess summarization quality, may provide a superior substitute, and we provide experiments showing our method's advantage under this evaluation metric as well. However, these lack interpretability and risk biasing evaluations in other ways.

%should also discuss how these datasets are challenging to fairly assess, since the gap between (e.g. on CNNDM) between the roughly .133 and .145 and .133 and .103 is huge, it's almost 3x as large gap, so the model performances on this (while they dont seem it) are actually strongly heterogeneous. if we instead defer to a model that better mirrors the relative gap, aka does much better (.2 rouge) we see the same sorts of curve shapes for both token level and semantic ensemble

\newcommand{\hlc}[2][yellow]{{%
    \colorlet{foo}{#1}%
    \sethlcolor{foo}\hl{#2}}%
}
\begin{table}[H]
    \centering
    \begin{tabular}{p{0.45\linewidth} | p{0.45\linewidth}}
        \texttt{The Dallas Mavericks player accused Ted Kritza of taking the money . He believes Kritza took it from his bank credit line without his permission . In a recorded phone call with Jefferson, Kritza 'confesses to wrongdoing' Recording of the conversation is now in the hands of FBI .} 
        & \texttt{Dallas Mavericks player Richard Jefferson helped FBI find \$2million dollars that was taken from his bank credit line without his permission. \hlc[orange!40]{Jefferson, 34, had reported the crime before the investigation began.} \hlc[red!30]{In a recorded phone call with Kritza, Jefferson 'confesses to wrongdoing'} The recording is now in the hands of the FBI.}\\
        \midrule
        \texttt{NBA player Richard Jefferson helped the FBI recover \$2 million after accusing his former business manager, Ted Kritza, of stealing the money from his bank credit line without permission. Jefferson recorded a phone call with Kritza, in which Kritza allegedly confessed to the crime, and the recording is now in the hands of the FBI. \hlc[orange!40]{Jefferson is seeking to put a hold on a lawsuit from a bank that is seeking part of the \$2 million until the federal investigation into Kritza is complete.}} 
        & \texttt{Richard Jefferson, a Dallas Mavericks basketball player, assisted the FBI in recovering \$2 million that his former business manager, Ted Kritza, allegedly stole from Jefferson's bank credit line. The case is currently under federal investigation, \hlc[orange!40]{and Jefferson is seeking to delay legal action from a bank that has sued him for a portion of the stolen funds until the investigation is complete.}} \\
    \end{tabular}
    \caption{Ground-truth and predicted summaries on CNN/DailyMail. Top left: ground-truth. Top right: FLAN-T5-Large (ROUGE-2: .4583). Bottom left: Llama3.1-8B (ROUGE-2: .2656). Bottom right: Qwen2.5-7B (ROUGE-2: .09346). Factual inaccuracies are highlighted in red, and superfluous information to the ground truth are highlighted in orange. Ground truth summary is comprised of sentence fragments, whereas models output longer full sentences and sometimes unnecessary detail. Qualitatively, FLAN-T5 exhibits better stylistic matching but higher rates of factual errors.}
    \label{tab:apdx-sum_ex}
\end{table}

%Ensemble deferral strategies further struggled on these datasets due to large relative differences in baseline model performance. For instance, on CNN/DailyMail, the difference in baseline performance between Qwen2.5-1.5B and Llama3.2-1B is almost three times as large as the difference in baseline performance between Llama3.2-1B and Llama3.2-8B.

\begin{figure}[H]
    \centering

    \begin{subfigure}{0.32\linewidth}
    \includegraphics[width=\linewidth]{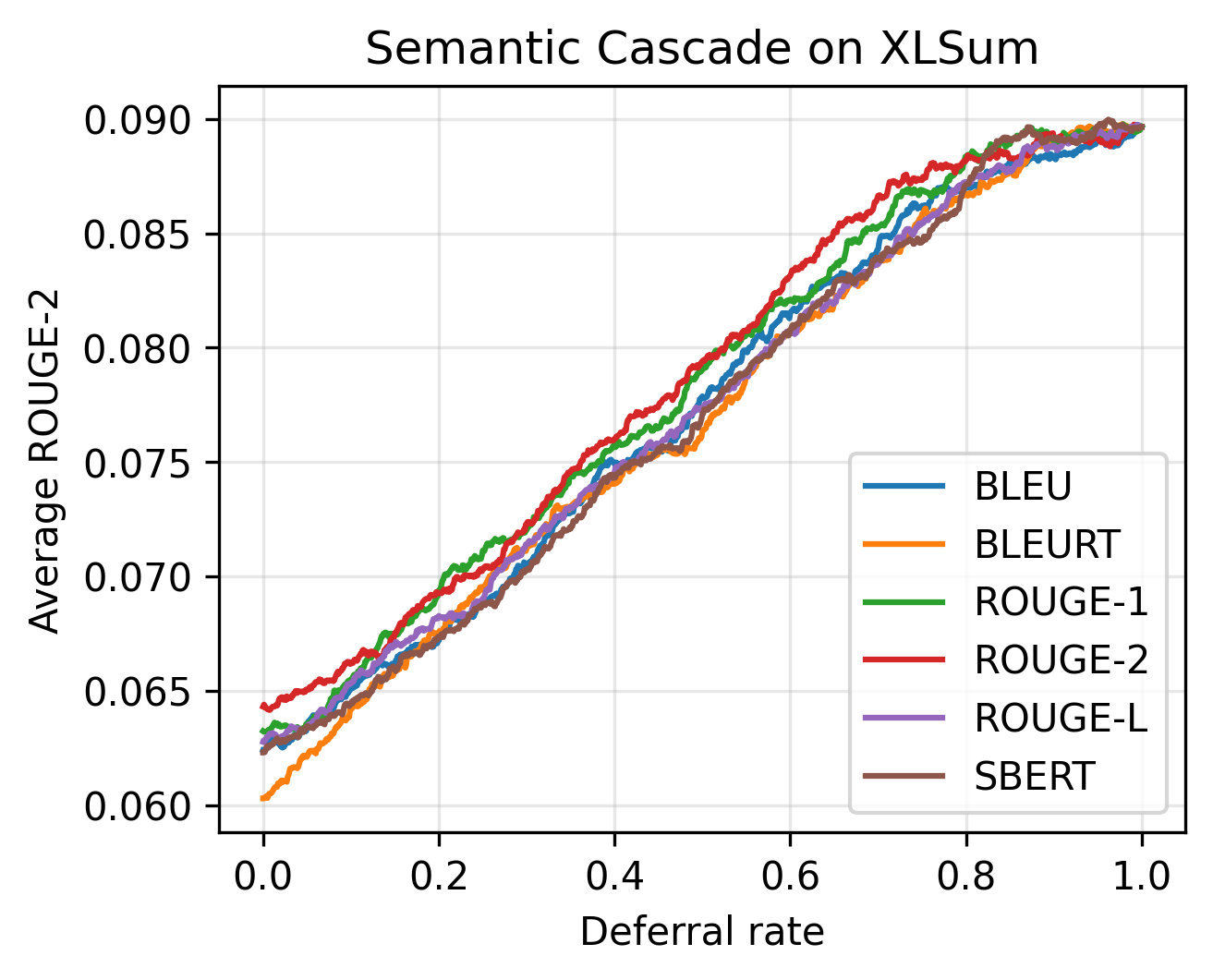}
    \end{subfigure}
    \hfill
    \begin{subfigure}{0.32\linewidth}
    \includegraphics[width=\linewidth]{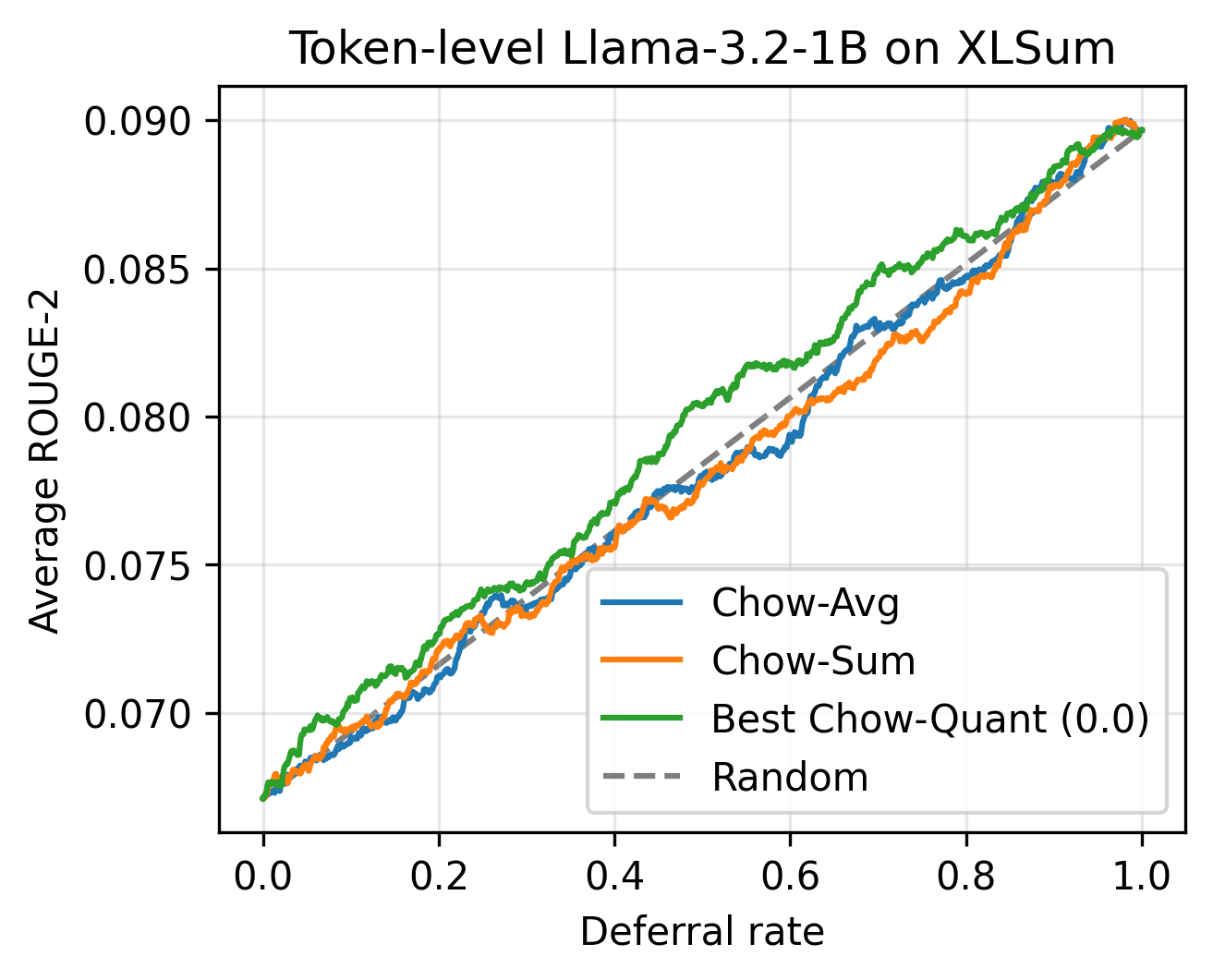}
    \end{subfigure}
    \hfill
    \begin{subfigure}{0.32\linewidth}
    \includegraphics[width=\linewidth]{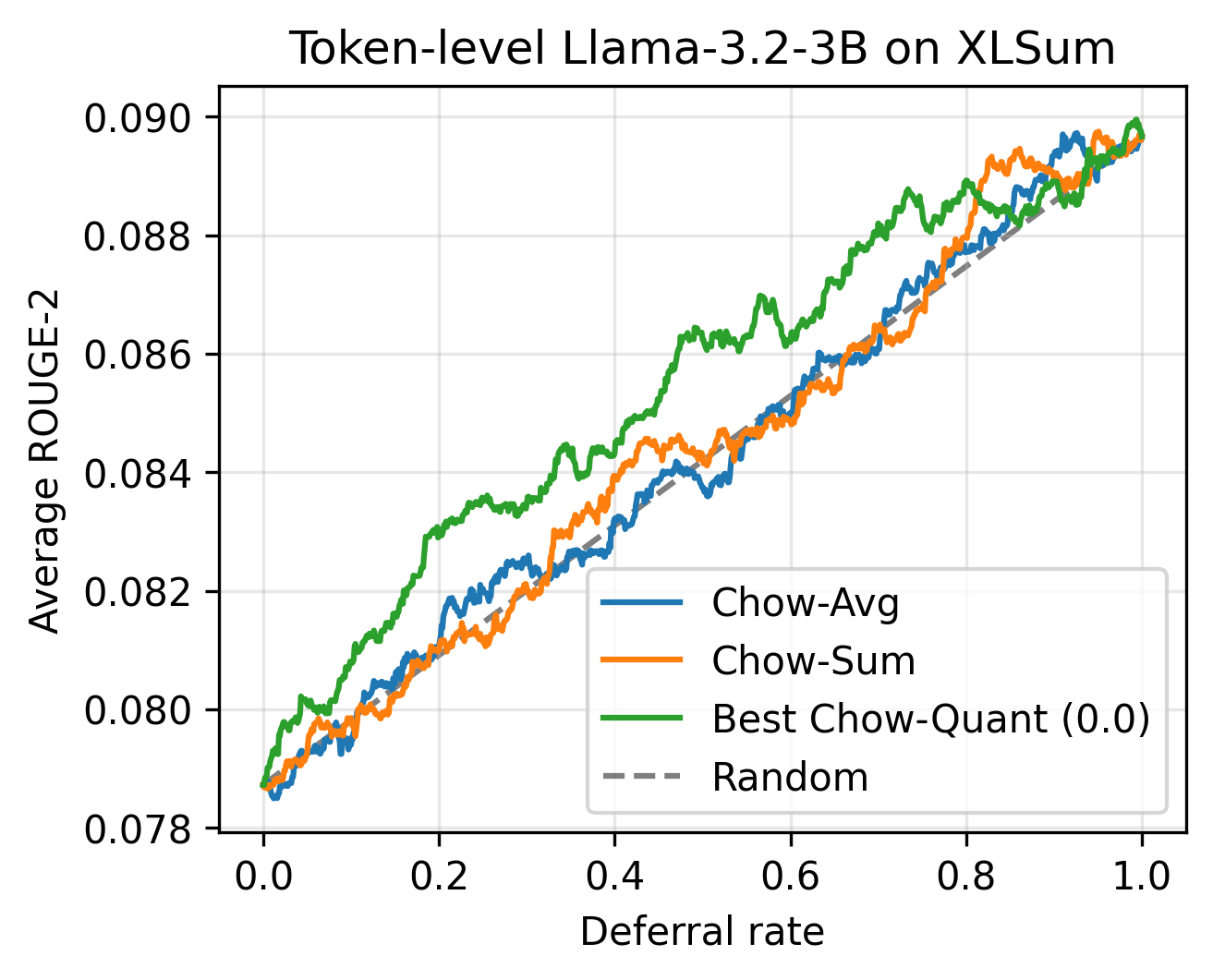}
    \end{subfigure}

    \begin{subfigure}{0.32\linewidth}
    \includegraphics[width=\linewidth]{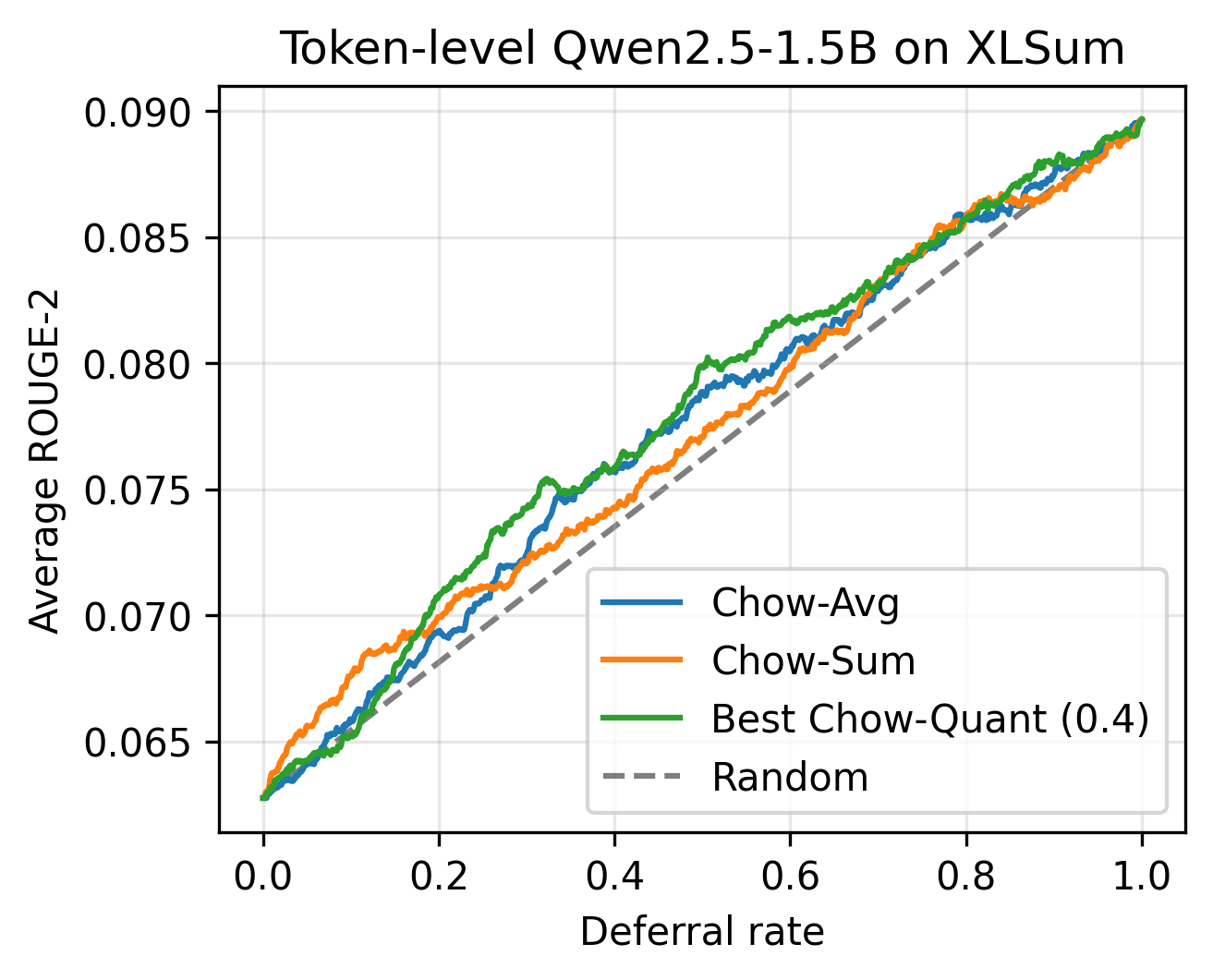}
    \end{subfigure}
    \hfill
    \begin{subfigure}{0.32\linewidth}
    \includegraphics[width=\linewidth]{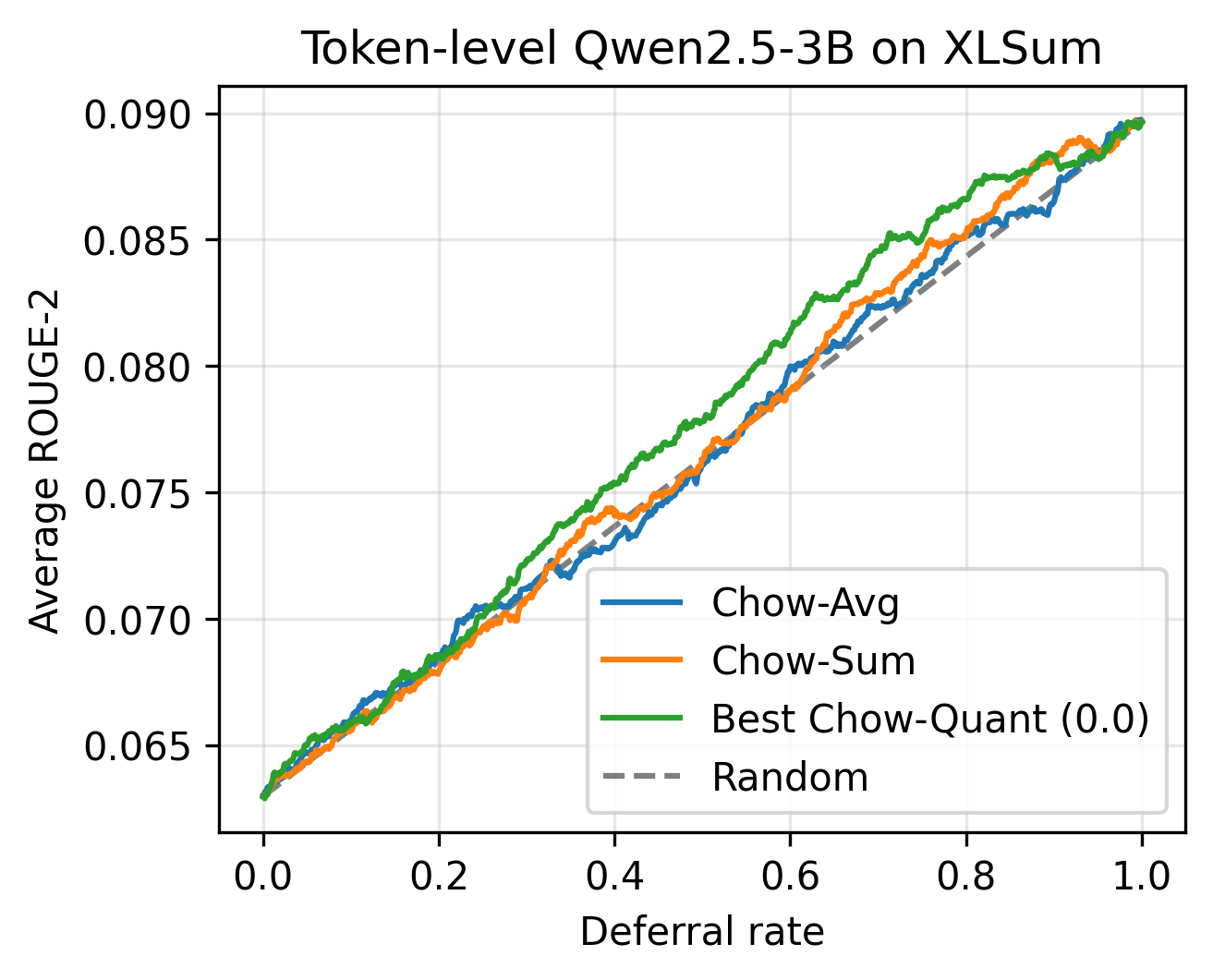}
    \end{subfigure}
    \hfill
    \begin{subfigure}{0.32\linewidth}
    \includegraphics[width=\linewidth]{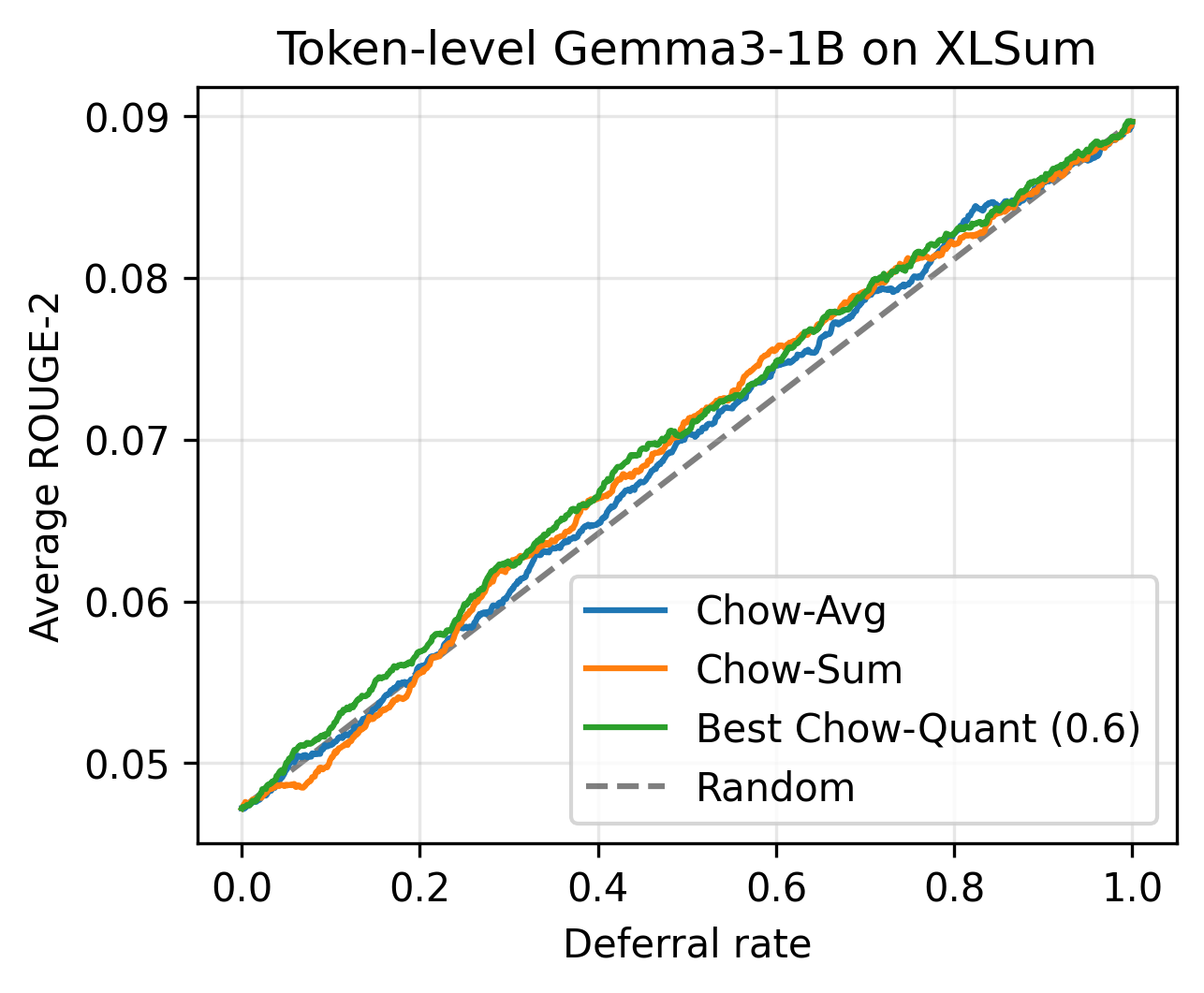}
    \end{subfigure}
    
    \caption{Deferral curves on XLSum. Semantic cascade's ensemble is comprised of Llama3.2-1B, Qwen2.5-1.5B, Gemma3-1B. All cascades defer to Llama3.1-8B. No cascades improve significantly on random deferral from their single-best ensemble model (only ensemble model, for token-level cascades), hence deferral curves are nearly flat. The semantic cascade has a slightly better curve shape than the token-level cascades, but begins at a lower baseline than its single-best ensemble model (Llama3.2-1B). This is enough for it to achieve similar AUC (.0786 versus 0.0795), and match or exceed performance at the 60\% deferral rate and above.}
    \label{fig:apdx-xlsum}
\end{figure}

Further, we note that on XLSum \emph{no deferral strategy} significantly improves on random deferral---especially at smaller scales---leading to flat deferral curves and baseline performance becoming the dominating factor (\cref{fig:apdx-xlsum}).

\section{Ensembling Token-level Models}\label{apdx-token-level-ensemble}
Due to differences in baseline confidence ranges, normalization was performed when ensembling token-level models. Table \ref{tab:apx-token-level-ensemble} presents the mean Chow-Avg and Chow-Sum values across responses in each dataset, providing a rationale for this decision.

\begin{table}[H] 
\centering
\caption{Mean Chow-Avg and Chow-Sum across datasets}
\label{tab:apx-token-level-ensemble}
\autosizeTable{
\begin{tabular}{lllllllll}
\hline
Model     & \multicolumn{2}{c}{TriviaQA}                                & \multicolumn{2}{c}{SQuAD}                                   & \multicolumn{2}{c}{WMT FR$\to$EN}                           & \multicolumn{2}{c}{CNN/DM}                                  \\
          & \multicolumn{1}{c}{Avg} & \multicolumn{1}{c}{Sum} & \multicolumn{1}{c}{Avg} & \multicolumn{1}{c}{Sum} & \multicolumn{1}{c}{Avg} & \multicolumn{1}{c}{Sum} & \multicolumn{1}{c}{Avg} & \multicolumn{1}{c}{Sum} \\ \hline
Llama3.1-8B   & -0.2378                      & -1.3165                      & -0.1209                      & -0.6837                      & -0.1162                      & -2.9247                      & -0.2189                      & -21.2436                     \\
Llama3.2-3B   & -0.2990                      & -1.2517                      & -0.1467                      & -0.7346                      & -0.1405                      & -3.5432                      & -0.2455                      & -29.0617                     \\
Llama3.2-1B   & -0.6003                      & -2.6633                      & -0.2863                      & -1.1109                      & -0.2442                      & -5.9975                      & -0.3268                      & -35.6698                     \\
Qwen2.5-7B   & -0.1929                      & -0.8508                      & -0.0288                      & -0.1835                      & -0.0461                      & -1.2387                      & -0.1945                      & -11.7805                     \\
Qwen2.5-3B   & -0.2733                      & -1.1296                      & -0.0541                      & -0.3209                      & -0.0664                      & -1.7102                      & -0.3397                      & -27.4709                     \\
Qwen2.5-1.5B & -0.3850                      & -1.7507                      & -0.0882                      & -0.5509                      & -0.1366                      & -3.5275                      & -0.3694                      & -15.1451                     \\
Qwen2.5-0.5B & -0.6137                      & -3.7378                      & -0.1955                      & -1.1930                      & -0.2966                      & -7.5582                      & -0.4268                      & -21.1389                     \\
Gemma3-1B   & -0.2602                      & -1.0226                      & -0.0667                      & -0.3466                      & -0.0635                      & -1.5018                      & -0.1625                      & -12.5079                     \\
FLAN-T5-Large  & -0.8618                      & -4.7093                      & -0.1070                      & -0.9355                      & -0.4093                      & -10.6007                     & -0.3233                      & -20.2584                     \\
mT0-Large & -1.2154                      & -5.3315                      & -0.1285                      & -0.5957                      & -0.4325                      & -12.2072                     & -0.4301                      & -24.4393                     \\ \hline
\end{tabular}
}
\end{table}

When model confidences within an ensemble do not fall within a close range, this leads to one model systematically dominating the others when it comes time to select model responses. Further, model confidence range is not always correlated to model performance, and in many instances worse models have higher average confidence. In order for naive token-level ensembling to be effective, the ensemble must consist of models whose baseline confidences \textit{and} response qualities fall within a similar range. Normalization removes the confidence range constraint, but still requires that baseline model qualities are similar---otherwise the worse model(s) will harm performance significantly as they are picked roughly as frequently as the stronger model(s).

\section{TriviaQA: Performance Gap and Oracle Insights}{\label{apdx:triviaqa-oracle}}
As noted, closed-book QA poses a challenge for semantic similarity. This is especially true at smaller scales: 3B models outperform all ensembles of the same total size by wide margins. For example, a token-level cascade with Llama3.2-3B deferring to Llama3.1-8B achieves an AUC of .6016, compared to an AUC of just .5044 for a semantic ensemble of the three best 1B models. 
Even still, semantic cascades typically nearly match, or occasionally exceed, their single-best ensemble model's performance.

\begin{figure}[H]
    \centering
    \begin{subfigure}{0.32\linewidth}
    \includegraphics[width=\linewidth]{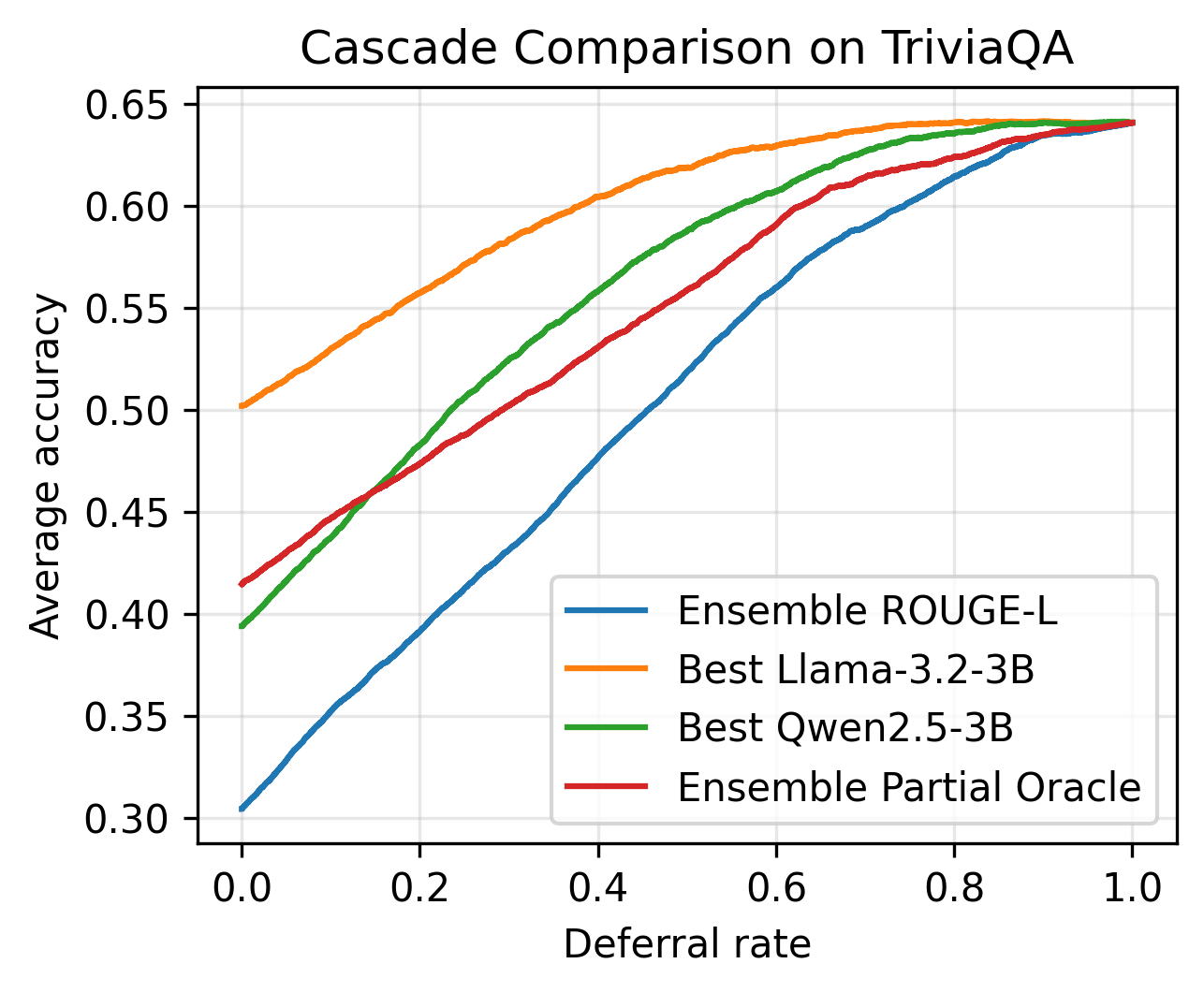}
    \caption{}
    \end{subfigure}
    \begin{subfigure}{0.32\linewidth}
    \includegraphics[width=\linewidth]{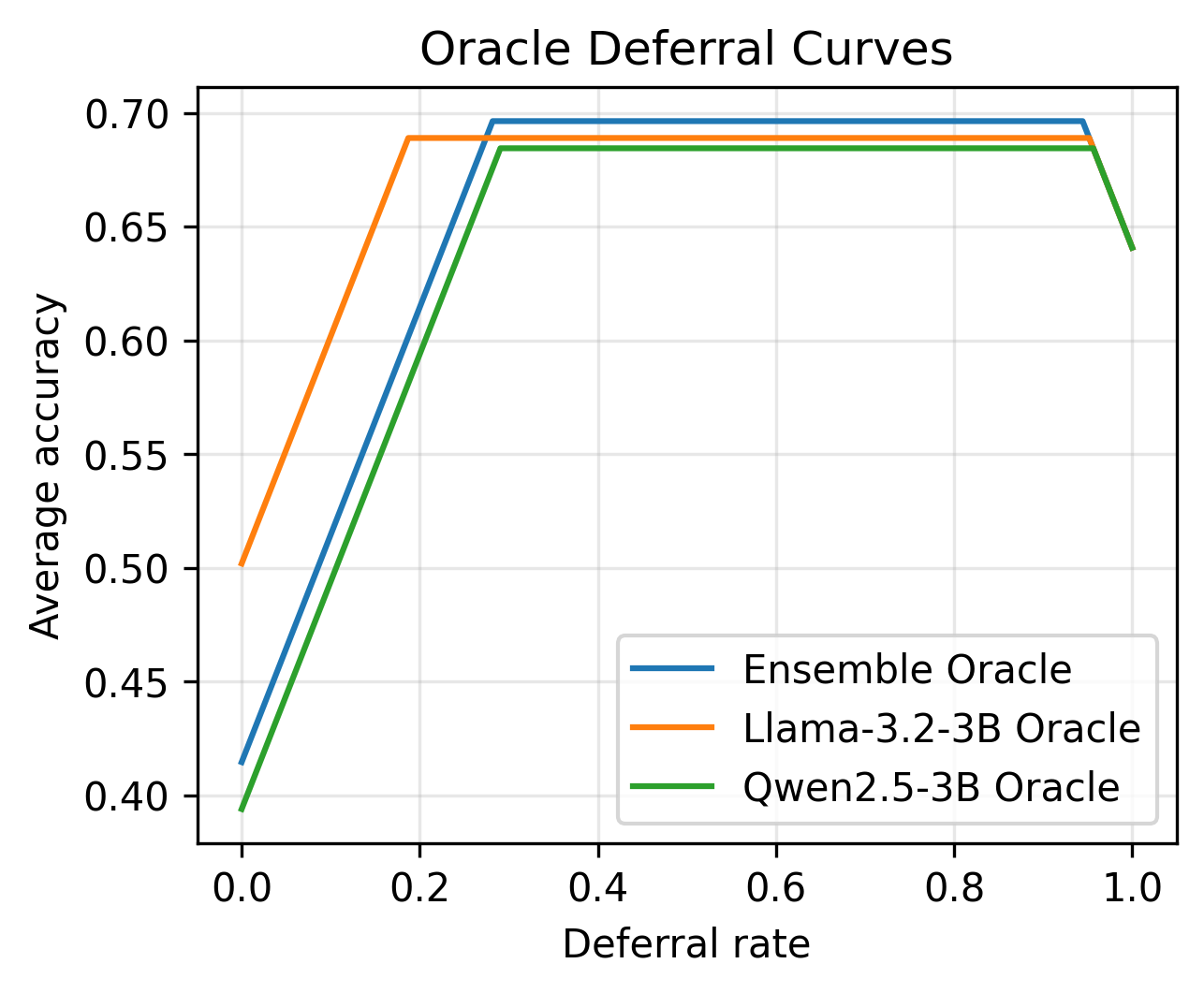}
    \caption{}
    \end{subfigure}
    \caption{\textbf{(a)} Deferral curves on TriviaQA for token-level cascades and a semantic cascade with ensemble models Qwen2.5-1.5B, Llama3.2-1B, Gemma3-1B. The partial oracle always selects the best output among the ensembles but uses semantic similarity as a deferral rule. All cascades defer to Llama3.1-8B. \textbf{(b)} Oracle deferral curves. Half of the initial performance gap between the ensemble method and Llama3.2-3B is closed simply by selecting the best ensemble outputs, and the ensemble oracle lies strictly above the Qwe2.5-3B oracle, demonstrating that differences in model-encoded knowledge cannot by themselves adequately explain the observed differences in baseline performance. }
    \label{fig:apdx-triviaqa-oracle}
\end{figure}

It is tempting to assume the performance gap between 3B parameter models and ensembles of 1B parameter models arises because the 3B parameter models simply encode more information than the smaller ensemble models. Surprisingly, however, the oracle deferral curves for this task suggest that this explanation is incomplete and that other dynamics are at play (\cref{fig:apdx-triviaqa-oracle}). We hypothesize that the gap in performance originates not simply from a lack of encoded information, but also due to the difficulty of the ensemble deferral task, which requires not only identifying when to defer to a large model, but also selecting the best answer from the ensemble's multiple outputs when not deferring. This challenge is particularly pronounced in TriviaQA, where answers are short and baseline ensemble models perform poorly---leading to frequent disagreement from which semantic similarity struggles to extract meaningful signal.

% maybe we could use the two kinds of partial oracle (we only discuss one here, but the other is deferral oracle that doesn't get to choose the outputs) to discuss something? like prove point that deferral is hard when we don't get signal from disagreements?

To illustrate this, we plot oracle curves representing the optimal performance achievable with perfect knowledge of each model's output quality on every example. For cascades involving a single baseline model, the oracle curve is computed by deferring based on the score difference between the large and small models. In the case of ensemble-based cascades, the oracle defers based on the score difference between the large model and the best-performing output among the small models in the ensemble; when not deferring, it selects the best small-model output. Additionally, we define a partial oracle, which always chooses the best output from the small models but relies on semantic similarity to decide whether to defer.

%\cref{fig:apdx-triviaqa-oracle} further demonstrates that, with perfect knowledge of which ensemble output is best (but \emph{not} its downstream accuracy), half of this gap can be closed. However, even with this, assessing when to defer from semantic information is still problematic, . However, even with this, properly assessing when to defer is still challenging in this case: baseline performances are low, and answers are short---resulting in frequent and semantically unhelpful disagreements.

% want a plot with the oracle curves for each (Llama 3B, Qwen 3B, ensemble 1B)
% a plot with baseline model performances, and  a partial oracle

\section{Additional Experimental Results}\label{apdx:experimental_results}

\subsection{Baseline Model Performances}\label{apdx:baselines}

\begin{table}[H]
    \centering
    \resizebox{\columnwidth}{!}{
\begin{tabular}{lccccccc}
    \toprule
    	Model &  \tworow{WMT19 DE$\to$FR \\ (BLEURT)}  & \tworow{WMT14 FR$\to$EN \\ (BLEURT)} & \tworow{WMT14 EN$\to$FR \\ (BLEURT)} & \tworow{CNN/DailyMail \\ (ROUGE-L)} & \tworow{XLSum \\ (ROUGE-L)} & \tworow{SQuAD1.1 \\ (Accuracy)} & \tworow{TriviaQA \\ (Accuracy)} \\
    \midrule
        Llama3.1-70B & .7260 & .7465 & .7069 & .2672 & .2454 & .8273 & .8073 \\
         Llama3.1-8B & .6802 & .7400 & .6791 & .2431 & .2248 & .7700 & .6532 \\
     	Llama3.2-3B  & .6177 & .7292 & .6486 & .2372 & .2032 & .6980 & .5095 \\
     	Llama3.2-1B  & .4715 & .6763 & .5772 & .2292 & .1777 & .4347 & .2457 \\
     	Qwen2.5-7B    & .6595 & .7411 & .6573 & .2487 & .2099 & .7833 & .5078 \\
     	Qwen2.5-3B    & .6120 & .7298 & .6282 & .2350 & .1763 & .6973 & .3932 \\
     	Qwen2.5-1.5B  & .5789 & .7255 & .6115 & .2277 & .1933 & .6913 & .3039 \\
     	Qwen2.5-0.5B  & .3989 & .6839 & .4878 & .2165 & .1698 & .4567 & .1064 \\
     	mT0-Large     & .4467 & .6636 & .5120 & .2383 & .1903 & .7767 & .0601 \\
     	FLAN-T5-Large & .3963 & .6883 & .4534 & .2955 & .3094 & .5973 & .1449 \\
     	Gemma3-1B     & .5670 & .7090 & .6240 & .2224 & .1747 &  .5247 & .2052 \\
    \bottomrule
	\end{tabular}
}

    \caption{Baseline model performances. At each base model tier (roughly 1B, 3B, 7-8B), no single model dominates across datasets.}
    \label{tab:baselines}
    
\end{table}

\subsection{AUC-DF Values}\label{appdx:auc_values}

\subsubsection{WMT19 DE$\to$FR} \label{apdx:wmt_de_fr}
\begin{table}[H]
\caption{AUC-DF values values on WMT19 DE$\to$FR for token-level cascades. All cascades defer to Llama3.1-70B. The Random column represents the expected value if queries are deferred uniformly at random.}
\autosizeTable{
\begin{tabular}{lccccc}
\toprule
Model & Random & Chow-Sum & Chow-Avg & Best Chow-Quantile & Best Quantile \\
\midrule
Qwen2.5-7B & 0.6928 & 0.6996 & \textbf{0.7032} & 0.7023 & 0.1 \\
Qwen2.5-3B & 0.669 & 0.6766 & \textbf{0.683} & 0.6826 & 0.3 \\
Qwen2.5-1.5B & 0.6525 & 0.6625 & \textbf{0.6701} & 0.6688 & 0.2 \\
Qwen2.5-0.5B & 0.5624 & 0.5766 & \textbf{0.5842} & 0.583 & 0.5 \\
Gemma3-1B & 0.6465 & 0.6573 & \textbf{0.6646} & 0.6641 & 0.3 \\
Llama3.2-1B & 0.5988 & 0.6149 & \textbf{0.6217} & 0.6206 & 0.3 \\
Llama3.2-3B & 0.6718 & 0.6797 & \textbf{0.6857} & 0.6854 & 0.2 \\
Llama3.1-8B & 0.7031 & 0.7065 & \textbf{0.7093} & 0.7089 & 0.2 \\
FLAN-T5-Large & 0.5611 & 0.5857 & \textbf{0.5886} & 0.5873 & 0.2 \\
mT0-Large & 0.5863 & 0.6044 & \textbf{0.6098} & 0.6085 & 0.3 \\
\bottomrule
\end{tabular}
}
\end{table}

\begin{table}[H]
\caption{AUC-DF values values on WMT19 DE$\to$FR for token-level ensemble cascades. All cascades defer to Llama3.1-70B.}
\autosizeTable{
\begin{tabular}{lcccc}
\toprule
Ensemble & Chow-Sum & Chow-Avg & Best Chow-Quantile & Best Quantile \\
\midrule
Gemma3-1B, Qwen2.5-1.5B & 0.6702 & \textbf{0.6738} & 0.6729 & 0.1 \\
Gemma3-1B, Qwen2.5-1.5B, Llama3.2-1B & 0.6616 & 0.6613 & \textbf{0.6627} & 0.2 \\
Gemma3-1B, Qwen2.5-1.5B, mT0-Large & 0.6595 & \textbf{0.661} & 0.6606 & 0.2 \\
Gemma3-1B, Qwen2.5-1.5B, FLAN-T5-Large & \textbf{0.6552} & 0.6532 & 0.6538 & 0.1 \\
Qwen2.5-1.5B, Llama3.2-1B, mT0-Large & \textbf{0.6486} & 0.6445 & 0.6484 & 1.0 \\
Llama3.1-8B, Qwen2.5-7B & 0.7075 & \textbf{0.7096} & 0.7084 & 0.1 \\
Llama3.1-8B, Qwen2.5-3B & 0.6976 & 0.7002 & \textbf{0.7025} & 0.6 \\
Llama3.1-8B, Qwen2.5-7B, FLAN-T5-Large & \textbf{0.6801} & 0.6758 & 0.6758 & 0.2 \\
Llama3.1-8B, Qwen2.5-7B, mT0-Large & 0.6808 & 0.6839 & \textbf{0.7019} & 1.0 \\
\bottomrule
\end{tabular}
}
\end{table}

\begin{table}[H]
\caption{AUC-DF values on WMT19 DE$\to$FR for various semantic cascades. All cascades defer to Llama3.1-70B.}
\autosizeTable{
\begin{tabular}{lllllll}
\toprule
Ensemble & BLEU & ROUGE-1 & ROUGE-2 & ROUGE-L & BLEURT & SBERT \\
\midrule
\tworow{Gemma3-1B, Qwen2.5-1.5B, \\ Llama3.2-1B} & 0.6621 & 0.6641 & 0.6637 & 0.6654 & \textbf{0.6806} & 0.6686 \\
\tworow{Gemma3-1B, Qwen2.5-1.5B, \\ mT0-Large} & 0.6632 & 0.6664 & 0.6662 & 0.6667 & \textbf{0.6861} & 0.6705 \\
\tworow{Qwen2.5-1.5B, Gemma3-1B, \\ FLAN-T5-Large} & 0.6626 & 0.6645 & 0.6648 & 0.6651 & \textbf{0.6854} & 0.6717 \\
\tworow{Qwen2.5-1.5B, Llama3.2-1B, \\ mT0-Large} & 0.6493 & 0.6521 & 0.6499 & 0.6524 & \textbf{0.6747} & 0.6581 \\
\tworow{Gemma3-1B, Qwen2.5-1.5B, \\ Llama3.2-1B, mT0-Large} & 0.6618 & 0.6637 & 0.6642 & 0.6657 & \textbf{0.6877} & 0.6702 \\
\tworow{Gemma3-1B, Qwen2.5-1.5B, \\ mT0-Large, FLAN-T5-Large} & 0.6609 & 0.6637 & 0.6636 & 0.6649 & \textbf{0.6889} & 0.6711 \\
\tworow{Llama3.1-8B, Qwen2.5-7B, \\ FLAN-T5-Large} & 0.7046 & 0.7048 & 0.7052 & 0.7052 & \textbf{0.7141} & 0.7063 \\
\tworow{Llama3.1-8B, Qwen2.5-7B, \\ mT0-Large} & 0.7045 & 0.7054 & 0.7051 & 0.7048 & \textbf{0.716} & 0.7059 \\
\tworow{Llama3.1-8B, Qwen2.5-7B, \\ Qwen2.5-1.5B} & 0.7043 & 0.705 & 0.7052 & 0.705 & \textbf{0.7141} & 0.7064 \\
\tworow{Llama3.2-3B, Qwen2.5-3B, \\ Gemma3-1B} & 0.6835 & 0.6849 & 0.6848 & 0.6857 & \textbf{0.7005} & 0.6892 \\
\bottomrule
\end{tabular}
}
\end{table}

\subsubsection{WMT14 FR$\to$EN} \label{apdx:wmt_fr_en}
\begin{table}[H]
\caption{AUC-DF values values on WMT14 FR$\to$EN for token-level cascades. All cascades defer to Llama3.1-70B. The Random column represents the expected value if queries are deferred uniformly at random.}
\autosizeTable{
\begin{tabular}{lccccc}
\toprule
Model & Random & Chow-Sum & Chow-Avg & Best Chow-Quantile & Best Quantile \\
\midrule
Qwen2.5-7B & 0.7438 & 0.7454 & 0.7453 & \textbf{0.7454} & 0.0 \\
Qwen2.5-3B & 0.7382 & 0.739 & \textbf{0.7397} & 0.7397 & 0.1 \\
Qwen2.5-1.5B & 0.736 & 0.7383 & \textbf{0.7391} & 0.739 & 0.1 \\
Qwen2.5-0.5B & 0.7152 & 0.7193 & \textbf{0.7224} & 0.7216 & 0.1 \\
Gemma3-1B & 0.7278 & 0.7299 & 0.7316 & \textbf{0.7321} & 0.5 \\
Llama3.2-1B & 0.7114 & 0.715 & \textbf{0.7178} & 0.7176 & 0.6 \\
Llama3.2-3B & 0.7379 & 0.7395 & \textbf{0.7404} & 0.7402 & 0.1 \\
Llama3.1-8B & 0.7433 & 0.7439 & 0.7445 & \textbf{0.7445} & 0.1 \\
FLAN-T5-Large & 0.7174 & 0.7221 & 0.7268 & \textbf{0.7268} & 0.1 \\
mT0-Large & 0.7051 & 0.7102 & \textbf{0.7149} & 0.7143 & 0.1 \\
\bottomrule
\end{tabular}
}
\end{table}

\begin{table}[H]
\caption{AUC-DF values values on WMT14 FR$\to$EN for token-level ensemble cascades. All cascades defer to Llama3.1-70B.}
\autosizeTable{
\begin{tabular}{lcccc}
\toprule
Ensemble & Chow-Sum & Chow-Avg & Best Chow-Quantile & Best Quantile \\
\midrule
Gemma3-1B, Qwen2.5-1.5B & 0.7354 & 0.7359 & \textbf{0.7377} & 0.7 \\
Gemma3-1B, Qwen2.5-1.5B, Llama3.2-1B & 0.7303 & 0.7302 & \textbf{0.7306} & 0.0 \\
Gemma3-1B, Qwen2.5-1.5B, mT0-Large & 0.7301 & 0.73 & \textbf{0.731} & 0.2 \\
Gemma3-1B, Qwen2.5-1.5B, FLAN-T5-Large & 0.7348 & \textbf{0.7348} & 0.7348 & 0.2 \\
Qwen2.5-1.5B, Llama3.2-1B, mT0-Large & \textbf{0.7277} & 0.7261 & 0.7269 & 0.0 \\
Llama3.1-8B, Qwen2.5-7B & 0.7453 & 0.7452 & \textbf{0.7453} & 0.1 \\
Llama3.1-8B, Qwen2.5-3B & 0.7429 & 0.7431 & \textbf{0.7436} & 0.7 \\
Llama3.1-8B, Qwen2.5-7B, FLAN-T5-Large & \textbf{0.7425} & 0.7402 & 0.7414 & 0.1 \\
Llama3.1-8B, Qwen2.5-7B, mT0-Large & 0.7382 & 0.7364 & \textbf{0.7433} & 1.0 \\
\bottomrule
\end{tabular}
}
\end{table}

\begin{table}[H]
\caption{AUC-DF values on WMT14 FR$\to$EN for various semantic cascades. All cascades defer to Llama3.1-70B.}
\autosizeTable{
\begin{tabular}{lllllll}
\toprule
Ensemble & BLEU & ROUGE-1 & ROUGE-2 & ROUGE-L & BLEURT & SBERT \\
\midrule
\tworow{Gemma3-1B, Qwen2.5-1.5B, \\ Llama3.2-1B} & 0.7362 & 0.7362 & 0.7358 & 0.7364 & \textbf{0.7397} & 0.737 \\
\tworow{Gemma3-1B, Qwen2.5-1.5B, \\ mT0-Large} & 0.737 & 0.7372 & 0.7359 & 0.7373 & \textbf{0.7408} & 0.7379 \\
\tworow{Qwen2.5-1.5B, Gemma3-1B, \\ FLAN-T5-Large} & 0.7385 & 0.7382 & 0.7378 & 0.7387 & \textbf{0.742} & 0.7395 \\
\tworow{Qwen2.5-1.5B, Llama3.2-1B, \\ mT0-Large} & 0.7342 & 0.7346 & 0.7345 & 0.7347 & \textbf{0.7384} & 0.7359 \\
\tworow{Gemma3-1B, Qwen2.5-1.5B, \\ Llama3.2-1B, mT0-Large} & 0.7368 & 0.7368 & 0.7364 & 0.737 & \textbf{0.7406} & 0.7373 \\
\tworow{Gemma3-1B, Qwen2.5-1.5B, \\ mT0-Large, FLAN-T5-Large} & 0.7374 & 0.7375 & 0.7362 & 0.7374 & \textbf{0.7418} & 0.7388 \\
\tworow{Llama3.1-8B, Qwen2.5-7B, \\ FLAN-T5-Large} & 0.7454 & 0.7453 & 0.7448 & 0.7452 & \textbf{0.747} & 0.7458 \\
\tworow{Llama3.1-8B, Qwen2.5-7B, \\ mT0-Large} & 0.7448 & 0.7445 & 0.7442 & 0.7445 & \textbf{0.7464} & 0.7451 \\
\tworow{Llama3.1-8B, Qwen2.5-7B, \\ Qwen2.5-1.5B} & 0.7451 & 0.7449 & 0.7451 & 0.7453 & \textbf{0.7461} & 0.7454 \\
\tworow{Llama3.2-3B, Qwen2.5-3B, \\ Gemma3-1B} & 0.7403 & 0.7404 & 0.7399 & 0.7403 & \textbf{0.7431} & 0.7415 \\
\bottomrule
\end{tabular}
}
\end{table}

\subsubsection{WMT14 EN$\to$FR} \label{apdx:wmt_en_fr}
\begin{table}[H]
\caption{AUC-DF values values on WMT14 EN$\to$FR for token-level cascades. All cascades defer to Llama3.1-70B. The Random column represents the expected value if queries are deferred uniformly at random.}
\autosizeTable{
\begin{tabular}{lccccc}
\toprule
Model & Random & Chow-Sum & Chow-Avg & Best Chow-Quantile & Best Quantile \\
\midrule
Qwen2.5-7B & 0.6821 & 0.6897 & \textbf{0.6931} & 0.6921 & 0.1 \\
Qwen2.5-3B & 0.6676 & 0.6728 & \textbf{0.6784} & 0.6775 & 0.2 \\
Qwen2.5-1.5B & 0.6592 & 0.6669 & \textbf{0.672} & 0.6711 & 0.3 \\
Qwen2.5-0.5B & 0.5973 & 0.6116 & \textbf{0.6173} & 0.6164 & 0.5 \\
Gemma3-1B & 0.6654 & 0.6724 & \textbf{0.6754} & 0.6744 & 0.1 \\
Llama3.2-1B & 0.642 & 0.6549 & \textbf{0.6601} & 0.6592 & 0.2 \\
Llama3.2-3B & 0.6777 & 0.6833 & \textbf{0.6879} & 0.6873 & 0.2 \\
Llama3.1-8B & 0.693 & 0.6953 & 0.6979 & \textbf{0.698} & 0.6 \\
FLAN-T5-Large & 0.5802 & 0.6012 & \textbf{0.605} & 0.6039 & 0.1 \\
mT0-Large & 0.6095 & 0.6259 & \textbf{0.6305} & 0.6287 & 0.4 \\
\bottomrule
\end{tabular}
}
\end{table}

\begin{table}[H]
\caption{AUC-DF values values on WMT14 EN$\to$FR for token-level ensemble cascades. All cascades defer to Llama3.1-70B.}
\autosizeTable{
\begin{tabular}{lcccc}
\toprule
Ensemble & Chow-Sum & Chow-Avg & Best Chow-Quantile & Best Quantile \\
\midrule
Gemma3-1B, Qwen2.5-1.5B & 0.6772 & \textbf{0.6774} & 0.677 & 0.1 \\
Gemma3-1B, Qwen2.5-1.5B, Llama3.2-1B & \textbf{0.6766} & 0.6763 & 0.6764 & 0.1 \\
Gemma3-1B, Qwen2.5-1.5B, mT0-Large & \textbf{0.6697} & 0.6688 & 0.6672 & 0.1 \\
Gemma3-1B, Qwen2.5-1.5B, FLAN-T5-Large & 0.6619 & \textbf{0.6625} & 0.6594 & 0.1 \\
Qwen2.5-1.5B, Llama3.2-1B, mT0-Large & 0.6639 & \textbf{0.6644} & 0.6624 & 0.2 \\
Llama3.1-8B, Qwen2.5-7B & 0.6962 & \textbf{0.6982} & 0.6976 & 0.7 \\
Llama3.1-8B, Qwen2.5-3B & 0.6885 & 0.6911 & \textbf{0.6935} & 1.0 \\
Llama3.1-8B, Qwen2.5-7B, FLAN-T5-Large & \textbf{0.6741} & 0.6733 & 0.6729 & 0.1 \\
Llama3.1-8B, Qwen2.5-7B, mT0-Large & 0.6825 & 0.6831 & \textbf{0.6936} & 1.0 \\
\bottomrule
\end{tabular}
}
\end{table}

\begin{table}[H]
\caption{AUC-DF values on WMT14 EN$\to$FR for various semantic cascades. All cascades defer to Llama3.1-70B.}
\autosizeTable{
\begin{tabular}{lllllll}
\toprule
Ensemble & BLEU & ROUGE-1 & ROUGE-2 & ROUGE-L & BLEURT & SBERT \\
\midrule
\tworow{Gemma3-1B, Qwen2.5-1.5B, \\ Llama3.2-1B} & 0.6724 & 0.672 & 0.6725 & 0.6727 & \textbf{0.6835} & 0.6749 \\
\tworow{Gemma3-1B, Qwen2.5-1.5B, \\ mT0-Large} & 0.6712 & 0.6714 & 0.6709 & 0.6718 & \textbf{0.6861} & 0.6733 \\
\tworow{Qwen2.5-1.5B, Gemma3-1B, \\ FLAN-T5-Large} & 0.6673 & 0.6695 & 0.6688 & 0.6688 & \textbf{0.683} & 0.6717 \\
\tworow{Qwen2.5-1.5B, Llama3.2-1B, \\ mT0-Large} & 0.6649 & 0.6635 & 0.6648 & 0.6651 & \textbf{0.6787} & 0.6664 \\
\tworow{Gemma3-1B, Qwen2.5-1.5B, \\ Llama3.2-1B, mT0-Large} & 0.6736 & 0.6727 & 0.6737 & 0.6739 & \textbf{0.6878} & 0.6752 \\
\tworow{Gemma3-1B, Qwen2.5-1.5B, \\ mT0-Large, FLAN-T5-Large} & 0.6678 & 0.6705 & 0.6696 & 0.6706 & \textbf{0.6864} & 0.6709 \\
\tworow{Llama3.1-8B, Qwen2.5-7B, \\ FLAN-T5-Large} & 0.6938 & 0.6933 & 0.6939 & 0.6938 & \textbf{0.7003} & 0.6948 \\
\tworow{Llama3.1-8B, Qwen2.5-7B, \\ mT0-Large} & 0.6944 & 0.6931 & 0.6936 & 0.6932 & \textbf{0.7013} & 0.6938 \\
\tworow{Llama3.1-8B, Qwen2.5-7B, \\ Qwen2.5-1.5B} & 0.6929 & 0.6933 & 0.6934 & 0.6933 & \textbf{0.6996} & 0.6938 \\
\tworow{Llama3.2-3B, Qwen2.5-3B, \\ Gemma3-1B} & 0.6859 & 0.6865 & 0.6861 & 0.6867 & \textbf{0.6961} & 0.6887 \\
\bottomrule
\end{tabular}
}
\end{table}

\subsubsection{CNN/DailyMail} \label{apdx:cnn}
\begin{table}[H]
\caption{AUC-DF values values on CNN/DailyMail for token-level cascades. All cascades defer to Llama3.1-70B. The Random column represents the expected value if queries are deferred uniformly at random.}
\autosizeTable{
\begin{tabular}{lccccc}
\toprule
Model & Random & Chow-Sum & Chow-Avg & Best Chow-Quantile & Best Quantile \\
\midrule
Qwen2.5-7B & 0.258 & 0.2577 & 0.2605 & \textbf{0.2607} & 0.2 \\
Qwen2.5-3B & 0.2511 & \textbf{0.2539} & 0.2537 & 0.2537 & 0.5 \\
Qwen2.5-1.5B & 0.2474 & 0.2407 & 0.2536 & \textbf{0.2559} & 1.0 \\
Qwen2.5-0.5B & 0.2418 & 0.2371 & 0.245 & \textbf{0.2467} & 1.0 \\
Gemma3-1B & 0.2448 & 0.2454 & 0.2473 & \textbf{0.2479} & 0.4 \\
Llama3.2-1B & 0.2482 & 0.2502 & 0.2508 & \textbf{0.2511} & 0.3 \\
Llama3.2-3B & 0.2522 & 0.2535 & 0.2548 & \textbf{0.2549} & 0.1 \\
Llama3.1-8B & 0.2551 & 0.2547 & 0.2564 & \textbf{0.2565} & 0.2 \\
mT0-Large & 0.2527 & 0.2528 & 0.2539 & \textbf{0.2542} & 0.8 \\
\bottomrule
\end{tabular}
}
\end{table}

\begin{table}[H]
\caption{AUC-DF values values on CNN/DailyMail for token-level ensemble cascades. All cascades defer to Llama3.1-70B.}
\autosizeTable{
\begin{tabular}{lcccc}
\toprule
Ensemble & Chow-Sum & Chow-Avg & Best Chow-Quantile & Best Quantile \\
\midrule
Llama3.2-1B, mT0-Large & \textbf{0.2554} & 0.2529 & 0.2553 & 0.1 \\
Llama3.2-1B, Qwen2.5-1.5B & 0.2532 & 0.2493 & \textbf{0.2548} & 0.2 \\
Qwen2.5-1.5B, Gemma3-1B & 0.2505 & 0.2489 & \textbf{0.2559} & 1.0 \\
Llama3.2-1B, mT0-Large, Qwen2.5-1.5B & 0.2568 & 0.2491 & \textbf{0.257} & 0.8 \\
Qwen2.5-7B, Llama3.1-8B & 0.2579 & 0.259 & \textbf{0.2611} & 0.2 \\
Qwen2.5-7B, mT0-Large & 0.258 & 0.2575 & \textbf{0.2588} & 0.2 \\
Qwen2.5-7B, Qwen2.5-3B & 0.259 & 0.2583 & \textbf{0.2602} & 0.2 \\
\bottomrule
\end{tabular}
}
\end{table}

\begin{table}[H]
\caption{AUC-DF values on CNN/DailyMail for various semantic cascades. All cascades defer to Llama3.1-70B.}
\autosizeTable{
\begin{tabular}{lllllll}
\toprule
Ensemble & BLEU & ROUGE-1 & ROUGE-2 & ROUGE-L & BLEURT & SBERT \\
\midrule
\tworow{Llama3.2-1B, mT0-Large, \\ Qwen2.5-1.5B} & 0.2611 & \textbf{0.2616} & 0.2598 & 0.2608 & 0.2567 & 0.2603 \\
\tworow{Llama3.2-1B, mT0-Large, \\ Qwen2.5-0.5B} & 0.2566 & 0.2567 & 0.2553 & \textbf{0.2574} & 0.2513 & 0.2547 \\
\tworow{Qwen2.5-1.5B, Qwen2.5-0.5B, \\ Gemma3-1B} & \textbf{0.2551} & 0.2546 & 0.2514 & 0.2519 & 0.2522 & 0.2515 \\
\tworow{Llama3.2-1B, Gemma3-1B, \\ Qwen2.5-0.5B} & 0.2537 & 0.2538 & 0.2534 & \textbf{0.2543} & 0.251 & 0.2528 \\
\tworow{Llama3.2-3B, Qwen2.5-3B, \\ Qwen2.5-0.5B} & 0.2568 & \textbf{0.258} & 0.2578 & 0.2578 & 0.2546 & 0.2576 \\
\tworow{Qwen2.5-7B, mT0-Large, \\ Llama3.2-3B} & \textbf{0.2635} & 0.2618 & 0.2615 & 0.2607 & 0.2584 & 0.2618 \\
\tworow{Qwen2.5-7B, Llama3.1-8B, \\ mT0-Large} & 0.2622 & 0.2613 & \textbf{0.2622} & 0.2619 & 0.2586 & 0.2612 \\
\tworow{Llama3.1-8B, mT0-Large, \\ Qwen2.5-3B} & \textbf{0.2589} & 0.2571 & 0.2583 & 0.2578 & 0.2555 & 0.2585 \\
\bottomrule
\end{tabular}
}
\end{table}

\subsubsection{SQuAD1.1} \label{apdx:squad}
\begin{table}[H]
\caption{AUC-DF values on SQuAD1.1 for token-level cascades. All cascades defer to Llama-3.1-70B. The Random column represents the expected value if queries are deferred uniformly at random.}
\autosizeTable{
\begin{tabular}{lccccc}
\toprule
Model & Random & Chow-Sum & Chow-Avg & Best Chow-Quantile & Best Quantile \\
\midrule
Qwen2.5-7B & 0.8053 & \textbf{0.8217} & 0.8201 & 0.821 & 0.0 \\
Qwen2.5-3B & 0.7623 & 0.7916 & \textbf{0.7921} & 0.7915 & 0.1 \\
Qwen2.5-1.5B & 0.7593 & \textbf{0.7897} & 0.7847 & 0.7868 & 0.0 \\
Qwen2.5-0.5B & 0.642 & \textbf{0.6898} & 0.6826 & 0.685 & 0.0 \\
Llama3.1-8B & 0.7987 & \textbf{0.8173} & 0.8151 & 0.8169 & 0.0 \\
Llama3.2-3B & 0.7627 & 0.7764 & 0.7791 & \textbf{0.7833} & 0.5 \\
Llama3.2-1B & 0.631 & 0.6695 & \textbf{0.677} & 0.6768 & 0.3 \\
mT0-Large & 0.802 & 0.8205 & 0.8211 & \textbf{0.8211} & 0.1 \\
FLAN-T5-Large & 0.7123 & \textbf{0.7598} & 0.6896 & 0.7437 & 0.0 \\
Gemma3-1B & 0.676 & \textbf{0.6984} & 0.6938 & 0.696 & 0.0 \\
\bottomrule
\end{tabular}
}
\end{table}

\begin{table}[H]
\caption{AUC-DF values values on SQuAD1.1 for token-level ensemble cascades. All cascades defer to Llama-3.1-70B.}
\autosizeTable{
\begin{tabular}{lcccc}
\toprule
Ensemble & Chow-Sum & Chow-Avg & Best Chow-Quantile & Best Quantile \\
\midrule
mT0-Large, Qwen2.5-1.5B & 0.8155 & \textbf{0.8168} & 0.8125 & 0.0 \\
mT0-Large, Qwen2.5-1.5B, FLAN-T5-Large & 0.7809 & \textbf{0.8081} & 0.8043 & 0.0 \\
Qwen2.5-1.5B, FLAN-T5-Large, Gemma3-1B & 0.7414 & 0.7511 & \textbf{0.7632} & 0.0 \\
Qwen2.5-7B, Llama3.1-8B & 0.8211 & 0.8211 & \textbf{0.8219} & 0.0 \\
Qwen2.5-7B, Llama3.1-8B, mT0-Large & 0.8236 & 0.8239 & \textbf{0.8255} & 0.0 \\
Qwen2.5-7B, Llama3.2-3B & \textbf{0.7993} & 0.7985 & 0.7988 & 0.0 \\
Qwen2.5-7B, Llama3.2-3B, Gemma3-1B & 0.7599 & 0.7461 & \textbf{0.7764} & 1.0 \\
Qwen2.5-7B, mT0-Large, FLAN-T5-Large & 0.7841 & \textbf{0.8165} & 0.8162 & 0.0 \\
\bottomrule
\end{tabular}
}
\end{table}

\begin{table}[H]
\caption{AUC-DF values on SQuAD1.1 for various semantic cascades. All cascades defer to Llama-3.1-70B.}
\autosizeTable{
\begin{tabular}{cllllll}
\toprule
Ensemble & BLEU & ROUGE-1 & ROUGE-2 & ROUGE-L & BLEURT & SBERT \\
\midrule
\tworow{mT0 Large, Qwen2.5-1.5B, \\ FLAN-T5 Large} & 0.8235 & 0.8241 & 0.8128 & \textbf{0.8247} & 0.8193 & 0.8235 \\
\tworow{mT0 Large, Qwen2.5-1.5B, \\ FLAN-T5 Large, Llama-3.2-1B} & 0.8169 & 0.822 & 0.8146 & \textbf{0.823} & 0.8143 & 0.8144 \\
\tworow{Qwen2.5-1.5B, FLAN-T5 Large, \\ Gemma3-1B} & 0.7889 & 0.7938 & 0.778 & \textbf{0.7944} & 0.7885 & 0.789 \\
\tworow{Qwen2.5-1.5B, FLAN-T5 Large, \\ Gemma3-1B, Llama-3.2-1B} & 0.7809 & \textbf{0.7916} & 0.7757 & 0.7915 & 0.788 & 0.7814 \\
\tworow{Qwen2.5-1.5B, FLAN-T5-Large, \\ Llama3.2-1B} & 0.7899 & 0.7937 & 0.7762 & \textbf{0.7944} & 0.7889 & 0.792 \\
\tworow{Llama3.2-3B, Qwen2.5-3B, \\ FLAN-T5-Large} & 0.8115 & 0.8151 & 0.8057 & 0.8151 & \textbf{0.8168} & 0.812 \\
\tworow{Qwen2.5-7B, mT0-Large, \\ FLAN-T5-Large} & 0.8333 & 0.8325 & 0.8253 & 0.833 & 0.8334 & \textbf{0.8353} \\
\tworow{Qwen2.5-7B, mT0-Large, \\ Llama3.2-3B} & 0.8272 & \textbf{0.8314} & 0.8248 & 0.831 & 0.8299 & 0.8297 \\
\tworow{Qwen2.5-7B, Llama3.1-8B, \\ mT0-Large} & 0.8315 & 0.8312 & 0.8262 & 0.8302 & \textbf{0.8324} & 0.8321 \\
\tworow{Llama3.1-8B, mT0-Large, \\ Qwen2.5-1.5B} & 0.8259 & 0.8254 & 0.8192 & 0.8251 & \textbf{0.827} & 0.8262 \\
\bottomrule
\end{tabular}
}
\end{table}

\subsubsection{TriviaQA} \label{apdx:triviaqa}
\begin{table}[H]
\caption{AUC-DF values values on TriviaQA for token-level cascades. All cascades defer to Llama3.1-70B. The Random column represents the expected value if queries are deferred uniformly at random.}
\autosizeTable{
\begin{tabular}{lccccc}
\toprule
Model & Random & Chow-Sum & Chow-Avg & Best Chow-Quantile & Best Quantile \\
\midrule
Qwen2.5-7B & 0.6575 & \textbf{0.7258} & 0.7238 & 0.7242 & 0.1 \\
Qwen2.5-3B & 0.6003 & \textbf{0.6684} & 0.6662 & 0.6664 & 0.1 \\
Qwen2.5-1.5B & 0.5556 & \textbf{0.6168} & 0.6147 & 0.6154 & 0.1 \\
Qwen2.5-0.5B & 0.4569 & \textbf{0.4832} & 0.4735 & 0.4812 & 0.0 \\
Gemma3-1B & 0.5063 & \textbf{0.5452} & 0.544 & 0.5442 & 0.0 \\
Llama3.2-1B & 0.5265 & \textbf{0.5745} & 0.5708 & 0.572 & 0.1 \\
Llama3.2-3B & 0.6584 & \textbf{0.7202} & 0.7183 & 0.7193 & 0.1 \\
Llama3.1-8B & 0.7303 & \textbf{0.7741} & 0.7711 & 0.7714 & 0.1 \\
FLAN-T5-Large & 0.4761 & 0.5092 & 0.5019 & \textbf{0.5102} & 0.0 \\
mT0-Large & 0.4337 & 0.4464 & 0.4436 & \textbf{0.4479} & 0.2 \\
\bottomrule
\end{tabular}
}
\end{table}

\begin{table}[H]
\caption{AUC-DF values values on TriviaQA for token-level ensemble cascades. All cascades defer to Llama3.1-70B.}
\autosizeTable{
\begin{tabular}{lcccc}
\toprule
Ensemble & Chow-Sum & Chow-Avg & Best Chow-Quantile & Best Quantile \\
\midrule
Qwen2.5-1.5B, Llama3.2-1B & 0.612 & \textbf{0.6183} & 0.6155 & 0.1 \\
Llama3.2-1B, Gemma3-1B & 0.5763 & \textbf{0.5788} & 0.5782 & 0.1 \\
Qwen2.5-1.5B, Llama3.2-1B, Gemma3-1B & 0.6079 & \textbf{0.6114} & 0.6111 & 0.1 \\
Qwen2.5-1.5B, Llama3.2-1B, FLAN-T5-Large & 0.5917 & \textbf{0.6002} & 0.6 & 0.0 \\
Llama3.1-8B, Qwen2.5-7B & 0.7596 & 0.7534 & \textbf{0.763} & 0.0 \\
Llama3.1-8B, Qwen2.5-7B, mT0-Large & 0.6443 & 0.6188 & \textbf{0.6618} & 0.0 \\
Qwen2.5-7B, Llama3.2-3B & 0.7402 & \textbf{0.7434} & 0.7421 & 0.1 \\
Qwen2.5-7B, Llama3.2-3B, Gemma3-1B & 0.689 & 0.6737 & \textbf{0.6939} & 0.0 \\
Qwen2.5-7B, Gemma3-1B & 0.6433 & 0.6429 & \textbf{0.6493} & 0.0 \\
Llama3.2-3B, Qwen2.5-3B & 0.7128 & 0.7119 & \textbf{0.7144} & 0.1 \\
Llama3.1-8B, Llama3.2-3B, Qwen2.5-3B & 0.7398 & 0.731 & \textbf{0.7455} & 0.1 \\
\bottomrule
\end{tabular}
}
\end{table}

\begin{table}[H]
\caption{AUC-DF values on TriviaQA for various semantic cascades. All cascades defer to Llama3.1-70B.}
\autosizeTable{
\begin{tabular}{lllllll}
\toprule
Ensemble & BLEU & ROUGE-1 & ROUGE-2 & ROUGE-L & BLEURT & SBERT \\
\midrule
\tworow{Qwen2.5-1.5B, Llama3.2-1B, \\ Gemma3-1B} & 0.6 & 0.6054 & 0.5699 & \textbf{0.6056} & 0.5966 & 0.5934 \\
\tworow{Qwen2.5-1.5B, Llama3.2-1B, \\ FLAN-T5-Large} & 0.5897 & 0.5963 & 0.565 & \textbf{0.5965} & 0.5849 & 0.5838 \\
\tworow{Llama3.2-1B, Gemma3-1B, \\ FLAN-T5-Large} & 0.5589 & \textbf{0.5656} & 0.5342 & 0.5652 & 0.5585 & 0.5551 \\
\tworow{Qwen2.5-1.5B, Llama3.2-1B, \\ Gemma3-1B, FLAN-T5-Large} & 0.5984 & \textbf{0.6027} & 0.5676 & 0.6026 & 0.5963 & 0.5885 \\
\tworow{Llama3.1-8B, Qwen2.5-7B, \\ Llama3.2-3B} & 0.7684 & 0.7718 & 0.744 & \textbf{0.772} & 0.7653 & 0.7584 \\
\tworow{Qwen2.5-7B, Llama3.2-3B, \\ Gemma3-1B} & 0.7169 & 0.7211 & 0.6823 & \textbf{0.7216} & 0.7146 & 0.7078 \\
\tworow{Qwen2.5-7B, Llama3.2-3B, \\ Qwen2.5-3B} & 0.729 & \textbf{0.7343} & 0.6881 & 0.7342 & 0.7322 & 0.7259 \\
\tworow{Llama3.1-8B, Llama3.2-3B, \\ Qwen2.5-3B} & 0.7622 & 0.7653 & 0.7422 & \textbf{0.7653} & 0.7551 & 0.7482 \\
\tworow{Llama3.1-8B, Qwen2.5-7B, \\ mT0-Large} & 0.7525 & 0.755 & 0.7408 & \textbf{0.7551} & 0.7319 & 0.7194 \\
\bottomrule
\end{tabular}
}
\end{table}

\subsection{Latencies at 98\% Target Model Performance}

\subsubsection{WMT19 DE$\to$FR} 
\begin{table}[H]
\caption{Latency (ms) at 98\% target model BLEURT values on WMT19 DE$\to$FR for token-level cascades. All cascades defer to Llama3.1-70B. The Random column represents the expected value if queries are deferred uniformly at random.}
\autosizeTable{
\begin{tabular}{lccccc}
\toprule
Model & Random & Chow-Sum & Chow-Avg & Best Chow-Quantile & Best Quantile \\
\midrule
Qwen2.5-7B & 3136 & 2495 & \textbf{2271} & 2310 & 0.1 \\
Qwen2.5-3B & 2725 & 2384 & 2258 & \textbf{2244} & 0.3 \\
Qwen2.5-1.5B & 2541 & 2268 & \textbf{2093} & 2127 & 0.2 \\
Qwen2.5-0.5B & 2416 & 2331 & \textbf{2300} & \textbf{2300} & 0.2 \\
Gemma3-1B & 2480 & 2233 & 2147 & \textbf{2126} & 0.2 \\
Llama3.2-1B & 2451 & 2316 & 2296 & \textbf{2283} & 0.0 \\
Llama3.2-3B & 2680 & 2313 & \textbf{2134} & 2151 & 0.2 \\
Llama3.1-8B & 3177 & 2278 & 2097 & \textbf{2055} & 0.2 \\
FLAN-T5-Large & 2469 & 2340 & 2340 & \textbf{2339} & 0.0 \\
mT0-Large & 2626 & 2484 & \textbf{2443} & 2449 & 0.3 \\
\bottomrule
\end{tabular}
}
\end{table}

\begin{table}[H]
\caption{Latency (ms) at 98\% target model BLEURT values on WMT19 DE$\to$FR for token-level ensemble cascades. All cascades defer to Llama3.1-70B.}
\autosizeTable{
\begin{tabular}{lcccc}
\toprule
Ensemble & Chow-Sum & Chow-Avg & Best Chow-Quantile & Best Quantile \\
\midrule
Gemma3-1B, Qwen2.5-1.5B & 2256 & \textbf{2102} & 2125 & 0.1 \\
Gemma3-1B, Qwen2.5-1.5B, Llama3.2-1B & 2239 & \textbf{2217} & 2220 & 0.1 \\
Gemma3-1B, Qwen2.5-1.5B, mT0-Large & 2370 & \textbf{2290} & 2300 & 0.1 \\
Gemma3-1B, Qwen2.5-1.5B, FLAN-T5-Large & 2273 & 2259 & \textbf{2236} & 0.1 \\
Qwen2.5-1.5B, Llama3.2-1B, mT0-Large & 2395 & 2386 & \textbf{2383} & 0.2 \\
Llama3.1-8B, Qwen2.5-7B & 2246 & \textbf{2117} & 2129 & 0.2 \\
Llama3.1-8B, Qwen2.5-3B & 2620 & 2388 & \textbf{2362} & 0.6 \\
Llama3.1-8B, Qwen2.5-7B, FLAN-T5-Large & \textbf{2745} & 2773 & 2761 & 0.1 \\
Llama3.1-8B, Qwen2.5-7B, mT0-Large & 2773 & 2696 & \textbf{2376} & 1.0 \\
\bottomrule
\end{tabular}
}
\end{table}

\begin{table}[H]
\caption{Latency (ms) at 98\% target model BLEURT on WMT19 DE$\to$FR for various semantic cascades. All cascades defer to Llama3.1-70B.}
\autosizeTable{
\begin{tabular}{lllllll}
\toprule
Ensemble & BLEU & ROUGE-1 & ROUGE-2 & ROUGE-L & BLEURT & SBERT \\
\midrule
\tworow{Gemma3-1B, Qwen2.5-1.5B, \\ Llama3.2-1B} & 2223 & 2239 & 2245 & 2239 & \textbf{2001} & 2215 \\
\tworow{Gemma3-1B, Qwen2.5-1.5B, \\ mT0-Large} & 2286 & 2296 & 2293 & 2280 & \textbf{2077} & 2282 \\
\tworow{Qwen2.5-1.5B, Gemma3-1B, \\ FLAN-T5-Large} & 2223 & 2237 & 2197 & 2187 & \textbf{1931} & 2156 \\
\tworow{Qwen2.5-1.5B, Llama3.2-1B, \\ mT0-Large} & 2359 & 2342 & 2361 & 2350 & \textbf{2162} & 2324 \\
\tworow{Gemma3-1B, Qwen2.5-1.5B, \\ Llama3.2-1B, mT0-Large} & 2272 & 2308 & 2293 & 2293 & \textbf{2033} & 2257 \\
\tworow{Gemma3-1B, Qwen2.5-1.5B, \\ mT0-Large, FLAN-T5-Large} & 2294 & 2276 & 2279 & 2251 & \textbf{1996} & 2238 \\
\tworow{Llama3.1-8B, Qwen2.5-7B, \\ FLAN-T5-Large} & 2336 & 2343 & 2233 & 2297 & \textbf{1737} & 2252 \\
\tworow{Llama3.1-8B, Qwen2.5-7B, \\ mT0-Large} & 2336 & 2315 & 2376 & 2329 & \textbf{1557} & 2301 \\
\tworow{Llama3.1-8B, Qwen2.5-7B, \\ Qwen2.5-1.5B} & 2311 & 2286 & 2275 & 2252 & \textbf{1693} & 2171 \\
\tworow{Llama3.2-3B, Qwen2.5-3B, \\ Gemma3-1B} & 2218 & 2225 & 2211 & 2188 & \textbf{1856} & 2131 \\
\bottomrule
\end{tabular}
}
\end{table}

\subsubsection{WMT14 FR$\to$EN} 
\begin{table}[H]
\caption{Latency (ms) at 98\% target model BLEURT values on WMT14 FR$\to$EN for token-level cascades. All cascades defer to Llama3.1-70B. The Random column represents the expected value if queries are deferred uniformly at random.}
\autosizeTable{
\begin{tabular}{lccccc}
\toprule
Model & Random & Chow-Sum & Chow-Avg & Best Chow-Quantile & Best Quantile \\
\midrule
Qwen2.5-7B & 1767 & \textbf{499.4} & \textbf{499.4} & \textbf{499.4} & 0.0 \\
Qwen2.5-3B & 1530 & 394 & 318.3 & \textbf{313} & 1.0 \\
Qwen2.5-1.5B & 1428 & 436.9 & \textbf{324.2} & 335.7 & 0.1 \\
Qwen2.5-0.5B & 1346 & 951 & \textbf{885.9} & 906.1 & 0.2 \\
Gemma3-1B & 1396 & 811 & \textbf{716.8} & 722.1 & 0.5 \\
Llama3.2-1B & 1373 & 1076 & 1077 & \textbf{1014} & 0.0 \\
Llama3.2-3B & 1505 & 357.4 & 309.9 & \textbf{283.5} & 0.6 \\
Llama3.1-8B & 1784 & \textbf{516.1} & \textbf{516.1} & \textbf{516.1} & 0.0 \\
FLAN-T5-Large & 1370 & 943.5 & \textbf{787.7} & 806.2 & 0.2 \\
mT0-Large & 1495 & 1208 & \textbf{1111} & 1130 & 0.1 \\
\bottomrule
\end{tabular}
}
\end{table}

\begin{table}[H]
\caption{Latency (ms) at 98\% target model BLEURT values on WMT14 FR$\to$EN for token-level ensemble cascades. All cascades defer to Llama3.1-70B.}
\autosizeTable{
\begin{tabular}{lcccc}
\toprule
Ensemble & Chow-Sum & Chow-Avg & Best Chow-Quantile & Best Quantile \\
\midrule
Gemma3-1B, Qwen2.5-1.5B & 597.1 & 575.1 & \textbf{416.7} & 0.8 \\
Gemma3-1B, Qwen2.5-1.5B, Llama3.2-1B & \textbf{809.2} & 866.5 & 835.7 & 0.0 \\
Gemma3-1B, Qwen2.5-1.5B, mT0-Large & 890 & 852.2 & \textbf{826.7} & 0.2 \\
Gemma3-1B, Qwen2.5-1.5B, FLAN-T5-Large & 643.8 & \textbf{636.7} & 653.4 & 0.3 \\
Qwen2.5-1.5B, Llama3.2-1B, mT0-Large & \textbf{916.4} & 988.6 & 986 & 0.0 \\
Llama3.1-8B, Qwen2.5-7B & \textbf{516.1} & \textbf{516.1} & \textbf{516.1} & 0.0 \\
Llama3.1-8B, Qwen2.5-3B & \textbf{516.1} & \textbf{516.1} & \textbf{516.1} & 0.0 \\
Llama3.1-8B, Qwen2.5-7B, FLAN-T5-Large & \textbf{516.1} & 565.4 & \textbf{516.1} & 0.2 \\
Llama3.1-8B, Qwen2.5-7B, mT0-Large & 724.7 & 883.2 & \textbf{516.1} & 1.0 \\
\bottomrule
\end{tabular}
}
\end{table}

\begin{table}[H]
\caption{Latency (ms) at 98\% target model BLEURT on WMT14 FR$\to$EN for various semantic cascades. All cascades defer to Llama3.1-70B.}
\autosizeTable{
\begin{tabular}{lllllll}
\toprule
Ensemble & BLEU & ROUGE-1 & ROUGE-2 & ROUGE-L & BLEURT & SBERT \\
\midrule
\tworow{Gemma3-1B, Qwen2.5-1.5B, \\ Llama3.2-1B} & 575.1 & 552.2 & 556.6 & 546.9 & \textbf{309.3} & 519.6 \\
\tworow{Gemma3-1B, Qwen2.5-1.5B, \\ mT0-Large} & 590.8 & 527.4 & 617.2 & 503.6 & \textbf{324.9} & 473.7 \\
\tworow{Qwen2.5-1.5B, Gemma3-1B, \\ FLAN-T5-Large} & 437.8 & 414 & 443.1 & 399 & \textbf{194.8} & 358.6 \\
\tworow{Qwen2.5-1.5B, Llama3.2-1B, \\ mT0-Large} & 737.8 & 644.4 & 725.4 & 678.8 & \textbf{479} & 647.1 \\
\tworow{Gemma3-1B, Qwen2.5-1.5B, \\ Llama3.2-1B, mT0-Large} & 610.1 & 574.9 & 622.4 & 551.1 & \textbf{324.9} & 572.3 \\
\tworow{Gemma3-1B, Qwen2.5-1.5B, \\ mT0-Large, FLAN-T5-Large} & 575.8 & 511.5 & 596.9 & 519.5 & \textbf{238.7} & 466.6 \\
\tworow{Llama3.1-8B, Qwen2.5-7B, \\ FLAN-T5-Large} & \textbf{516.1} & \textbf{516.1} & \textbf{516.1} & \textbf{516.1} & \textbf{516.1} & \textbf{516.1} \\
\tworow{Llama3.1-8B, Qwen2.5-7B, \\ mT0-Large} & \textbf{516.1} & \textbf{516.1} & \textbf{516.1} & \textbf{516.1} & \textbf{516.1} & \textbf{516.1} \\
\tworow{Llama3.1-8B, Qwen2.5-7B, \\ Qwen2.5-1.5B} & \textbf{516.1} & \textbf{516.1} & \textbf{516.1} & \textbf{516.1} & \textbf{516.1} & \textbf{516.1} \\
\tworow{Llama3.2-3B, Qwen2.5-3B, \\ Gemma3-1B} & 290.1 & 345.5 & 326.2 & 319.1 & \textbf{262.8} & \textbf{262.8} \\
\bottomrule
\end{tabular}
}
\end{table}

\subsubsection{WMT14 EN$\to$FR} 
\begin{table}[H]
\caption{Latency (ms) at 98\% target model BLEURT values on WMT14 EN$\to$FR for token-level cascades. All cascades defer to Llama3.1-70B. The Random column represents the expected value if queries are deferred uniformly at random.}
\autosizeTable{
\begin{tabular}{lccccc}
\toprule
Model & Random & Chow-Sum & Chow-Avg & Best Chow-Quantile & Best Quantile \\
\midrule
Qwen2.5-7B & 2583 & 1672 & \textbf{1460} & 1556 & 0.1 \\
Qwen2.5-3B & 2229 & 1817 & \textbf{1659} & 1696 & 0.1 \\
Qwen2.5-1.5B & 2083 & 1690 & \textbf{1547} & 1590 & 0.1 \\
Qwen2.5-0.5B & 1979 & 1845 & 1797 & \textbf{1794} & 0.5 \\
Gemma3-1B & 2051 & 1630 & \textbf{1489} & 1536 & 0.1 \\
Llama3.2-1B & 2007 & 1673 & \textbf{1581} & 1584 & 0.2 \\
Llama3.2-3B & 2205 & 1578 & \textbf{1347} & 1356 & 0.1 \\
Llama3.1-8B & 2604 & 1528 & 1215 & \textbf{1166} & 0.6 \\
FLAN-T5-Large & 2037 & 1902 & 1897 & \textbf{1864} & 0.1 \\
mT0-Large & 2170 & 1959 & \textbf{1912} & 1918 & 0.4 \\
\bottomrule
\end{tabular}
}
\end{table}

\begin{table}[H]
\caption{Latency (ms) at 98\% target model BLEURT values on WMT14 EN$\to$FR for token-level ensemble cascades. All cascades defer to Llama3.1-70B.}
\autosizeTable{
\begin{tabular}{lcccc}
\toprule
Ensemble & Chow-Sum & Chow-Avg & Best Chow-Quantile & Best Quantile \\
\midrule
Gemma3-1B, Qwen2.5-1.5B & 1594 & 1541 & \textbf{1509} & 0.1 \\
Gemma3-1B, Qwen2.5-1.5B, Llama3.2-1B & 1608 & 1545 & \textbf{1533} & 0.1 \\
Gemma3-1B, Qwen2.5-1.5B, mT0-Large & 1745 & 1659 & \textbf{1606} & 0.1 \\
Gemma3-1B, Qwen2.5-1.5B, FLAN-T5-Large & 1662 & 1622 & \textbf{1538} & 0.1 \\
Qwen2.5-1.5B, Llama3.2-1B, mT0-Large & 1776 & 1682 & \textbf{1629} & 0.1 \\
Llama3.1-8B, Qwen2.5-7B & 1661 & 1477 & \textbf{1442} & 0.1 \\
Llama3.1-8B, Qwen2.5-3B & 1901 & 1681 & \textbf{1677} & 0.2 \\
Llama3.1-8B, Qwen2.5-7B, FLAN-T5-Large & 1682 & 1463 & \textbf{1412} & 0.1 \\
Llama3.1-8B, Qwen2.5-7B, mT0-Large & 1735 & 1477 & \textbf{1454} & 0.1 \\
\bottomrule
\end{tabular}
}
\end{table}

\begin{table}[H]
\caption{Latency (ms) at 98\% target model BLEURT on WMT14 EN$\to$FR for various semantic cascades. All cascades defer to Llama3.1-70B.}
\autosizeTable{
\begin{tabular}{lllllll}
\toprule
Ensemble & BLEU & ROUGE-1 & ROUGE-2 & ROUGE-L & BLEURT & SBERT \\
\midrule
\tworow{Gemma3-1B, Qwen2.5-1.5B, \\ Llama3.2-1B} & 1564 & 1594 & 1571 & 1568 & \textbf{1309} & 1538 \\
\tworow{Gemma3-1B, Qwen2.5-1.5B, \\ mT0-Large} & 1719 & 1696 & 1706 & 1705 & \textbf{1331} & 1666 \\
\tworow{Qwen2.5-1.5B, Gemma3-1B, \\ FLAN-T5-Large} & 1660 & 1645 & 1661 & 1670 & \textbf{1359} & 1617 \\
\tworow{Qwen2.5-1.5B, Llama3.2-1B, \\ mT0-Large} & 1725 & 1770 & 1749 & 1774 & \textbf{1503} & 1713 \\
\tworow{Gemma3-1B, Qwen2.5-1.5B, \\ Llama3.2-1B, mT0-Large} & 1620 & 1661 & 1625 & 1642 & \textbf{1330} & 1639 \\
\tworow{Gemma3-1B, Qwen2.5-1.5B, \\ mT0-Large, FLAN-T5-Large} & 1715 & 1727 & 1735 & 1744 & \textbf{1339} & 1691 \\
\tworow{Llama3.1-8B, Qwen2.5-7B, \\ FLAN-T5-Large} & 1447 & 1575 & 1482 & 1586 & \textbf{1025} & 1460 \\
\tworow{Llama3.1-8B, Qwen2.5-7B, \\ mT0-Large} & 1475 & 1695 & 1536 & 1666 & \textbf{861.4} & 1597 \\
\tworow{Llama3.1-8B, Qwen2.5-7B, \\ Qwen2.5-1.5B} & 1580 & 1550 & 1592 & 1572 & \textbf{1036} & 1530 \\
\tworow{Llama3.2-3B, Qwen2.5-3B, \\ Gemma3-1B} & 1507 & 1456 & 1434 & 1465 & \textbf{962.2} & 1375 \\
\bottomrule
\end{tabular}
}
\end{table}

\subsubsection{CNN/DailyMail}
\begin{table}[H]
\caption{Latency (ms) at 98\% target model ROUGE-L values on CNN/DailyMail for token-level cascades. All cascades defer to Llama3.1-70B. The Random column represents the expected value if queries are deferred uniformly at random.}
\autosizeTable{
\begin{tabular}{lccccc}
\toprule
Model & Random & Chow-Sum & Chow-Avg & Best Chow-Quantile & Best Quantile \\
\midrule
Qwen2.5-7B & 7824 & 5787 & 5123 & \textbf{4705} & 0.2 \\
Qwen2.5-3B & 7217 & 5982 & 5803 & \textbf{5318} & 0.5 \\
Qwen2.5-1.5B & 7095 & 6790 & 5196 & \textbf{4639} & 1.0 \\
Qwen2.5-0.5B & 6682 & 6609 & 6098 & \textbf{5752} & 0.9 \\
Gemma3-1B & 6954 & 6264 & 5959 & \textbf{5779} & 0.3 \\
Llama3.2-1B & 6890 & 5808 & 5489 & \textbf{5356} & 0.2 \\
Llama3.2-3B & 7585 & 6251 & 6284 & \textbf{5574} & 0.1 \\
Llama3.1-8B & 8936 & 7947 & 7701 & \textbf{6818} & 0.0 \\
mT0-Large & 6964 & 6062 & 5677 & \textbf{5444} & 0.2 \\
\bottomrule
\end{tabular}
}
\end{table}

\begin{table}[H]
\caption{Latency (ms) at 98\% target model ROUGE-L values on CNN/DailyMail for token-level ensemble cascades. All cascades defer to Llama3.1-70B.}
\autosizeTable{
\begin{tabular}{lcccc}
\toprule
Ensemble & Chow-Sum & Chow-Avg & Best Chow-Quantile & Best Quantile \\
\midrule
Llama3.2-1B, mT0-Large & 5511 & 5650 & \textbf{5172} & 0.1 \\
Llama3.2-1B, Qwen2.5-1.5B & 5714 & 6318 & \textbf{5462} & 0.2 \\
Qwen2.5-1.5B, Gemma3-1B & 6670 & 5827 & \textbf{4626} & 1.0 \\
Llama3.2-1B, mT0-Large, Qwen2.5-1.5B & \textbf{5103} & 6564 & 5402 & 0.2 \\
Qwen2.5-7B, Llama3.1-8B & 7011 & 6692 & \textbf{5763} & 0.2 \\
Qwen2.5-7B, mT0-Large & \textbf{5733} & 6291 & 5806 & 0.4 \\
Qwen2.5-7B, Qwen2.5-3B & 5541 & 5621 & \textbf{4751} & 0.4 \\
\bottomrule
\end{tabular}
}
\end{table}

\begin{table}[H]
\caption{Latency (ms) at 98\% target model ROUGE-L on CNN/DailyMail for various semantic cascades. All cascades defer to Llama3.1-70B.}
\autosizeTable{
\begin{tabular}{lllllll}
\toprule
Ensemble & BLEU & ROUGE-1 & ROUGE-2 & ROUGE-L & BLEURT & SBERT \\
\midrule
\tworow{Llama3.2-1B, mT0-Large, \\ Qwen2.5-1.5B} & 3816 & \textbf{3245} & 4174 & 4055 & 5070 & 4187 \\
\tworow{Llama3.2-1B, mT0-Large, \\ Qwen2.5-0.5B} & 5039 & \textbf{4873} & 4986 & 4999 & 6254 & 5570 \\
\tworow{Qwen2.5-1.5B, Qwen2.5-0.5B, \\ Gemma3-1B} & \textbf{5422} & 5834 & 6551 & 6444 & 6199 & 6571 \\
\tworow{Llama3.2-1B, Gemma3-1B, \\ Qwen2.5-0.5B} & 5262 & 5149 & 5235 & \textbf{4790} & 5799 & 5096 \\
\tworow{Llama3.2-3B, Qwen2.5-3B, \\ Qwen2.5-0.5B} & 5733 & \textbf{5242} & 5408 & 5448 & 6444 & 5322 \\
\tworow{Qwen2.5-7B, mT0-Large, \\ Llama3.2-3B} & \textbf{4041} & 4744 & 4691 & 4997 & 6125 & 4525 \\
\tworow{Qwen2.5-7B, Llama3.1-8B, \\ mT0-Large} & \textbf{5059} & 6022 & 5557 & 5431 & 7310 & 5803 \\
\tworow{Llama3.1-8B, mT0-Large, \\ Qwen2.5-3B} & \textbf{6633} & 7257 & 6945 & 6925 & 7575 & 6745 \\
\bottomrule
\end{tabular}
}
\end{table}

\subsubsection{SQuAD1.1}
\begin{table}[H]
\caption{Latency (ms) at 98\% target model accuracy values on SQuAD1.1 for token-level cascades. All cascades defer to Llama3.1-70B. The Random column represents the expected value if queries are deferred uniformly at random.}
\autosizeTable{
\begin{tabular}{lccccc}
\toprule
Model & Random & Chow-Sum & Chow-Avg & Best Chow-Quantile & Best Quantile \\
\midrule
Qwen2.5-7B & 557.1 & 215 & 221.3 & \textbf{212.3} & 0.1 \\
Qwen2.5-3B & 464.1 & 302.8 & 298.4 & \textbf{297} & 0.1 \\
Qwen2.5-1.5B & 442.5 & \textbf{281.2} & 300.6 & 296.2 & 0.1 \\
Qwen2.5-0.5B & 419.5 & 365.1 & \textbf{364.3} & 368.9 & 0.1 \\
Gemma3-1B & 435.2 & 411.4 & 409.2 & \textbf{405.1} & 0.4 \\
Llama3.2-1B & 419.2 & 385.2 & 375.4 & \textbf{372.7} & 0.3 \\
Llama3.2-3B & 455.3 & 393.8 & 373 & \textbf{310.6} & 0.5 \\
Llama3.1-8B & 536.9 & \textbf{247.1} & 273.9 & \textbf{247.1} & 0.0 \\
FLAN-T5-Large & 453.3 & \textbf{314.7} & 432.5 & 348.8 & 0.0 \\
mT0-Large & 442.1 & 130.7 & 118.7 & \textbf{109.1} & 0.1 \\
\bottomrule
\end{tabular}
}
\end{table}

\begin{table}[H]
\caption{Latency (ms) at 98\% target model accuracy values on SQuAD1.1 for token-level ensemble cascades. All cascades defer to Llama3.1-70B.}
\autosizeTable{
\begin{tabular}{lcccc}
\toprule
Ensemble & Chow-Sum & Chow-Avg & Best Chow-Quantile & Best Quantile \\
\midrule
mT0-Large, Qwen2.5-1.5B & \textbf{135.9} & 146 & 184.6 & 0.0 \\
mT0-Large, Qwen2.5-1.5B, FLAN-T5-Large & 339.3 & \textbf{227.2} & 231.8 & 0.0 \\
Qwen2.5-1.5B, FLAN-T5-Large, Gemma3-1B & 398.1 & 392 & \textbf{389.6} & 0.0 \\
Qwen2.5-7B, Llama3.1-8B & 212.3 & 224.3 & \textbf{203.6} & 0.0 \\
Qwen2.5-7B, Llama3.1-8B, mT0-Large & \textbf{163.1} & \textbf{163.1} & \textbf{163.1} & 0.0 \\
Qwen2.5-7B, Llama3.2-3B & 467.4 & 455.9 & \textbf{415.2} & 0.5 \\
Qwen2.5-7B, Llama3.2-3B, Gemma3-1B & 513 & 540.7 & \textbf{450.7} & 0.5 \\
Qwen2.5-7B, mT0-Large, FLAN-T5-Large & 441.7 & \textbf{171.3} & 230.9 & 0.0 \\
\bottomrule
\end{tabular}
}
\end{table}

\begin{table}[H]
\caption{Latency (ms) at 98\% target model accuracy on SQuAD1.1 for various semantic cascades. All cascades defer to Llama3.1-70B.}
\autosizeTable{
\begin{tabular}{lllllll}
\toprule
Ensemble & BLEU & ROUGE-1 & ROUGE-2 & ROUGE-L & BLEURT & SBERT \\
\midrule
\tworow{mT0-Large, Qwen2.5-1.5B, \\ FLAN-T5-Large} & 122.2 & 98.95 & 217.9 & 98.95 & 149.8 & \textbf{86.65} \\
\tworow{mT0-Large, Qwen2.5-1.5B, \\ FLAN-T5-Large, Llama3.2-1B} & 152 & 122.2 & 214.6 & \textbf{117.5} & 179.1 & 159.9 \\
\tworow{Qwen2.5-1.5B, FLAN-T5-Large, \\ Gemma3-1B} & 291.4 & 285.4 & 329.2 & \textbf{285.1} & 291.4 & 316.6 \\
\tworow{Qwen2.5-1.5B, FLAN-T5-Large, \\ Gemma3-1B, Llama3.2-1B} & 327.2 & 314.1 & 314.4 & 314.4 & \textbf{289.2} & 335.4 \\
\tworow{Qwen2.5-1.5B, FLAN-T5-Large, \\ Llama3.2-1B} & 318.5 & 302.1 & 313.6 & 301.5 & \textbf{281.9} & 293.3 \\
\tworow{Llama3.2-3B, Qwen2.5-3B, \\ FLAN-T5-Large} & 216.4 & 232 & 250.8 & 236.1 & \textbf{176.2} & 239.1 \\
\tworow{Qwen2.5-7B, mT0-Large, \\ FLAN-T5-Large} & \textbf{163.1} & \textbf{163.1} & 213.9 & \textbf{163.1} & \textbf{163.1} & \textbf{163.1} \\
\tworow{Qwen2.5-7B, mT0-Large, \\ Llama3.2-3B} & 176.5 & \textbf{163.1} & 231.7 & \textbf{163.1} & \textbf{163.1} & \textbf{163.1} \\
\tworow{Qwen2.5-7B, Llama3.1-8B, \\ mT0-Large} & \textbf{163.1} & \textbf{163.1} & 200.8 & \textbf{163.1} & \textbf{163.1} & \textbf{163.1} \\
\tworow{Llama3.1-8B, mT0-Large, \\ Qwen2.5-1.5B} & 173.9 & 168.1 & 277.2 & 169.8 & 168.4 & \textbf{156.1} \\
\bottomrule
\end{tabular}
}
\end{table}

\subsubsection{TriviaQA}
\begin{table}[H]
\caption{Latency (ms) at 98\% target model accuracy values on TriviaQA for token-level cascades. All cascades defer to Llama3.1-70B. The Random column represents the expected value if queries are deferred uniformly at random.}
\autosizeTable{
\begin{tabular}{lccccc}
\toprule
Model & Random & Chow-Sum & Chow-Avg & Best Chow-Quantile & Best Quantile \\
\midrule
Qwen2.5-7B & 448.7 & \textbf{332.8} & 333.8 & 334.2 & 0.1 \\
Qwen2.5-3B & 402.7 & 335.9 & \textbf{333.4} & 335.5 & 0.1 \\
Qwen2.5-1.5B & 392.6 & 351.4 & \textbf{351.1} & 351.7 & 0.0 \\
Qwen2.5-0.5B & 386.2 & \textbf{377.4} & 378.8 & 378.2 & 0.0 \\
Gemma3-1B & 385.1 & \textbf{373} & 374.9 & 373.3 & 0.0 \\
Llama3.2-1B & 379.1 & 353.9 & \textbf{353.8} & 353.9 & 0.0 \\
Llama3.2-3B & 405.8 & 302.5 & 303.2 & \textbf{301.6} & 0.1 \\
Llama3.1-8B & 490.2 & \textbf{311.2} & 328 & 336.5 & 0.1 \\
FLAN-T5-Large & 388.8 & 372.3 & 379 & \textbf{371.4} & 0.0 \\
mT0-Large & 400.8 & \textbf{396.4} & 396.6 & 396.8 & 0.1 \\
\bottomrule
\end{tabular}
}
\end{table}

\begin{table}[H]
\caption{Latency (ms) at 98\% target model accuracy values on TriviaQA for token-level ensemble cascades. All cascades defer to Llama3.1-70B.}
\autosizeTable{
\begin{tabular}{lcccc}
\toprule
Ensemble & Chow-Sum & Chow-Avg & Best Chow-Quantile & Best Quantile \\
\midrule
Qwen2.5-1.5B, Llama3.2-1B & \textbf{356.8} & 358.4 & 357.7 & 0.0 \\
Llama3.2-1B, Gemma3-1B & 361.2 & 361.9 & \textbf{360.5} & 0.0 \\
Qwen2.5-1.5B, Llama3.2-1B, Gemma3-1B & 358.8 & 359.3 & \textbf{356.8} & 0.1 \\
Qwen2.5-1.5B, Llama3.2-1B, FLAN-T5-Large & \textbf{366.5} & 367.3 & 368.5 & 0.1 \\
Llama3.1-8B, Qwen2.5-7B & 359.3 & 390.7 & \textbf{354.9} & 0.1 \\
Llama3.1-8B, Qwen2.5-7B, mT0-Large & \textbf{490.3} & 492.4 & 491.9 & 0.0 \\
Qwen2.5-7B, Llama3.2-3B & 332.1 & 331.7 & \textbf{331.3} & 0.0 \\
Qwen2.5-7B, Llama3.2-3B, Gemma3-1B & 404.3 & 429.9 & \textbf{404} & 0.5 \\
Qwen2.5-7B, Gemma3-1B & \textbf{433.3} & 434.5 & 433.7 & 0.0 \\
Llama3.2-3B, Qwen2.5-3B & \textbf{321.7} & 331.5 & 325.6 & 0.1 \\
Llama3.1-8B, Llama3.2-3B, Qwen2.5-3B & 400.6 & 415.3 & \textbf{394.9} & 0.5 \\
\bottomrule
\end{tabular}
}
\end{table}

\begin{table}[H]
\caption{Latency (ms) at 98\% target model accuracy on TriviaQA for various semantic cascades. All cascades defer to Llama3.1-70B.}
\autosizeTable{
\begin{tabular}{lllllll}
\toprule
Ensemble & BLEU & ROUGE-1 & ROUGE-2 & ROUGE-L & BLEURT & SBERT \\
\midrule
\tworow{Qwen2.5-1.5B, Llama3.2-1B, \\ Gemma3-1B} & 360.6 & \textbf{354.7} & 378.3 & 355.4 & 357.2 & 361.5 \\
\tworow{Qwen2.5-1.5B, Llama3.2-1B, \\ FLAN-T5-Large} & 367.9 & \textbf{364.3} & 381.6 & 364.7 & 364.5 & 370.9 \\
\tworow{Llama3.2-1B, Gemma3-1B, \\ FLAN-T5-Large} & 368.7 & 364.4 & 380.7 & \textbf{364.3} & 364.4 & 369 \\
\tworow{Qwen2.5-1.5B, Llama3.2-1B, \\ Gemma3-1B, FLAN-T5-Large} & 364.2 & \textbf{358.5} & 379.7 & 361.2 & 358.6 & 365.5 \\
\tworow{Llama3.1-8B, Qwen2.5-7B, \\ Llama3.2-3B} & 329.5 & 314.7 & 397.9 & 315.2 & \textbf{313.6} & 338.7 \\
\tworow{Qwen2.5-7B, Llama3.2-3B, \\ Gemma3-1B} & 354.7 & 357.9 & 391.9 & \textbf{350.4} & 356 & 370 \\
\tworow{Qwen2.5-7B, Llama3.2-3B, \\ Qwen2.5-3B} & 328.4 & 320.9 & 390.2 & \textbf{320.8} & 323.5 & 338.8 \\
\tworow{Llama3.1-8B, Llama3.2-3B, \\ Qwen2.5-3B} & 346.1 & 334.3 & 403.5 & \textbf{334.2} & 337.1 & 362.5 \\
\tworow{Llama3.1-8B, Qwen2.5-7B, \\ mT0-Large} & 364.1 & 335 & 412.4 & \textbf{331.5} & 375.4 & 414.7 \\
\bottomrule
\end{tabular}
}
\end{table}

\subsection{Performances at 40\% Target Model FLOPs}

\subsubsection{WMT19 DE$\to$FR}
\begin{table}[H]
\caption{BLEURT at 40\% target model FLOPs values on WMT19 DE$\to$FR for token-level cascades. All cascades defer to Llama3.1-70B. The Random column represents the expected value if queries are deferred uniformly at random.}
\autosizeTable{
\begin{tabular}{lccccc}
\toprule
Model & Chow-Sum & Chow-Avg & Best Chow-Quantile & Best Quantile \\
\midrule
Qwen2.5-7B & 0.6906 & \textbf{0.6968} & 0.6948 & 0.2 \\
Qwen2.5-3B & 0.6667 & 0.6735 & \textbf{0.6743} & 0.3 \\
Qwen2.5-1.5B & 0.6506 & \textbf{0.6594} & 0.6583 & 0.3 \\
Qwen2.5-0.5B & 0.5483 & \textbf{0.5588} & 0.5585 & 0.5 \\
Gemma3-1B & 0.6441 & \textbf{0.6543} & 0.6537 & 0.3 \\
Llama3.2-1B & 0.5954 & 0.603 & \textbf{0.6036} & 0.2 \\
Llama3.2-3B & 0.6707 & 0.6773 & \textbf{0.6776} & 0.3 \\
Llama3.1-8B & 0.6985 & \textbf{0.7048} & 0.7043 & 0.2 \\
FLAN-T5-Large & 0.5584 & 0.5638 & \textbf{0.5643} & 0.3 \\
mT0-Large & 0.5799 & \textbf{0.588} & 0.5867 & 0.1 \\
\bottomrule
\end{tabular}
}
\end{table}

\begin{table}[H]
\caption{BLEURT at 40\% target model FLOPs values on WMT19 DE$\to$FR for token-level ensemble cascades. All cascades defer to Llama3.1-70B.}
\autosizeTable{
\begin{tabular}{lcccc}
\toprule
Ensemble & Chow-Sum & Chow-Avg & Best Chow-Quantile & Best Quantile \\
\midrule
Gemma3-1B, Qwen2.5-1.5B & 0.6458 & 0.6486 & \textbf{0.6527} & 0.1 \\
Gemma3-1B, Qwen2.5-1.5B, Llama3.2-1B & 0.6413 & 0.6446 & \textbf{0.6469} & 0.1 \\
Gemma3-1B, Qwen2.5-1.5B, mT0-Large & 0.641 & 0.6458 & \textbf{0.6503} & 0.2 \\
Gemma3-1B, Qwen2.5-1.5B, FLAN-T5-Large & 0.6424 & 0.6448 & \textbf{0.65} & 0.1 \\
Qwen2.5-1.5B, Llama3.2-1B, mT0-Large & 0.6456 & 0.6436 & \textbf{0.6457} & 0.3 \\
Llama3.1-8B, Qwen2.5-7B & 0.6924 & 0.6956 & \textbf{0.6958} & 0.0 \\
Llama3.1-8B, Qwen2.5-3B & 0.6871 & \textbf{0.6889} & 0.6876 & 0.2 \\
Llama3.1-8B, Qwen2.5-7B, FLAN-T5-Large & 0.6903 & 0.6929 & \textbf{0.6943} & 0.0 \\
Llama3.1-8B, Qwen2.5-7B, mT0-Large & 0.6913 & \textbf{0.6942} & 0.6927 & 0.0 \\
\bottomrule
\end{tabular}
}
\end{table}

\begin{table}[H]
\caption{BLEURT at 40\% target model FLOPs on WMT19 DE$\to$FR for various semantic cascades. All cascades defer to Llama3.1-70B.}
\autosizeTable{
\begin{tabular}{lllllll}
\toprule
Ensemble & BLEU & ROUGE-1 & ROUGE-2 & ROUGE-L & BLEURT & SBERT \\
\midrule
\tworow{Gemma3-1B, Qwen2.5-1.5B, \\ Llama3.2-1B} & 0.6468 & 0.6487 & 0.6478 & 0.6507 & \textbf{0.6712} & 0.6545 \\
\tworow{Gemma3-1B, Qwen2.5-1.5B, \\ mT0-Large} & 0.6473 & 0.6521 & 0.6512 & 0.6535 & \textbf{0.6801} & 0.6576 \\
\tworow{Qwen2.5-1.5B, Gemma3-1B, \\ FLAN-T5-Large} & 0.6467 & 0.6491 & 0.6496 & 0.6503 & \textbf{0.6796} & 0.6595 \\
\tworow{Qwen2.5-1.5B, Llama3.2-1B, \\ mT0-Large} & 0.6296 & 0.6339 & 0.6298 & 0.6331 & \textbf{0.6649} & 0.6413 \\
\tworow{Gemma3-1B, Qwen2.5-1.5B, \\ Llama3.2-1B, mT0-Large} & 0.6436 & 0.6483 & 0.6475 & 0.6498 & \textbf{0.6786} & 0.6545 \\
\tworow{Gemma3-1B, Qwen2.5-1.5B, \\ mT0-Large, FLAN-T5-Large} & 0.6442 & 0.6472 & 0.6482 & 0.6501 & \textbf{0.682} & 0.6571 \\
\tworow{Llama3.1-8B, Qwen2.5-7B, \\ FLAN-T5-Large} & 0.6934 & 0.6947 & 0.6942 & 0.6947 & \textbf{0.7081} & 0.6965 \\
\tworow{Llama3.1-8B, Qwen2.5-7B, \\ mT0-Large} & 0.6944 & 0.6956 & 0.6958 & 0.695 & \textbf{0.7112} & 0.6963 \\
\tworow{Llama3.1-8B, Qwen2.5-7B, \\ Qwen2.5-1.5B} & 0.6932 & 0.6946 & 0.6955 & 0.6944 & \textbf{0.7079} & 0.6957 \\
\tworow{Llama3.2-3B, Qwen2.5-3B, \\ Gemma3-1B} & 0.6676 & 0.6719 & 0.6703 & 0.6729 & \textbf{0.6937} & 0.6779 \\
\bottomrule
\end{tabular}
}
\end{table}

\subsubsection{WMT14 FR$\to$EN}
\begin{table}[H]
\caption{BLEURT at 40\% target model FLOPs values on WMT14 FR$\to$EN for token-level cascades. All cascades defer to Llama3.1-70B. The Random column represents the expected value if queries are deferred uniformly at random.}
\autosizeTable{
\begin{tabular}{lccccc}
\toprule
Model &  Chow-Sum & Chow-Avg & Best Chow-Quantile & Best Quantile \\
\midrule
Qwen2.5-7B & \textbf{0.7457} & 0.745 & 0.7454 & 0.0 \\
Qwen2.5-3B  & 0.737 & 0.7381 & \textbf{0.7388} & 0.1 \\
Qwen2.5-1.5B & 0.7367 & 0.7374 & \textbf{0.738} & 0.1 \\
Qwen2.5-0.5B & 0.7127 & \textbf{0.7188} & 0.7181 & 0.2 \\
Gemma3-1B & 0.7268 & 0.7291 & \textbf{0.7301} & 0.8 \\
Llama3.2-1B & 0.7078 & 0.7133 & \textbf{0.7134} & 0.1 \\
Llama3.2-3B & 0.7381 & \textbf{0.7393} & 0.7391 & 0.7 \\
Llama3.1-8B & 0.743 & 0.7441 & \textbf{0.7445} & 0.1 \\
FLAN-T5-Large & 0.7169 & \textbf{0.7245} & 0.7244 & 0.1 \\
mT0-Large & 0.7023 & \textbf{0.7099} & 0.7091 & 0.4 \\
\bottomrule
\end{tabular}
}
\end{table}

\begin{table}[H]
\caption{BLEURT at 40\% target model FLOPs values on WMT14 FR$\to$EN for token-level ensemble cascades. All cascades defer to Llama3.1-70B.}
\autosizeTable{
\begin{tabular}{lcccc}
\toprule
Ensemble & Chow-Sum & Chow-Avg & Best Chow-Quantile & Best Quantile \\
\midrule
Gemma3-1B, Qwen2.5-1.5B & 0.7334 & 0.7333 & \textbf{0.737} & 0.5 \\
Gemma3-1B, Qwen2.5-1.5B, Llama3.2-1B & \textbf{0.7263} & 0.7252 & 0.7259 & 0.1 \\
Gemma3-1B, Qwen2.5-1.5B, mT0-Large & 0.725 & 0.726 & \textbf{0.7276} & 0.2 \\
Gemma3-1B, Qwen2.5-1.5B, FLAN-T5-Large & 0.7317 & \textbf{0.7319} & 0.7314 & 0.3 \\
Qwen2.5-1.5B, Llama3.2-1B, mT0-Large & \textbf{0.7225} & 0.7197 & 0.7222 & 0.0 \\
Llama3.1-8B, Qwen2.5-7B & 0.7446 & \textbf{0.7449} & 0.7449 & 0.2 \\
Llama3.1-8B, Qwen2.5-3B & 0.7414 & 0.7411 & \textbf{0.7433} & 0.7 \\
Llama3.1-8B, Qwen2.5-7B, FLAN-T5-Large & \textbf{0.741} & 0.7371 & 0.7391 & 0.1 \\
Llama3.1-8B, Qwen2.5-7B, mT0-Large & 0.7343 & 0.731 & \textbf{0.7416} & 1.0 \\
\bottomrule
\end{tabular}
}
\end{table}

\begin{table}[H]
\caption{BLEURT at 40\% target model FLOPs on WMT14 FR$\to$EN for various semantic cascades. All cascades defer to Llama3.1-70B.}
\autosizeTable{
\begin{tabular}{lllllll}
\toprule
Ensemble & BLEU & ROUGE-1 & ROUGE-2 & ROUGE-L & BLEURT & SBERT \\
\midrule
\tworow{Gemma3-1B, Qwen2.5-1.5B, \\ Llama3.2-1B} & 0.7337 & 0.734 & 0.7336 & 0.7341 & \textbf{0.7382} & 0.7343 \\
\tworow{Gemma3-1B, Qwen2.5-1.5B, \\ mT0-Large} & 0.7345 & 0.7351 & 0.734 & 0.735 & \textbf{0.7402} & 0.7353 \\
\tworow{Qwen2.5-1.5B, Gemma3-1B, \\ FLAN-T5-Large} & 0.7366 & 0.7366 & 0.7359 & 0.737 & \textbf{0.7415} & 0.7376 \\
\tworow{Qwen2.5-1.5B, Llama3.2-1B, \\ mT0-Large} & 0.731 & 0.7323 & 0.7312 & 0.7328 & \textbf{0.7368} & 0.7335 \\
\tworow{Gemma3-1B, Qwen2.5-1.5B, \\ Llama3.2-1B, mT0-Large} & 0.7337 & 0.734 & 0.7338 & 0.7342 & \textbf{0.7398} & 0.7347 \\
\tworow{Gemma3-1B, Qwen2.5-1.5B, \\ mT0-Large, FLAN-T5-Large} & 0.735 & 0.7361 & 0.734 & 0.7358 & \textbf{0.741} & 0.7359 \\
\tworow{Llama3.1-8B, Qwen2.5-7B, \\ FLAN-T5-Large} & 0.7443 & 0.7443 & 0.7431 & 0.744 & \textbf{0.7473} & 0.7449 \\
\tworow{Llama3.1-8B, Qwen2.5-7B, \\ mT0-Large} & 0.7429 & 0.7431 & 0.7421 & 0.7432 & \textbf{0.7466} & 0.7446 \\
\tworow{Llama3.1-8B, Qwen2.5-7B, \\ Qwen2.5-1.5B} & 0.7438 & 0.7437 & 0.7443 & 0.744 & \textbf{0.7455} & 0.7444 \\
\tworow{Llama3.2-3B, Qwen2.5-3B, \\ Gemma3-1B} & 0.7379 & 0.7386 & 0.7377 & 0.7382 & \textbf{0.7423} & 0.7398 \\
\bottomrule
\end{tabular}
}
\end{table}

\subsubsection{WMT14 EN$\to$FR}
\begin{table}[H]
\caption{BLEURT at 40\% target model FLOPs values on WMT14 EN$\to$FR for token-level cascades. All cascades defer to Llama3.1-70B. The Random column represents the expected value if queries are deferred uniformly at random.}
\autosizeTable{
\begin{tabular}{lccccc}
\toprule
Model & Chow-Sum & Chow-Avg & Best Chow-Quantile & Best Quantile \\
\midrule
Qwen2.5-7B & 0.684 & \textbf{0.6903} & 0.6887 & 0.1 \\
Qwen2.5-3B & 0.665 & \textbf{0.6745} & 0.673 & 0.2 \\
Qwen2.5-1.5B & 0.6579 & 0.6659 & \textbf{0.6667} & 0.3 \\
Qwen2.5-0.5B & 0.5947 & \textbf{0.6027} & 0.6014 & 0.3 \\
Gemma3-1B & 0.6664 & \textbf{0.6701} & 0.67 & 0.2 \\
Llama3.2-1B & 0.6467 & \textbf{0.6528} & 0.6522 & 0.4 \\
Llama3.2-3B & 0.6774 & \textbf{0.6872} & 0.6843 & 0.1 \\
Llama3.1-8B & 0.6909 & 0.6956 & \textbf{0.6968} & 0.6 \\
FLAN-T5-Large & 0.5834 & \textbf{0.5896} & 0.5894 & 0.4 \\
mT0-Large & 0.6107 & \textbf{0.6186} & 0.6154 & 0.2 \\
\bottomrule
\end{tabular}
}
\end{table}

\begin{table}[H]
\caption{BLEURT at 40\% target model FLOPs values on WMT14 EN$\to$FR for token-level ensemble cascades. All cascades defer to Llama3.1-70B.}
\autosizeTable{
\begin{tabular}{lcccc}
\toprule
Ensemble & Chow-Sum & Chow-Avg & Best Chow-Quantile & Best Quantile \\
\midrule
Gemma3-1B, Qwen2.5-1.5B & 0.6673 & 0.6691 & \textbf{0.6705} & 0.1 \\
Gemma3-1B, Qwen2.5-1.5B, Llama3.2-1B & 0.6657 & 0.6692 & \textbf{0.6692} & 0.1 \\
Gemma3-1B, Qwen2.5-1.5B, mT0-Large & 0.6625 & 0.6649 & \textbf{0.6663} & 0.0 \\
Gemma3-1B, Qwen2.5-1.5B, FLAN-T5-Large & 0.6645 & 0.6626 & \textbf{0.667} & 0.0 \\
Qwen2.5-1.5B, Llama3.2-1B, mT0-Large & 0.6592 & 0.6634 & \textbf{0.6638} & 0.1 \\
Llama3.1-8B, Qwen2.5-7B & 0.6869 & \textbf{0.6903} & 0.6898 & 0.1 \\
Llama3.1-8B, Qwen2.5-3B & 0.6818 & 0.6854 & \textbf{0.6876} & 0.2 \\
Llama3.1-8B, Qwen2.5-7B, FLAN-T5-Large & 0.6855 & 0.689 & \textbf{0.6892} & 0.1 \\
Llama3.1-8B, Qwen2.5-7B, mT0-Large & 0.6843 & 0.6883 & \textbf{0.6899} & 0.1 \\
\bottomrule
\end{tabular}
}
\end{table}

\begin{table}[H]
\caption{BLEURT at 40\% target model FLOPs on WMT14 EN$\to$FR for various semantic cascades. All cascades defer to Llama3.1-70B.}
\autosizeTable{
\begin{tabular}{lllllll}
\toprule
Ensemble & BLEU & ROUGE-1 & ROUGE-2 & ROUGE-L & BLEURT & SBERT \\
\midrule
\tworow{Gemma3-1B, Qwen2.5-1.5B, \\ Llama3.2-1B} & 0.6623 & 0.6622 & 0.6639 & 0.6632 & \textbf{0.6787} & 0.6682 \\
\tworow{Gemma3-1B, Qwen2.5-1.5B, \\ mT0-Large} & 0.6615 & 0.6617 & 0.6619 & 0.6625 & \textbf{0.6823} & 0.6638 \\
\tworow{Qwen2.5-1.5B, Gemma3-1B, \\ FLAN-T5-Large} & 0.6575 & 0.6603 & 0.6601 & 0.6603 & \textbf{0.6785} & 0.6634 \\
\tworow{Qwen2.5-1.5B, Llama3.2-1B, \\ mT0-Large} & 0.6544 & 0.6513 & 0.6536 & 0.6542 & \textbf{0.6732} & 0.6541 \\
\tworow{Gemma3-1B, Qwen2.5-1.5B, \\ Llama3.2-1B, mT0-Large} & 0.6631 & 0.6626 & 0.6642 & 0.6644 & \textbf{0.6846} & 0.666 \\
\tworow{Gemma3-1B, Qwen2.5-1.5B, \\ mT0-Large, FLAN-T5-Large} & 0.6582 & 0.6602 & 0.6604 & 0.662 & \textbf{0.6821} & 0.6618 \\
\tworow{Llama3.1-8B, Qwen2.5-7B, \\ FLAN-T5-Large} & 0.6868 & 0.6861 & 0.6883 & 0.688 & \textbf{0.6982} & 0.6887 \\
\tworow{Llama3.1-8B, Qwen2.5-7B, \\ mT0-Large} & 0.6885 & 0.6859 & 0.6864 & 0.6865 & \textbf{0.6986} & 0.686 \\
\tworow{Llama3.1-8B, Qwen2.5-7B, \\ Qwen2.5-1.5B} & 0.6841 & 0.6858 & 0.6856 & 0.6855 & \textbf{0.6964} & 0.6853 \\
\tworow{Llama3.2-3B, Qwen2.5-3B, \\ Gemma3-1B} & 0.6791 & 0.6794 & 0.679 & 0.6795 & \textbf{0.6936} & 0.6817 \\
\bottomrule
\end{tabular}
}
\end{table}

\subsubsection{CNN/DailyMail}
\begin{table}[H]
\caption{ROUGE-L at 40\% target model FLOPs values on CNN/DailyMail for token-level cascades. All cascades defer to Llama3.1-70B. The Random column represents the expected value if queries are deferred uniformly at random.}
\autosizeTable{
\begin{tabular}{lccccc}
\toprule
Model & Chow-Sum & Chow-Avg & Best Chow-Quantile & Best Quantile \\
\midrule
Qwen2.5-7B & 0.2531 & 0.2578 & \textbf{0.259} & 0.8 \\
Qwen2.5-3B & \textbf{0.252} & 0.2511 & 0.2509 & 0.3 \\
Qwen2.5-1.5B & 0.2315 & 0.2511 & \textbf{0.2545} & 0.9 \\
Qwen2.5-0.5B & 0.2293 & 0.2416 & \textbf{0.2452} & 1.0 \\
Gemma3-1B  & 0.2411 & 0.2431 & \textbf{0.2441} & 0.9 \\
Llama3.2-1B & 0.2468 & \textbf{0.2479} & \textbf{0.2479} & 0.5 \\
Llama3.2-3B & 0.2505 & \textbf{0.2527} & \textbf{0.2527} & 0.1 \\
Llama3.1-8B & 0.251 & 0.2532 & \textbf{0.2534} & 0.9 \\
mT0-Large & 0.2513 & 0.2512 & \textbf{0.2524} & 0.8 \\
\bottomrule
\end{tabular}
}
\end{table}

\begin{table}[H]
\caption{ROUGE-L at 40\% target model FLOPs on CNN/DailyMail for token-level ensemble cascades. All cascades defer to Llama3.1-70B.}
\autosizeTable{
\begin{tabular}{lcccc}
\toprule
Ensemble & Chow-Sum & Chow-Avg & Best Chow-Quantile & Best Quantile \\
\midrule
Llama3.2-1B, mT0-Large & \textbf{0.2543} & 0.2494 & 0.2517 & 0.1 \\
Llama3.2-1B, Qwen2.5-1.5B & 0.2483 & 0.2457 & \textbf{0.2526} & 0.3 \\
Qwen2.5-1.5B, Gemma3-1B & 0.2454 & 0.2453 & \textbf{0.2549} & 1.0 \\
Llama3.2-1B, mT0-Large, Qwen2.5-1.5B & \textbf{0.2536} & 0.2448 & 0.2535 & 0.4 \\
Qwen2.5-7B, Llama3.1-8B & 0.2533 & 0.2538 & \textbf{0.2566} & 0.2 \\
Qwen2.5-7B, mT0-Large & 0.2533 & 0.2537 & \textbf{0.2561} & 0.2 \\
Qwen2.5-7B, Qwen2.5-3B & 0.2558 & 0.2542 & \textbf{0.2568} & 0.1 \\
\bottomrule
\end{tabular}
}
\end{table}

\begin{table}[H]
\caption{ROUGE-L at 40\% target model FLOPs on CNN/DailyMail for various semantic cascades. All cascades defer to Llama3.1-70B.}
\autosizeTable{
\begin{tabular}{lllllll}
\toprule
Ensemble & BLEU & ROUGE-1 & ROUGE-2 & ROUGE-L & BLEURT & SBERT \\
\midrule
\tworow{Llama3.2-1B, mT0-Large, \\ Qwen2.5-1.5B} & 0.2595 & \textbf{0.2611} & 0.2593 & 0.2588 & 0.2537 & 0.2587 \\
\tworow{Llama3.2-1B, mT0-Large, \\ Qwen2.5-0.5B} & \textbf{0.2551} & 0.2542 & 0.2533 & 0.2551 & 0.2478 & 0.2523 \\
\tworow{Qwen2.5-1.5B, Qwen2.5-0.5B, \\ Gemma3-1B} & 0.2525 & \textbf{0.2542} & 0.2482 & 0.2502 & 0.251 & 0.249 \\
\tworow{Llama3.2-1B, Gemma3-1B, \\ Qwen2.5-0.5B} & 0.2504 & 0.2507 & 0.2511 & \textbf{0.2519} & 0.2471 & 0.2497 \\
\tworow{Llama3.2-3B, Qwen2.5-3B, \\ Qwen2.5-0.5B} & 0.254 & \textbf{0.2563} & 0.255 & 0.255 & 0.2514 & 0.2539 \\
\tworow{Qwen2.5-7B, mT0-Large, \\ Llama3.2-3B} & \textbf{0.2605} & 0.2598 & 0.2581 & 0.2574 & 0.2556 & 0.2584 \\
\tworow{Qwen2.5-7B, Llama3.1-8B, \\ mT0-Large} & 0.2574 & 0.2578 & \textbf{0.2595} & 0.2585 & 0.2548 & 0.2583 \\
\tworow{Llama3.1-8B, mT0-Large, \\ Qwen2.5-3B} & \textbf{0.2543} & 0.2532 & 0.2543 & 0.2536 & 0.2501 & 0.2535 \\
\bottomrule
\end{tabular}
}
\end{table}

\subsubsection{SQuAD1.1} 
\begin{table}[H]
\caption{accuracy at 40\% target model FLOPs values on SQuAD1.1 for token-level cascades. All cascades defer to Llama3.1-70B. The Random column represents the expected value if queries are deferred uniformly at random.}
\autosizeTable{
\begin{tabular}{lccccc}
\toprule
Model & Chow-Sum & Chow-Avg & Best Chow-Quantile & Best Quantile \\
\midrule
Qwen2.5-7B & \textbf{0.8227} & 0.818 & \textbf{0.8227} & 0.0 \\
Qwen2.5-3B & 0.7833 & 0.7853 & \textbf{0.786} & 0.1 \\
Qwen2.5-1.5B & \textbf{0.788} & 0.7813 & 0.782 & 0.1 \\
Qwen2.5-0.5B & \textbf{0.67} & 0.6593 & 0.6647 & 0.0 \\
Gemma3-1B & \textbf{0.674} & 0.666 & 0.6727 & 0.0 \\
Llama3.2-1B & 0.638 & 0.648 & \textbf{0.6527} & 0.8 \\
Llama3.2-3B & 0.766 & 0.7687 & \textbf{0.7747} & 0.8 \\
Llama3.1-8B & \textbf{0.8207} & 0.814 & 0.816 & 0.1 \\
FLAN-T5-Large & \textbf{0.7567} & 0.6453 & 0.7247 & 0.0 \\
mT0-Large & 0.8267 & \textbf{0.8307} & \textbf{0.8307} & 0.2 \\
\bottomrule
\end{tabular}
}
\end{table}

\begin{table}[H]
\caption{accuracy at 40\% target model FLOPs values on SQuAD1.1 for token-level ensemble cascades. All cascades defer to Llama3.1-70B.}
\autosizeTable{
\begin{tabular}{lcccc}
\toprule
Ensemble & Chow-Sum & Chow-Avg & Best Chow-Quantile & Best Quantile \\
\midrule
mT0-Large, Qwen2.5-1.5B & 0.8153 & \textbf{0.8207} & 0.8113 & 0.0 \\
mT0-Large, Qwen2.5-1.5B, FLAN-T5-Large & 0.7807 & \textbf{0.8073} & 0.8013 & 0.0 \\
Qwen2.5-1.5B, FLAN-T5-Large, Gemma3-1B & 0.7193 & 0.724 & \textbf{0.746} & 0.0 \\
Qwen2.5-7B, Llama3.1-8B & 0.8207 & 0.8213 & \textbf{0.8227} & 0.0 \\
Qwen2.5-7B, Llama3.1-8B, mT0-Large & \textbf{0.824} & 0.8227 & 0.8233 & 0.1 \\
Qwen2.5-7B, Llama3.2-3B & 0.7867 & \textbf{0.7907} & 0.788 & 0.0 \\
Qwen2.5-7B, Llama3.2-3B, Gemma3-1B & 0.7273 & 0.7167 & \textbf{0.7607} & 1.0 \\
Qwen2.5-7B, mT0-Large, FLAN-T5-Large & 0.7747 & \textbf{0.816} & 0.8127 & 0.0 \\
\bottomrule
\end{tabular}
}
\end{table}

\begin{table}[H]
\caption{accuracy at 40\% target model FLOPs on SQuAD1.1 for various semantic cascades. All cascades defer to Llama3.1-70B.}
\autosizeTable{
\begin{tabular}{lllllll}
\toprule
Ensemble & BLEU & ROUGE-1 & ROUGE-2 & ROUGE-L & BLEURT & SBERT \\
\midrule
\tworow{mT0-Large, Qwen2.5-1.5B, \\ FLAN-T5-Large} & \textbf{0.8313} & 0.8233 & 0.8053 & 0.824 & 0.8227 & 0.8247 \\
\tworow{mT0-Large, Qwen2.5-1.5B, \\ FLAN-T5-Large, Llama3.2-1B} & 0.8227 & 0.826 & 0.8067 & \textbf{0.8273} & 0.8167 & 0.8187 \\
\tworow{Qwen2.5-1.5B, FLAN-T5-Large, \\ Gemma3-1B} & 0.7813 & 0.79 & 0.7653 & \textbf{0.7913} & 0.78 & 0.786 \\
\tworow{Qwen2.5-1.5B, FLAN-T5-Large, \\ Gemma3-1B, Llama3.2-1B} & 0.7667 & \textbf{0.7893} & 0.762 & 0.7887 & 0.7873 & 0.7713 \\
\tworow{Qwen2.5-1.5B, FLAN-T5-Large, \\ Llama3.2-1B} & 0.7833 & 0.7967 & 0.7573 & \textbf{0.7987} & 0.782 & 0.782 \\
\tworow{Llama3.2-3B, Qwen2.5-3B, \\ FLAN-T5-Large} & 0.808 & 0.808 & 0.7913 & 0.8087 & \textbf{0.8187} & 0.806 \\
\tworow{Qwen2.5-7B, mT0-Large, \\ FLAN-T5-Large} & 0.8427 & 0.8333 & 0.8207 & 0.8347 & 0.838 & \textbf{0.8433} \\
\tworow{Qwen2.5-7B, mT0-Large, \\ Llama3.2-3B} & 0.8287 & 0.83 & 0.8227 & 0.8287 & 0.8293 & \textbf{0.832} \\
\tworow{Qwen2.5-7B, Llama3.1-8B, \\ mT0-Large} & 0.836 & 0.8353 & 0.8207 & 0.834 & 0.8373 & \textbf{0.838} \\
\tworow{Llama3.1-8B, mT0-Large, \\ Qwen2.5-1.5B} & 0.83 & 0.8313 & 0.8107 & \textbf{0.832} & 0.8287 & 0.83 \\
\bottomrule
\end{tabular}
}
\end{table}

\subsubsection{TriviaQA}
\begin{table}[H]
\caption{accuracy at 40\% target model FLOPs values on TriviaQA for token-level cascades. All cascades defer to Llama3.1-70B. The Random column represents the expected value if queries are deferred uniformly at random.}
\autosizeTable{
\begin{tabular}{lccccc}
\toprule
Model &  Chow-Sum & Chow-Avg & Best Chow-Quantile & Best Quantile \\
\midrule
Qwen2.5-7B & \textbf{0.709} & 0.7017 & 0.7073 & 0.0 \\
Qwen2.5-3B & \textbf{0.6422} & 0.6373 & 0.6373 & 0.1 \\
Qwen2.5-1.5B  & \textbf{0.5796} & 0.5752 & 0.5767 & 0.1 \\
Qwen2.5-0.5B  & \textbf{0.4149} & 0.4014 & 0.4111 & 0.0 \\
Gemma3-1B  & \textbf{0.4928} & 0.4915 & 0.4906 & 0.1 \\
Llama3.2-1B  & \textbf{0.5249} & 0.5193 & 0.5208 & 0.2 \\
Llama3.2-3B  & \textbf{0.7137} & 0.7071 & 0.7113 & 0.1 \\
Llama3.1-8B  & \textbf{0.7763} & 0.7711 & 0.7726 & 0.1 \\
FLAN-T5-Large  & 0.4403 & 0.435 & \textbf{0.4438} & 0.0 \\
mT0-Large  & 0.3637 & 0.3623 & \textbf{0.3676} & 0.2 \\
\bottomrule
\end{tabular}
}
\end{table}

\begin{table}[H]
\caption{accuracy at 40\% target model FLOPs values on TriviaQA for token-level ensemble cascades. All cascades defer to Llama3.1-70B.}
\autosizeTable{
\begin{tabular}{lcccc}
\toprule
Ensemble & Chow-Sum & Chow-Avg & Best Chow-Quantile & Best Quantile \\
\midrule
Qwen2.5-1.5B, Llama3.2-1B & 0.5641 & \textbf{0.5708} & 0.5699 & 0.1 \\
Llama3.2-1B, Gemma3-1B & 0.5193 & 0.5229 & \textbf{0.5239} & 0.2 \\
Qwen2.5-1.5B, Llama3.2-1B, Gemma3-1B & 0.5519 & 0.5559 & \textbf{0.5571} & 0.0 \\
Qwen2.5-1.5B, Llama3.2-1B, FLAN-T5-Large & 0.5322 & 0.547 & \textbf{0.5487} & 0.0 \\
Llama3.1-8B, Qwen2.5-7B & 0.7301 & 0.7266 & \textbf{0.7401} & 0.0 \\
Llama3.1-8B, Qwen2.5-7B, mT0-Large & 0.5807 & 0.5478 & \textbf{0.6098} & 0.0 \\
Qwen2.5-7B, Llama3.2-3B & 0.7154 & \textbf{0.7216} & 0.7203 & 0.0 \\
Qwen2.5-7B, Llama3.2-3B, Gemma3-1B & 0.6377 & 0.6207 & \textbf{0.6465} & 0.0 \\
Qwen2.5-7B, Gemma3-1B & 0.5925 & 0.5922 & \textbf{0.6004} & 0.0 \\
Llama3.2-3B, Qwen2.5-3B & 0.6857 & 0.6852 & \textbf{0.6883} & 0.1 \\
Llama3.1-8B, Llama3.2-3B, Qwen2.5-3B & 0.7026 & 0.6889 & \textbf{0.7158} & 0.1 \\
\bottomrule
\end{tabular}
}
\end{table}

\begin{table}[H]
\caption{accuracy at 40\% target model FLOPs on TriviaQA for various semantic cascades. All cascades defer to Llama3.1-70B.}
\autosizeTable{
%
}
\end{table}

%Note that comparing to the single-best model from the ensemble is not an entirely fair comparison since the best model may change between tasks.

\end{document}